\documentclass[nohyperref]{article}

\usepackage{microtype}
\usepackage{graphicx}
\usepackage{subfigure}
\usepackage{booktabs} 

\usepackage{hyperref}

\usepackage[accepted]{icml2022}

\usepackage{amsmath}
\usepackage{amssymb}
\usepackage{mathtools}
\usepackage{amsthm}
\usepackage{soul}

\usepackage[capitalize,noabbrev]{cleveref}

\theoremstyle{plain}

\theoremstyle{definition}

\theoremstyle{remark}

\usepackage[textsize=tiny]{todonotes}

\icmltitlerunning{Quantification and Analysis of Layer-wise and Pixel-wise Information Discarding}

\begin{document}

\twocolumn[
\icmltitle{Quantification and Analysis of Layer-wise and Pixel-wise Information Discarding}



\icmlsetsymbol{equal}{*}
\renewcommand{\thefootnote}{\dag}
\icmlsetsymbol{cor}{\dag}

\begin{icmlauthorlist}
\textcolor{white}{\icmlauthor{}{yyy}}
\icmlauthor{Haotian Ma}{equal,xxx}
\icmlauthor{Hao Zhang}{equal,yyy}
\icmlauthor{Fan Zhou}{yyy}
\icmlauthor{Yinqing Zhang}{yyy}
\icmlauthor{Quanshi Zhang}{cor,yyy}
\end{icmlauthorlist}

\icmlaffiliation{xxx}{Southern University of Science and Technology, Shenzhen, China}
\icmlaffiliation{yyy}{Shanghai Jiao Tong University, Shanghai, China}

\icmlcorrespondingauthor{Quanshi Zhang}{qszhang@sjtu.edu.cn}

\icmlkeywords{Machine Learning, ICML}

\vskip 0.3in
]



\printAffiliationsAndNotice{\icmlEqualContribution} 


\begin{abstract}
This paper presents a method to explain how the information of each input variable is gradually discarded during the forward propagation in a deep neural network (DNN), which provides new perspectives to explain DNNs. We define two types of entropy-based metrics, \emph{i.e.} (1) the discarding of pixel-wise information used in the forward propagation, and (2) the uncertainty of the input reconstruction, to measure input information contained by a specific layer from two perspectives. Unlike previous attribution metrics, the proposed metrics ensure the fairness of comparisons between different layers of different DNNs. We can use these metrics to analyze the efficiency of information processing in DNNs, which exhibits strong connections to the performance of DNNs. We analyze information discarding in a pixel-wise manner, which is different from the information bottleneck theory measuring feature information \emph{w.r.t.} the sample distribution. Experiments have shown the effectiveness of our metrics in analyzing classic DNNs and explaining existing deep-learning techniques.
\emph{The code is available at \href{https://github.com/haotianSustc/deepinfo}{https://github.com/haotianSustc/deepinfo}.}
\end{abstract}

\section{Introduction}



\begin{figure}[t]
\centering
\includegraphics[width=0.99\linewidth]{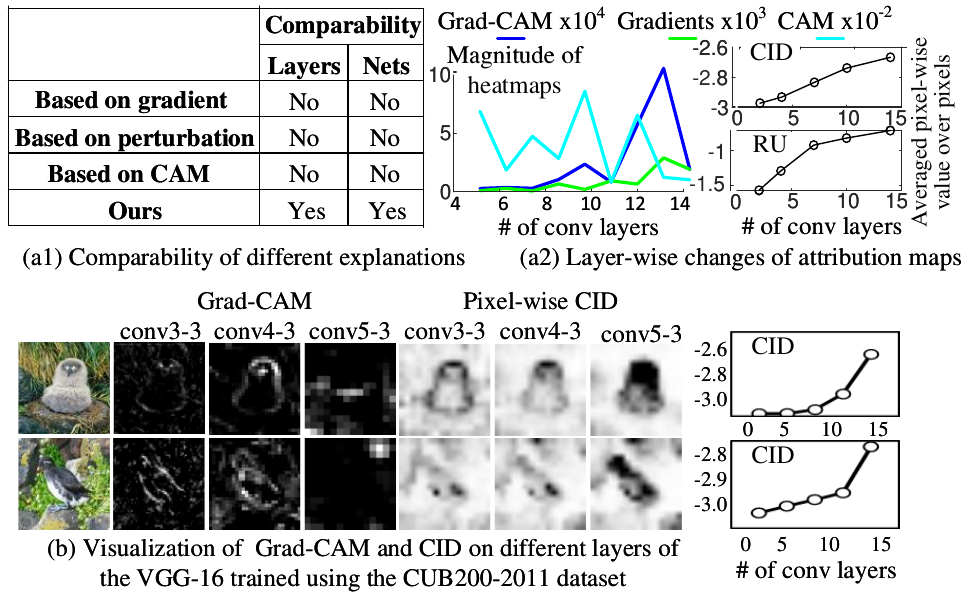}
\vspace{-10pt}
\caption{Subfigure (a1, a2) shows that our metrics (CID and RU) enable the fair comparison of the representation capacity of different layers. In comparison, the magnitude of explanations from previous methods is not comparable through different layers (see a2).
More analysis and proof are presented in Section~\ref{sec:other_coherency}. Subfigure (b) visualizes the CID of each input pixel and the Grad-CAM at different layers. Appendix~\ref{appsec:cam} has also shown importance maps generated by CAM, Gradient, and Grad-CAM on different layers of the DNN.}
\vspace{-15pt}
\label{fig:coherency}
\end{figure}

The interpretability of DNNs has received increasing attention in recent years. To this end, many methods have been proposed to measure the importance/saliency/attribution score of each input variable~\cite{visualCNN_grad_2, simonyan2013deep, shrikumar2016not, shapley1953value, springenberg2014striving, lundberg2017unified, shrikumar2016not, trust, visualCNN_grad, visualCNN_grad_2, CAM}.
However, these attribution maps lack the ability to reflect the representation capacity of intermediate-layer features.
For example, Figure~\ref{fig:coherency}(a2) shows that the magnitudes of these attribution maps among different layers are quite unstable, and therefore cannot objectively reflect layer-wise changes of the representation capacity of intermediate-layer features.


Therefore, instead of merely studying the attribution score, in this paper, we aim to quantify the representation capacity of intermediate features, which provides new insight into the explanation of DNNs.
To this end, the core challenge of quantifying the representation capacity is to ensure \textbf{the fair comparability of the representation capacity over different layers of the same DNN, or even over different DNNs.} We have shown that previous explanation methods cannot ensure the fairness of comparisons in Section~\ref{sec:other_coherency}. Therefore, in this study, we aim to explain how the information of each input variable is gradually discarded by intermediate-layer features during the forward propagation. Unlike previous studies, information discarding provides the following new perspectives to fairly compare features through different layers in DNNs.\\
\emph{1. Quantification of the pixel-wise information discarding:} In general, information propagation through the cascaded layers of a DNN can be considered as a process of information selection. Figure~\ref{fig:coherency}(b) shows that as the number of layers increases, the DNN discards more information.\\
\emph{2. Efficiency of information processing:} Based on our metrics, we develop a new method to quantify the efficiency of information processing of DNNs. Here, the efficiency refers to the efficiency of feature extraction of DNNs. For example, Figure~\ref{fig:coherency}(b) shows that from low layers to high layers, the DNN gradually shifts the attention from low-level concepts (edges) to middle-level concepts (parts), and to high-level concepts (objects).\\
\emph{3. Analysis of classic DNNs and classic deep-learning methods:} We use our metrics to evaluate the representation capacity of classic DNNs, and analyze the effectiveness of network compression and knowledge distillation.

\textbf{Metrics:} To quantify the discarded information of input variables, we design two new metrics as follows.

(1) The first metric aims to quantify how much information of each input pixel is used to compute the feature, namely \emph{pixel-wise computational information discarding} (pixel-wise CID).
The information discarding refers to the phenomenon that a DNN usually selectively discards redundant information of input units (\emph{e.g.} some pixels are not related to the task) when computing the intermediate-layer feature representation.
Recently,~\citet{NLPID} proposed a method to estimate the information discarding of words in natural language processing.
In this work, we extend the information discarding to the CID metric to quantify the discarded information of input pixels, and boost the fairness of layer-wise comparisons.

More crucially, based on the pixel-wise CID, we further develop a metric, namely \textit{concentration} to measure the efficiency of the information processing of a DNN. The concentration measures the relative magnitude of information discarding on the foreground \textit{w.r.t.} that on the background.
\textit{We theoretically explain and experimentally verify the relationship between the concentration metric and the efficiency of the information processing of the DNN (see Figure 3(a)).}

(2) The second metric aims to quantify how much input information can be recovered from the intermediate-layer feature, which is termed \emph{pixel-wise reconstruction uncertainty} (pixel-wise RU). The RU handles the following case. Some pixels may be discarded during the forward propagation, but their information can still be well recovered by other pixels due to information redundancy.



\textbf{Analysis of DNNs and findings:}
Unlike previous pixel-wise attribution metrics, the generality of the proposed metrics CID, RU, and concentration enables us to fairly compare DNNs, \emph{i.e.} fairly compare intermediate-layer features (1) between different DNNs, and (2) between different layers of the same DNN, as Figure\ref{fig:coherency} shows. It is because our metrics are all formulated in the form of entropy, which is a generic metric in information theory, and enables fair comparisons of the DNN's representation capacity. Furthermore, based on the metrics, we obtain the following finding.

\emph{Finding 1:} The last paragraph of Section~\ref{sec:SidAndConcentration} proves a close relationship between the concentration and the DNN's performance.

\emph{Finding 2:} Network compression makes the DNN less powerful to remove the information of redundant pixels, but it still maintains the representation power of the DNN, \emph{i.e.} the feature can still well reconstruct the input. On the other hand, the feature still concentrates on the foreground.

\emph{Finding 3:} Knowledge distillation helps DNNs to preserve more information.

Besides, Appendix~\ref{appsec:exp} also shows the proof of the relationship between the CID value and the adversarial noise.


\textbf{Connection to the information bottleneck theory:}
The information bottleneck theory~\cite{InformationBottleneck, InformationBottleneck2,tishby2015deep} quantifies the layer-wise feature information $I(X;F)$ and $I(F;Y)$ at the \emph{sample level}, where $X$ represents input samples, $Y$ represents ground-truth labels, and $F$ denotes intermediate-layer features. In comparison, our method measures fine-grained, \textbf{pixel-wise} information discarding through layer-wise propagation. More interestingly, we prove that the metric can represent the sample-wise efficiency of feature extraction \emph{w.r.t} $I(X;F)$, \emph{i.e.}, $I(F;Y)/I(X;F)$.

Contributions of this study can be summarized as follows. In this study, we propose metrics CID, concentration, and RU, to measure the discarding of input information during the forward propagation, in order to quantify the representation capacity between intermediate-layer features in a DNN. Our metrics enable fair comparisons of the representation capacity between different layers in different DNNs. Based on the proposed metrics, we analyze classic DNNs and deep learning techniques. Experiments have demonstrated the effectiveness of our method.

\section{Related work}
\textbf{Explaining DNNs visually or semantically:} The visualization of DNNs is the most direct way of explaining knowledge hidden inside a DNN~\cite{CNNVisualization_1,CNNVisualization_2, FeaVisual, CNNSemanticDeep, Interpretability, net2vector}.
Beyond visualization, attribution methods~\cite{simonyan2013deep, visualCNN_grad_2, visualCNN_grad, binder2016layer, trust, lundberg2017unified, springenberg2014striving, CAM} estimated image regions that directly contribute to the network output.
As Table~\ref{tab:relat_pre} shows, our research has an essential difference from previous attribution methods. We propose to use information discarding to analyze the representation capacity of the DNN and explain classic deep learning techniques. More crucially, we prove the close relationship between our metric and the information processing of the DNN.


\begin{table}[t]
\centering
\caption{Comparisons of objectives of different explanation methods. Unlike previous methods, our method aims to quantify the representation capacity of DNNs.}
\label{tab:relat_pre}
\resizebox{0.99\linewidth}{!}{
\begin{tabular}{c|c}
\hline
Objective & Methods \\
\hline
Feature importance & CAM, Grad-CAM  \\
\hline
Pixel attribution & LRP, Shapley value, SHAP, LIME, Gradient, Guided-BP \\
\hline
Information discarding & Our method \\
\hline
\end{tabular}
}
\vspace{-5pt}
\end{table}


\begin{figure}[t]
\centering
\includegraphics[width=0.95\linewidth]{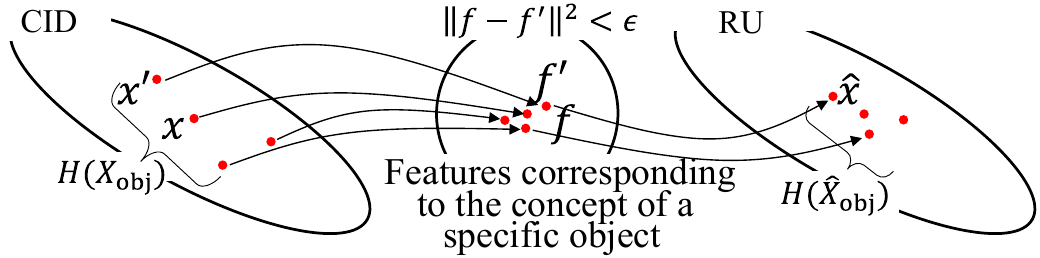}
\vspace{-5pt}
\caption{Illustration of the computation of CID and RU. Given a trained DNN, we compute the maximal entropy of the input $H(X_\text{obj})$ and the maximal entropy of image reconstruction $H(\hat{X}_\text{obj})$, when we constrain the intermediate-layer feature $f$ within a small range to represent a specific object instance.}
\vspace{-10pt}
\label{fig:concept}
\end{figure}

\textbf{Mathematical evaluation of the representation capacity:} Formulating and evaluating the representation capacity of DNNs is another emerging direction. The analysis of representation similarity between DNNs based on canonical correlation analysis is widely used to analyze DNN representations~\cite{RepresentationSimilarity2,RepresentationSimilarity3,RepresentationSimilarity4}.
~\citet{NetSensitivity} measured the sensitivity of network outputs \emph{w.r.t.} parameters of neural networks.
~\citet{NetRethinking} discussed the relationship between the parameter number and the generalization capacity of DNNs. Network-attack methods~\cite{CNNInfluence} could also be used to evaluate representation robustness by computing adversarial samples for a CNN.
~\citet{Schulz2020Restricting} and~\citet{InfoMask} analyzed the feature processing from the intermediate layer to the final output of the DNN, and computing attention on intermediate layers. In comparison, this paper focuses on the processing from the input to the intermediate layer.

In particular, the information-bottleneck theory~\cite{tishby1999the} provides a generic metric to quantify the information contained in DNNs. The information-bottleneck theory can be extended to evaluate the representation capacity of DNNs~\cite{InfoFlow,IBGeneralization}.
~\citet{InformationDropout} further used the information-bottleneck theory to revise the dropout layer in a DNN. Our study is also inspired by the information-bottleneck theory. Unlike analyzing the final output of a DNN in~\cite{InformationPlane}, we pursue new model-agnostic and task-agnostic metrics of input information to enable comparisons over different layers of networks in a pixel-wise manner.

\section{Analyze feature representations of DNNs}

In order to conduct comparative studies to analyze DNNs learned by various deep-learning techniques, in this section, we introduce three generic metrics, CID, concentration, and RU. Theoretically, these metrics can be applied to various tasks, but to simplify the story, we limit our discussions to the task of object classification.

The basic idea is that we represent metrics CID, concentration, and RU as the entropy of the input information, given the feature of a specific intermediate layer. In other words, the entropy measures the uncertainty of the input when the feature represents the same object instance, \emph{i.e.} how much input information can be discarded. Let {\small$x\in \mathbb{R}^{n}$} and {\small$f=h(x)\in\mathbb{R}^m$} denote the object instance and an intermediate-layer feature of the DNN, respectively.
We assume that the DNN represents a specific object instance {\small$x$} using a very limited range of features with an average feature {\small$f$}.
Similarly, there exists a latent space {\small$X_\text{obj}=\{x'|\Vert h(x')-f\Vert^2\le\epsilon\}$} for {\small$x$} that represents the same specific object, which ensures {\small$h(x')$} to localize in the manifold of feature $f$, where $x'$ represents the perturbed input around $x$. {\small$\epsilon$} is a small constant. {\small$p(x'|X=x)$} denotes the possibility of the perturbed input {\small$x'$} given the input {\small$x$}.
Let {\small$f'=h(x')\in\mathbb{R}^m$} denote the feature in the limited range.
Our method can be regarded to add perturbations to the input $x$ to approximate the domain of $f'$ (see Equation~\eqref{eqn:pixel_entropy}), subject to {\small$\Vert f'-f\Vert^2\le \epsilon$}.

Figure~\ref{fig:concept} illustrates the basic idea of the algorithm. We compute the entropy of the input (\emph{i.e.} the CID) when the input represents the same object instance. We also use features of the object instance to reconstruct the input {\small$\hat{x}=g(f)$} and measure the entropy of the reconstructed input (\emph{i.e.} the RU). In this way, two types of information discarding (CID and RU) of a specific layer can be represented using the same prototype formulation, \emph{s.t.} {\small$\Vert f'-f\Vert^2\!\!\le\!\!\epsilon$}, as follows,

\vspace{-10pt}
\begin{small}
\begin{equation}
H(X_\text{obj})=-\sum\nolimits_{x'}p(x'|X=x)\log p(x'|X=x)
\label{eqn:entropy}
\end{equation}
\end{small}

\subsection{CID, concentration, and efficiency}
\label{sec:SidAndConcentration}
The CID quantifies the discarding of input information during the \textbf{computation} of intermediate-layer features, which is derived from the entropy in Equation~(\ref{eqn:entropy}) from the perspective of feature extraction {\small$f=h(x)$}.
The core challenge is that the explicit low-dimensional manifold of features \emph{w.r.t.} the input {\small$x$} is unknown. Therefore, we approximate the manifold by adding noises to the original input {\small$x$}. Let {\small$x'$} denote new inputs around {\small$x$}, \emph{i.e.} {\small$x'=x+\Delta x, x'\in\mathbb{R}^n$}, which satisfy {\small$\Vert f'-f\Vert^2\le\epsilon$}.

Although strictly speaking, pixels in an input image are not independent, similar to~\cite{chen2018explaining}, we assume that {\small$\Delta x$} is a Gaussian noise for simplicity, thereby {\small$x'=x+\Delta x$} can be represented as {\small$x'\sim\mathcal{N}(\mu=x,\Sigma),\mu\in\mathbb{R}^n,\Sigma\in\mathbb{R}^{n\times n}$}.
This also relaxes the constraint to {\small$\text{Prob}(\Vert f'-f\Vert^2\le\epsilon)\ge 1-\tau$}, where {\small$\tau\ll 1$} is a tiny positive scalar.
Considering the local linearity within a small feature range of {\small$\epsilon$} and {\small$f=h(x)$}, {\small$\mu$} can be approximated as {\small$\mu=x$}.
Although different dimensions of the input can be dependent on each other, different dimensions of the added noise can be assumed to be independent of each other. Thus, we further simplify the covariance matrix as a diagonal matrix {\small$\Sigma=\text{diag}[\sigma_1^2,\ldots,\sigma_n^2]$} to ease the computation.
In this way, the CID can be decomposed to pixel-wise entropy.

\vspace{-10pt}
\begin{small}
\begin{equation}
H(X_\text{obj})=\sum\nolimits_{i=1}^{n}\!\!H_i(\sigma_i),\;
\textrm{s.t.}\;\text{Prob}(\Vert f'-f\Vert^2\le\epsilon)=1\!-\!\tau
\label{eqn:pixel_entropy}
\end{equation}
\end{small}where {\small$H_i(\sigma_i)=\log\sigma_i+C$}, {\small$C=\frac{1}{2}\log(2\pi e)$}. The relationship between Equations~\eqref{eqn:entropy} and~\eqref{eqn:pixel_entropy} is discussed in Appendix~\ref{appsec:eqn2}.
The overall CID value {\small$H(X_\text{obj})$} can be decomposed to the pixel-wise entropy (\textbf{pixel-wise CID}) {\small$\{H_i(\sigma_i)\}$}.
Figure~\ref{fig:coherency}(b) shows such information discarding of each input pixel.
A larger value of CID indicates that the DNN discards more input information during the forward propagation.

In real applications, the overall CID can be used to compare DNNs learned on the same dataset when different DNNs share the same input size $n$.
Our method follows the \textbf{maximum-entropy principle}, which maximizes {\small$H(X_\text{obj})$} subject to constraining features within the scope of a specific object instance {\small$\Vert f'-f\Vert^2\le\epsilon$}. \emph{I.e.} we enumerate all perturbation directions in $x'$ within a small variance of {\small$f'$}, in order to approximate the local manifold of {\small$f'$}. We use the Langrange multiplier to relax Equation~\eqref{eqn:pixel_entropy} as follows.

\vspace{-5pt}
\begin{small}
\begin{equation}
\begin{aligned}
Loss({\boldsymbol\sigma})&=\frac{1}{\delta_f^2}\underset{f'}{\mathbb{E}}\left[\Vert f'-f\Vert^2\right]-\lambda\sum\nolimits_{i=1}^{n}H_i(\boldsymbol\sigma)
\label{eqn:ign}
\end{aligned}
\end{equation}
\end{small}where {\small${\boldsymbol\sigma}=[\sigma_1,\ldots,\sigma_{n}]^{\top}$} is \emph{the parameter that we aim to learn}. {\small$\lambda$} is a positive scalar, and {\small$\delta_f^2=\lim_{\xi\rightarrow 0^+}\mathbb{E}_{x'\sim \mathcal{N}(x,\xi^2\bf{I})}[\|h(x')-f\|^2]/\xi^2$} is the inherent variance of intermediate-layer features, which is used for normalization. Note that {\small$\delta_f^2$} is only used to normalize the intermediate-layer feature, instead of normalizing the CID value.
We use {\small$x'=x+{\boldsymbol\sigma}\circ{\boldsymbol\delta}$, $\delta\sim\mathcal{N}(0,I)$} to simplify the computation of the gradient \emph{w.r.t.} {\small${\boldsymbol\sigma}$}, where {\small$\circ$} denotes the element-wise multiplication.
Equation~\eqref{eqn:ign} is tractable, and we can learn {\small{$\boldsymbol\sigma$}} via gradient descent.

\textbf{For fair layer-wise comparisons:} In order to ensure fair layer-wise comparisons, we need to control the value range of the first term in Equation~(\ref{eqn:ign}). Features of different layers need to be perturbed at a comparable level. To this end, we use {\small$\delta_f^2$} to normalize the first term in Equation~(\ref{eqn:ign}). In this way, the stop criterion of learning {\small${\boldsymbol\sigma}$} is given as
\begin{equation}
\label{eqn:lambda}
\min_\sigma\text{Loss}(\sigma)
\;\text{s.t.}\;\mathbb{E}_{\delta\sim\mathcal{N}(0,I)}[\Vert f'-f\Vert^2]\approx\beta\delta_f^2,\beta<\alpha,
\end{equation}
where {\small$\alpha$} is a positive scalar, and {\small$0<\beta<\alpha$} satisfies {\small$\mathbb{E}_{\delta\sim\mathcal{N}(0,I)}[\Vert f'-f\Vert^2]\approx\beta\delta_f^2$}.
The value of {\small$\lambda$} in Equation~\eqref{eqn:ign} is slightly adjusted (manually or automatically) to make {\small${\boldsymbol\sigma}$} satisfy {\small$\text{Prob}(\Vert f'-f\Vert^2\le\alpha\delta_f^2)>1-\tau$}. Specifically, {\small$\lambda$} is determined according to the value of {\small$\beta$}, and we will discuss the value of {\small$\beta$} in Section~\ref{sec:comparative}. Please see Appendix~\ref{appsec:eqn4} for more details about the derivation of Equation~\eqref{eqn:lambda}.

\textbf{Using the metric concentration to evaluate the efficiency of information processing:} Based on the CID, we design the concentration metric to evaluate the efficiency of the feature extraction of DNNs. Given an input image {\small$x$} containing both the target object and some background area, let {\small$\Lambda$} denote the ground-truth segment (or the bounding box) of the target object in {\small$x$}. {\small$\forall i\in\Lambda, x_{i}$} represents pixels within {\small$\Lambda$}. Thus, the concentration is formulated as follows.

\vspace{-10pt}
\begin{small}
\begin{equation}
\text{concentration}=\frac{1}{n-|\Lambda|}\sum_{i\not\in\Lambda}\left[H_i(\sigma_i)\right]-\frac{1}{|\Lambda|}\sum_{i\in\Lambda}\left[H_i(\sigma_i)\right]
\end{equation}
\end{small}Ideally, a DNN for object classification is supposed to discard background information, rather than foreground information. Note that this assumption cannot be applied to tasks depending on the background. Thus, the concentration measures the relative background information discarding \emph{w.r.t.} foreground information discarding, which reflects the efficiency of feature extraction.

\begin{figure*}[t]
\centering
\includegraphics[width=0.99\linewidth]{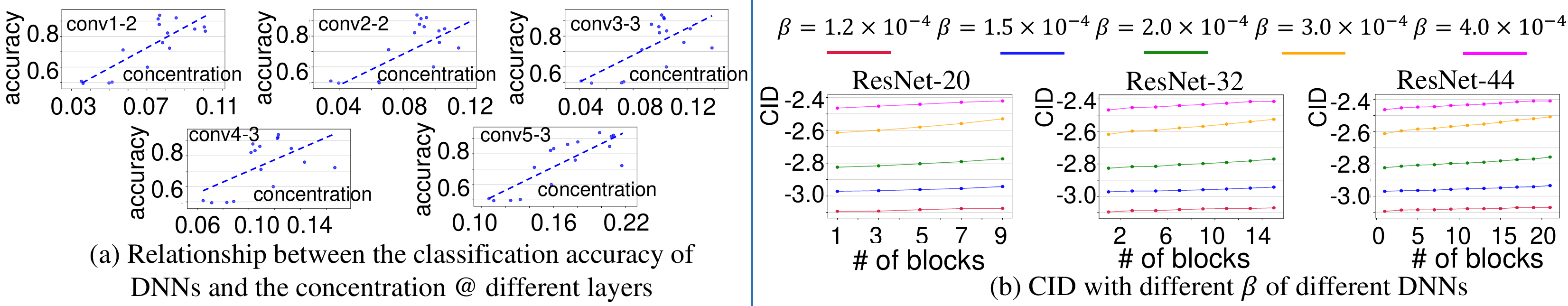}
\vspace{-5pt}
\caption{(a) The positive correlation between the concentration value and the classification accuracy of DNNs. Besides, the connection between the concentration and the efficiency of signal processing is explained in Appendix~\ref{appsec:information-bottleneck}. Each point corresponds to a DNN. We computed the concentration value of features in different layers of each DNN. As dashed lines show, a DNN with a high accuracy usually had a high concentration value.
(b) CID computed with different values of {\small$\beta$}. A high value of {\small$\beta$} in Equation~\eqref{eqn:lambda} led to a high value of CID. When we fixed a specific value of $\beta$, CID increased stably along with the number of blocks in the ResNet, which ensures convincing comparisons through layers.}
\vspace{-5pt}
\label{fig:AlphaAndPerformance}
\end{figure*}

\textbf{Theoretical connection of the connection between the concentration and the information bottleneck theory:}
The information bottleneck theory~\cite{InformationBottleneck,  InformationBottleneck2} formulates the relationship between the mutual information $I(X;F)$ and $I(F;Y)$, where $X$ denotes the input samples, and $Y$ denotes ground-truth labels.
Let $\rho=I(F;Y)/I(X;F)$ denote the efficiency of the extraction of the feature $F$. We can prove that a high concentration usually indicates a high value of efficiency. Specifically, we can roughly consider the foreground is related to the classification, while the background is not. Based on this, we can obtain the following relationship between the concentration value and the efficiency $\rho$ in the form of $\rho=C_1+\frac{C_2 \text{concentration}-C_3}{2[C_4-\text{CID}]}$,
where $C_1, C_2, C_3, C_4$ are four constants, and $C_2>0$, $C_4>\text{CID}$.
Please see Appendix~\ref{appsec:information-bottleneck} for the proof and more discussions.
In addition, we find that the concentration has a close relationship with the performance of DNNs. As Figure~\ref{fig:AlphaAndPerformance} (a) shows, DNNs with better performance usually had a higher concentration.

\subsection{Reconstruction uncertainty}

The metric of RU is also derived from the entropy in Equation~(\ref{eqn:entropy}). The CID focuses on the input information used to compute a feature, while the RU describes the discarding of input information that can be recovered from the feature. Due to the redundancy of the input information, a pixel may be well recovered from the feature, even when the pixel is not used for feature extraction.

We use a decoder net {\small$g$} to reconstruct the input {\small$\hat{x}'=g(f')$}. We consider the reconstructed result {\small$\hat{x}'$} as the information represented by {\small$f'$}. Although the architecture of $g$ affects the measurement of RU, RU values are still comparable through different layers and between DNNs when we fix $g$'s architecture in all comparisons. Thus, the metric RU can guarantee the fairness of layer-wise comparison (see Figure~\ref{fig:coherency}(a2)). Given a target DNN, $g$ is pre-trained using the MSE loss {\small$Loss^{\rm dec}=\Vert x'-\hat{x}'\Vert^2$}. In this way, the RU is formulated as the entropy of the reconstruction {\small$\hat{x}'=g(f')$}.

\vspace{-10pt}
\begin{small}
\begin{equation}
\begin{aligned}
H(\hat{X}_\text{obj})\!=\!-\!\!\sum_{\hat{x}'}\!p(\hat{x}')\log p(\hat{x}')\;\textrm{s.t.}\; \text{Prob}(\Vert f'\!-\!f\Vert^2\le\epsilon)
\!=\!\!1\!-\!\tau\!\!
\end{aligned}
\end{equation}
\end{small}where {\small$\hat{X}_\text{obj}$} denotes a set of images that are reconstructed using intermediate-layer features.
The above entropy {\small$H(\hat{X}_\text{obj})$} is computed in the same manner as the quantification of the CID. First, we synthesize the feature distribution {\small$F_\text{obj}$} by assuming that inputs follow a Gaussian distribution {\small$x'\sim\mathcal{N}(\mu=x,\Sigma)$, $f'=h(x')$}. {\small$\hat{x}'=g(f')$} denotes the reconstructed result using {\small$f'$}. Second, we can also assume {\small$\hat{x}'$} follows a Gaussian distribution with i.i.d. random variables {\small$\mathcal{N}(\mu^{\rm rec}=x,\Sigma^{\rm rec})$}.
As a result, the entropy of RU {\small$H(\hat{X}_\text{obj})$} can be decomposed into each pixel.

\begin{figure}[t]
\centering
\includegraphics[width=0.9\linewidth]{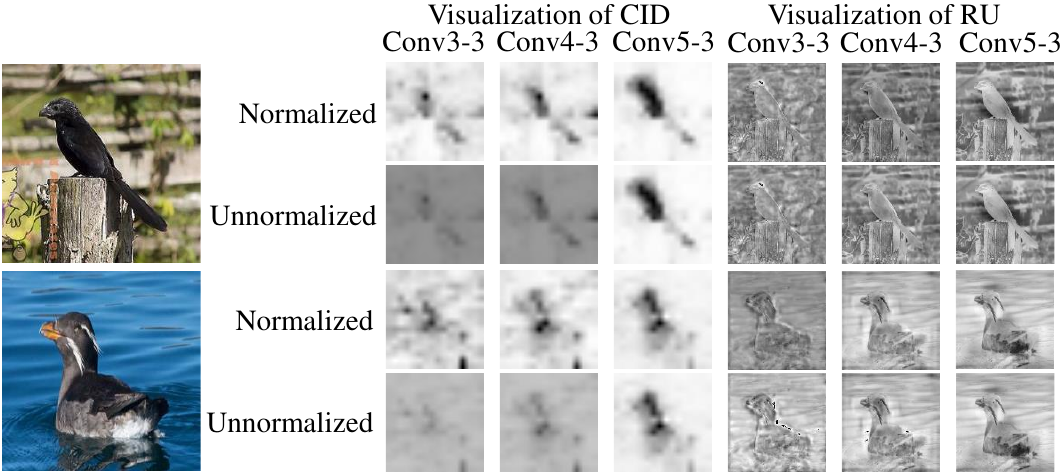}
\vspace{-5pt}
\caption{Visualization of pixel-wise CID and RU of different layers. The visualized pixel-wise CID and pixel-wise RU have been normalized to the value range of $[0,1]$ to clarify the difference between foreground and background. We find that low layers mainly focus on local patterns, and high layers mainly focus on large-scale patterns. We further visualize unnormalized CID and RU values to fairly compare the information discarding between different layers. Please see Appendix~\ref{appsec:visCID} and Appendix~\ref{appsec:visRU} for more results.}
\label{fig:CIDRUConcentration-vis}
\vspace{-5pt}
\end{figure}

\vspace{-10pt}
\begin{small}
\begin{equation}
\begin{aligned}
&H(\hat{X}_\text{obj})=\sum\nolimits_{i=1}^{n} \hat{H}_i({\boldsymbol\sigma}),\qquad\hat{H}_i({\boldsymbol\sigma})=\log\hat{\sigma}_i+C
\\=\frac{1}{2}&\log\Big(\underset{\substack{x'\sim\mathcal{N}(\mu=x,\Sigma=diag[\sigma_1,\sigma_2,\ldots])}}{\mathbb{E}}\!\!\left[\Vert \mu^{\rm rec}_i-\hat{x}'_i\Vert^2\right]\Big)\!\!+\!\!C
\end{aligned}
\end{equation}
\end{small}{\small$\hat{H}_i({\boldsymbol\sigma})$} is referred to as the \textbf{pixel-wise RU} for the $i$-th pixel (unit) in the input (see Figure~\ref{fig:CIDRUConcentration-vis}). Just like the CID, {\small$H(\hat{X}_\text{obj})$} is also estimated via the maximum-entropy principle.

\vspace{-5pt}
\begin{small}
\begin{equation}
Loss({\boldsymbol\sigma})=\frac{1}{\delta_f^2}\underset{f'}{\mathbb{E}}\left[\Vert f'-f\Vert^2\right]-\lambda\sum_{i=1}^{n}\hat{H}_i({\boldsymbol\sigma})\\
\nonumber
\end{equation}
\end{small}
We use the learned $\boldsymbol\sigma$ to compute {\small$\hat{H}_i({\boldsymbol\sigma})$} as the pixel-wise RU.
Like the computation of CID, $\lambda$ is also adjusted to ensure {\small$\mathbb{E}_{f'}[\Vert f'-f\Vert^2]\approx\beta\delta_f^2$}.
The above equation is tractable and can be solved by gradient descent.

\subsection{Discussions}
\label{subsec:discussions}
\textbf{Relationship between CID and RU:} CID and RU seem to be similar metrics, but they may be significantly different in some cases. Let us consider the following two cases.
(1) In the first case, redundant pixels ignored by the DNN increase the CID value, but they may still be well recovered via input reconstruction.
A toy example is that given an image with a white wall, and a white pixel in the wall has the same color as its neighboring pixels. If a DNN assigns a zero weight to this pixel, then we can consider this pixel is ignored by the DNN. Thus, this pixel will have an infinite CID value.
However, because this pixel and its neighboring pixels have the same color, this pixel can still be well reconstructed based on its neighboring pixels. In this case, the RU value of this pixel is still low.
(2) In the second case, pixels used for feature extraction may not be reconstructed. An example is the following function {\small$f=h(x)=\sum_i x_i$}, where all pixels are used to compute the feature $f$. However, no pixel can be well reconstructed from $h(x)$.

\textbf{Relationship between the CID and the metric in \cite{NLPID}:} Guan \emph{et al.}~\cite{NLPID} also measured the entropy of the input information, but there was no quantitative definition for the range of the target object. In other words, for each intermediate layer, the entropy may be measured within a different range of features, which significantly hurts the fairness of layer-wise comparisons. In comparison, we clearly define the feature range $\epsilon=\beta\delta_f^2$ to enable fair layer-wise comparisons.

\textbf{About information discarding in invertible networks:} Strictly speaking, there is no strict way to quantify the discarding of the input information during the computation of an intermediate-layer feature. The RU metric is related to image inversion based on invertible nets~\cite{invertible,invertible4,invertible5}, which also focuses on whether the feature can recover the input (theoretically, the decoder $g$ can be implemented as the inversion operations in invertible nets). In comparison, the CID metric is defined from another perspective, \emph{i.e.} whether the input information can contribute significant numerical values to the intermediate-layer feature or the final output. Please see Appendix~\ref{appsec:invertible} for details.

\textbf{Relationship with perturbation-based methods:} Our method is related to \cite{du2018towards,visualCNN_grad}. These studies extract input pixels responsible for the intermediate-layer feature by deleting as many input pixels as possible while keeping the feature unchanged.
They remove inputs by replacing inputs with human-designed values, which actually are not always meaningless.
Du \emph{et al.}~\cite{du2018towards}, Fong and Vedaldi~\cite{visualCNN_grad} computed pixel-wise importance.
However, these methods did not enable fair comparison over layers or evaluate the representation capacity of DNNs. Please see Section~\ref{sec:other_coherency} for details.
In comparison, our entropy-based metrics can provide fair comparisons without specific requirements for model parameters, model architectures, and tasks.

\begin{figure*}[t]
\centering
\includegraphics[width=0.99\linewidth]{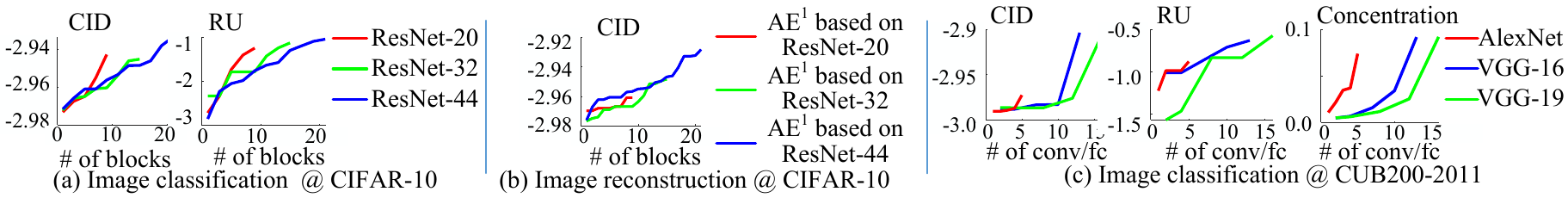}
\vspace{-10pt}
\caption{Layer-wise CID, RU, and concentration. Subfigure (a), (b) show that a deep DNN has high CID and RU values. Subfigure (c) shows that high layers can be more concentrated on the foreground than low layers.}
\label{fig:SIDRUConcentration}
\vspace{-10pt}
\end{figure*}

\textbf{High CID $\rightarrow$ robustness:} We can regard the forward propagation as a process of gradually discarding noisy information in the input that is irrelevant to the task, in order to extract features relevant to the task. In other words, a high CID value usually indicates that the DNN has discarded a large amount of noisy information, making the extracted features robust to noises.
Specifically, people usually understand the robustness of DNNs in two aspects. The first aspect mainly considers whether the DNN's output is largely influenced by noises, and the second aspect is whether the DNN can exhibit discrimination power on noisy samples. Strictly speaking, we can conceptually disentangle the two aspects of robustness. \emph{I.e.}, the first aspect cares about the insensitivity to noises, and even a toy model {\small$\forall x, h(x)=0$} can be considered the most robust model, although it does not have any discrimination power. Whereas, the second aspect cares about the classification accuracy under noises, no matter how large the output score is changed by the noise. Appendix~\ref{appsubsec:advnoise} shows the experiment proving the relationship between the CID and the first aspect of robustness.

\textbf{Limitations of concentration and RU:} The concentration metric is based on the assumption that the information in the foreground is related to the task. Therefore, concentration is not suitable for tasks depending on the background. Besides, we admit that DNNs with different architectures usually need different decoder architectures. However, theoretically, our algorithm can be adapted to different decoders. Although we use the \textit{same decoder} to fairly compare different DNNs, we still conduct experiments to test decoders with different architectures.
We find that RU values do not change significantly over different decoders, which proves the trustworthiness of RU (see Appendix~\ref{appsec:diffdecoder} for results).

\textbf{Computational cost} of the CID and RU is comparable with classical explanation methods, such as IG~\cite{sundararajan2017axiomatic} and LIME~\cite{trust}. Please see Appendix~\ref{appsec:cost} for details.

\subsection{Fairness of layer-wise comparisons}
\label{sec:other_coherency}
In this section, we discuss the fairness of layer-wise comparisons of existing explanation metrics, as follows.
\\
$\bullet$ SHAP~\cite{lundberg2017unified} is an explanation metric based on the Shapley value~\cite{shapley1953value}.
The Shapley value directly measures the numerical contribution of each input variable to the network output, instead of the contribution to the intermediate-layer feature. Thus, the Shapley value cannot be directly used to compare the attention distribution of different layers. Besides, according to the efficiency axiom~\cite{shapley1953value}, the sum of Shapley values of all input variables is equal to the output score. In other words, the Shapley value is sensitive to the magnitude of the network output, which disables fair comparisons between different DNNs.
\\
$\bullet$ LRP~\cite{bach2015on} computes the relevance score of each variable by layer-wise
relevance propagation. Compared with our metrics, the LRP mainly estimates the attention distribution over input variables, rather than explaining the information flow inside DNNs. Therefore, the LRP cannot examine the DNN's capacity of memorizing input information.
\\
$\bullet$
Most gradient-based methods do not generate explanations that ensures the fairness of layer-wise comparisons, such as CAM~\cite{CAM}, Grad-CAM~\cite{visualCNN_grad_2}, and gradient explanations~\cite{simonyan2013deep}. It is because the gradient map $\frac{\partial Loss}{\partial f}$ cannot ensure the fairness of comparison between different layers. Theoretically, we can easily construct two DNNs representing exactly the same knowledge but with different magnitudes of gradients, as follows.
A VGG-16 was learned to classify birds based on the CUB200-2011 dataset~\cite{CUB200}.
Given a pre-trained DNN, we slightly revised the magnitude of parameters in every pair of neighboring convolutional layers {\small$y=x\otimes w+b$} to examine our metrics.
For the {\small$L$}-th and {\small$(L+1)$}-th layers, parameters were revised as {\small$w^{(L)}\leftarrow w^{(L)}/4$}, {\small$w^{(L+1)}\leftarrow 4w^{(L+1)}$}, {\small$b^{(L)}\leftarrow b^{(L)}/4$}, {\small$b^{(L+1)}\leftarrow 4b^{(L+1)}$}.
Such revisions did not change knowledge representations or the network output, but changed the gradient magnitude. As Figure~\ref{fig:coherency} (a2) shows, magnitudes of explanation results of baseline methods are sensitive to the magnitude of parameters. In comparison, our metrics are not affected by the magnitude of parameters, and produce reliable results. Therefore, our metrics enable layer-wise comparisons.

\section{Comparative studies}
\label{sec:comparative}

We designed various experiments, in order to demonstrate the utility of the proposed metrics in comparing feature representations of various DNNs, analyzing inner mechanisms of knowledge distillation, and network compression.
In order to learn the parameter $\sigma$, we used the learning rate $1\times 10^{-4}$, and learned $\sigma$ for $100$ epochs.

In all experiments for image classification, we used object images cropped by object bounding boxes for both training and testing, except for experiments of computing concentration in Figure~\ref{fig:SIDRUConcentration} where images were cropped by the box of {\small$1.5\,width\times 1.5\,height$} of the object, which was similar to~\cite{interpretableCNN}. For the computation of RU, all experiments used a decoder with six residual blocks. We have tested decoders with different architectures, \emph{e.g.} ResNet with different numbers of blocks. The decoder with six residual blocks had enough sophisticated architecture for feature inversion, and was relatively easy to learn. Thus, we used this decoder in experiments. To invert low-resolution features back to high-resolution images, we added two transposed conv-layers to two parallel tracks in the residual block to enlarge the feature map. Considering the size of the input feature of the decoder, we added transposed conv-layers to the first 2--4 residual blocks.
The effects of {\small$\alpha$} and {\small$\tau$} is controlled by {\small$\beta$}, and Figure~\ref{fig:AlphaAndPerformance} (b) shows that a low value of {\small$\beta$} led to a low value of CID. For a specific {\small$\beta$}, the CID stably increased along with the number of blocks. Therefore, the selection of {\small$\beta$} did not affect the conclusion when we used CID to analyze DNNs. In the following experiments, we set {\small$\beta=1.5\times 10^{-4}$}. Figure~\ref{fig:SIDRUConcentration} visualizes the pixel-wise CID and RU for VGG-16 on the CUB200-2011 datasets.
We also applied the CID to the U-Net~\cite{UNet} trained for segmenting neuronal structures in medical images as a real-world application. The U-Net is trained using images in the ISBI cell tracking challenge~\cite{ISBI}, and we visualized the pixel-wised CID in Appendix~\ref{appsec:visCID}.
Appendix~\ref{appsec:visCID} also shows pixel-wise CID and RU for DNNs learned on the ImageNet dataset~\cite{ILSVRC15}.

\textbf{Comparisons between different DNNs for various tasks:} We compared layer-wise measures of CID and RU of different DNNs. We trained various DNNs for image classification using different datasets, and trained auto-encoders{\footnote[1]{To construct the auto-encoder, the encoder was set as all layers of the residual network before the FC layer. The decoder was the same as that for the computation of RU.}} (AEs) for image reconstruction (by revising architectures of ResNets-20/32/44~\cite{ResNet}). Figure~\ref{fig:SIDRUConcentration} (a), (b) compares input information discarding of intermediate layers of both DNNs for classification and DNNs for reconstruction. We found that the CID curve of image classification and the curve of image reconstruction were similar. \emph{A deep DNN usually had higher CID and RU values than a shallow DNN. Thus, a deep DNN usually discards more input information than a shallow DNN.}

Figure~\ref{fig:SIDRUConcentration} (c) illustrates the layer-wise concentration of various DNNs, which were learned to classify birds in the CUB200-2011~\cite{CUB200} dataset. Compared to the AlexNet~\cite{CNNImageNet}, we found that VGG nets~\cite{VGG} distracted attention to the background to learn diverse features in low layers, but more concentrated on the foreground object in high layers. Besides, curves became sharp at the last few layers, which indicated that \emph{fully-connected layers made the DNN quickly discard information that is irrelevant to the task.}


\begin{figure}[t]
 \centering
 \begin{minipage}[c]{.5\linewidth}
  \centering
  \vspace{-5pt}
  \includegraphics[width=\linewidth]{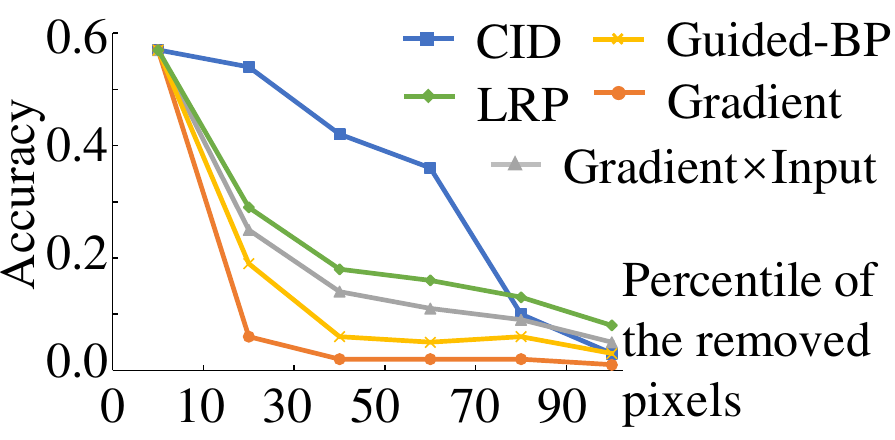}
 \end{minipage}%
 \hfill
 \begin{minipage}[c]{.47\linewidth}
 \centering
 \vspace{-10pt}
 \caption{Accuracy of the DNN when we gradually removed pixels with the lowest importance value. A slower decrease of the accuracy indicates a higher descriptive accuracy.}
 \label{fig:RankOrder}
\end{minipage}
\vspace{-10pt}
\end{figure}


\begin{table}[t]
\centering
\begin{minipage}[c]{.4\linewidth}
\resizebox{\linewidth}{!}{
\begin{tabular}{|l|cc|}
\hline
Model & ResNet-50 & VGG-16\\
\hline
CAM & 0.367 & - \\
Grad-CAM & 0.355 & 0.507 \\
LRP & - & 0.489 \\
CID & \textbf{0.493} & \textbf{0.578}\\
\hline
\end{tabular}
}
\end{minipage}
\hfill
\begin{minipage}[c]{.58\linewidth}
\vspace{-5pt}
\caption{Accuracy of the weakly-supervised localization task. A higher value indicates a better object-localization performance.}
\end{minipage}
\vspace{-15pt}
\label{tab:local}
\end{table}

\begin{figure*}[!ht]
\centering
\includegraphics[width=0.99\linewidth]{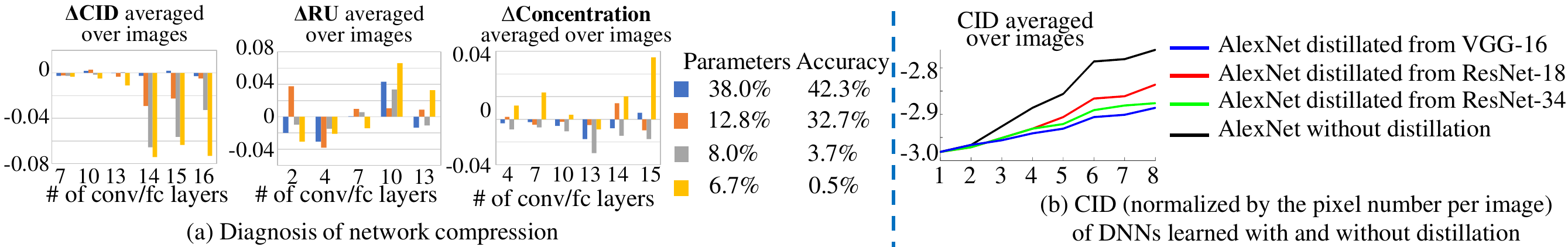}
\vspace{-10pt}
\caption{Analysis of network compression (a) and knowledge distillation (b). Subfigure (a) shows the change of CID, RU, and concentration after compression. We found that network compression decreased the CID, but there was no clear conclusion about the influence on the value of RU and concentration.
Subfigure (b) compares layerwise information discarding between DNNs learned with and without distillations. AlexNet distilled from other DNNs discarded less information.}
\label{fig:CompressAndDistill}
\vspace{-5pt}
\end{figure*}

\textbf{Evaluation using the descriptive accuracy~\cite{warnecke2020evaluating}:} In order to verify the effectiveness of the proposed metric, we evaluated the CID using the \emph{descriptive accuracy}~\cite{warnecke2020evaluating}. Specifically, we gradually removed pixels with the lowest importance values, and measured the accuracy of the DNN using images after the removal. In this case, a slower decrease of the accuracy indicated a higher descriptive accuracy. We conducted this experiment with the AlexNet trained on the CUB200-2011 dataset. We compared CID with various explanation methods, including Gradient~\cite{simonyan2013deep}, Gradient$\times$Input~\cite{shrikumar2016not}, Guided-BP~\cite{springenberg2014striving}, and LRP~\cite{binder2016layer}. As Figure~\ref{fig:RankOrder} shows, the CID outperformed other methods.

\textbf{Weakly-supervised localization:} We further evaluated the CID via the weakly-supervised localization task~\cite{CAM}.
We trained the VGG-16 and ResNet-50 using uncropped images from the CUB200-2011 dataset, and used ground-truth object bounding boxes for evaluation.
We compared CID with several previous methods, including the CAM~\cite{CAM}, Grad-CAM~\cite{visualCNN_grad_2}, LRP~\cite{binder2016layer}.
We followed~\cite{Schulz2020Restricting} to evaluate localization results by measuring the recall rate of pixels in the bounding box, \emph{i.e.} the number of pixels in the bounding box with high importance values.
Table~\ref{tab:local} shows the result of the evaluation. CID had a better performance than CAM, Grad-CAM, and LRP. Note that CID was not proposed to localize objects in the image. Instead, CID aimed to measure the information discarding during the forward propagation.

\textbf{Analysis of network compression:} We used our metrics to analyze the compressed DNN. We trained another VGG-16 using the CUB200-2011 dataset~\cite{CUB200} for fine-grained classification. Then, the VGG-16 was compressed using the method of~\cite{NetCompress} with different pruning thresholds. Figure~\ref{fig:CompressAndDistill} (a) compares layerwise information discarding of the original VGG-16 and the compressed VGG-16 nets with different numbers of parameters.
Specifically, let {\small$\text{CID}_\text{compressed net}$} and {\small$\text{CID}_\text{original net}$} denote the CID value of the compressed VGG-16 and the original VGG-16, respectively. We computed the change of CID during the compression as {\small$\Delta\text{CID}=\text{CID}_\text{compressed net}-\text{CID}_\text{original net}$}. On the other hand, we also compared the reconstruction capacity and the concentration of the compressed VGG-16 and the original VGG-16. Similar to {\small$\Delta \text{CID}$}, the change of RU and concentration is computed as {\small$\Delta\text{RU}=\text{RU}_\text{compressed net}-\text{RU}_\text{original net}$} and {\small$\Delta\text{concentration}=\text{concentration}_\text{compressed net}-\text{concentration}_\text{original net}$}, respectively.

Based on Figure~\ref{fig:CompressAndDistill} (a), we found that \textbf{network compression decreased the CID of features, which indicated that compressed DNNs were more sensitive to adversarial noises.} It was because the CID value could indicate the robustness to the adversarial noise (please see Appendix~\ref{appsubsec:advnoise} for more details and results
). Besides, Figure~\ref{fig:CompressAndDistill} shows that \textbf{network compression did not significantly affect the reconstruction capacity and the concentration of intermediate-layer features.}
It meant that \textbf{the network compression made the DNN less powerful to remove the information of redundant pixels, but it still maintained the representation power of the DNN, \emph{i.e.} the feature could still well reconstruct the input. On the other hand, the feature still concentrated on the foreground.}

\textbf{Analysis of knowledge distillation:} We used our metrics to analyze the inner mechanism of knowledge distillation. We trained the VGG-16, ResNet-18, and ResNet-34 using the CUB200-2011 dataset~\cite{CUB200} as three teacher nets for fine-grained classification. Each teacher net was used to guide the learning of an AlexNet. Figure~\ref{fig:CompressAndDistill} (b) compares layerwise information discarding between AlexNets learned with and without knowledge distillation. We found that AlexNets learned using knowledge distillation had lower information discarding than the ordinarily learned AlexNet.
Therefore, we can conclude that knowledge distillation helped AlexNets to preserve more information. Meanwhile, knowledge distillation may make intermediate-layer features more sensitive to noises, because AlexNets were mainly learned from distillation and used less noisy information from real training data during the distillation process.

\textbf{Further experiments:}
In Appendix~\ref{appsec:exp}, we used the proposed metrics to \textit{analyze flaws of the network architecture}, and \textit{explored the relationship between the CID value and the adversarial noises}. \textit{Furthermore, we found that the adversarial trained DNNs discarded more information than the normally trained DNNs. Besides, the adversarial trained DNNs more focused on the foreground than the normally trained DNNs.}


\section{Conclusion}
In this paper, we have defined three metrics to quantify information discarding during the forward propagation. A model-agnostic method is developed to measure the proposed metrics for each specific layer of a DNN. Comparing existing methods of visualizing network features and extracting important pixels, our metrics provide consistent and faithful results across different layers. Therefore, our metrics enable a fair analysis of the efficiency of signal processing of DNNs. The concentration value is highly correlated with the performance of the DNN. In experiments, we have used our metrics to analyze and understand the inner mechanisms of existing deep-learning techniques.

\textbf{Acknowledgments} This work is partially supported by National Key R\&D Program of China (2021ZD0111602), the National Nature Science Foundation of China (No. 61906120, U19B2043), Shanghai Natural Science Foundation (21JC1403800, 21ZR1434600), Shanghai Municipal Science and Technology Major Project (2021SHZDZX0102). This work is also partially supported by Huawei Technologies Inc.


\bibliography{TheBib}
\bibliographystyle{icml2022}

\newpage
\appendix
\onecolumn
\section{Relationship between Equation (1) and Equation (2)}
\label{appsec:eqn2}
We introduce the computation of the entropy of the input in Equation (1) of the paper, and the computation of the CID value in Equation (2) of the paper. In this section, we introduce how to approximate Equation (1) using Equation (2). For the convenience of readers, we rewrite these equations as follows.
{\small
\begin{equation*}
  H(X_\text{obj})\!=\!-\!\sum\nolimits_{x'}p(x')\log p(x')\;\textrm{ s.t. }\; \Vert f'-f\Vert^2\le\epsilon
\end{equation*}
\begin{equation*}
  H(X_\text{obj})=\sum\nolimits_{i=1}^{n}\!\!H_i(\sigma_i),\;
  \textrm{s.t.}\;\text{Prob}(\Vert f'-f\Vert^2\le\epsilon)=1\!-\!\tau
\end{equation*}
}
$x'\in X_\text{obj}$ is a perturbed input around the original input $x$, \emph{i.e.} $x'=x+\Delta x$. We assume that $x'$ follows the Gaussian distribution $\mathcal{N}(\mu,\Sigma)$. However, this assumption cannot ensure $\Vert f'-f\Vert^2\le\epsilon$, because there is a small probability that $x'$ is significantly different from $x$. Therefore, we relax the constraint as $\text{Prob}(\Vert f'-f\Vert^2)\ge 1-\tau$, where $\tau$ is a small positive number.
We simplify the covariance matrix as $\Sigma=\text{diag}[\sigma_1^2,...,\sigma_n^2]$, and we can approximate $H(X_\text{obj})$ as $\sum_{i=1}^nH_i(\sigma_i)$. In this way, we can approximate Equation (1) using Equation (2).

\section{The derivation of Equation (4) in the paper}
\label{appsec:eqn4}
In the paper, we introduce the computation of the metric CID in the ``For fair layer-wise comparisons'' paragraph of Section~\ref{sec:SidAndConcentration}. We learn the {\small$\boldsymbol\sigma$} to compute the CID value using Equation (3). In this section, we introduce how to get Equation (4) to ensure the fairness of the layer-wise comparison. For the convenience of readers, we rewrite Equation~\eqref{eqn:ign} as follows.

\begin{equation*}
\begin{aligned}
Loss({\boldsymbol\sigma})&=\frac{1}{\delta_f^2}\underset{f'}{\mathbb{E}}\left[\Vert f'-f\Vert^2\right]-\lambda\sum\nolimits_{i=1}^{n}H_i(\boldsymbol\sigma),
\label{appeqn:ign}
\end{aligned}
\end{equation*}
where {\small${\boldsymbol\sigma}=[\sigma_1,\ldots,\sigma_{n}]^{\top}$} is the parameter that we aim to learn. {\small$\lambda$} is a positive scalar, and {\small$\delta_f^2=\lim_{\xi\rightarrow 0^+}\mathbb{E}_{x'\sim \mathcal{N}(x,\xi^2\bf{I})}[\|h(x')-f\|^2]/\xi^2$} is the inherent variance of intermediate-layer features, which is used for normalization.

In order to ensure fair layer-wise comparisons, we need to control the value range of the first term in Equation (3). Features of different layers need to be perturbed at a comparable level.
Therefore, for each layer, we measure and compare {\small$H_i(\sigma_i)$} when {\small$\Vert f'-f\Vert^2\le\epsilon=\alpha\delta_f^2$}, where {\small$\alpha$} is a positive scalar.
In this way, the value of {\small$\lambda$} need to be adjusted (manually or automatically) to make {\small${\boldsymbol\sigma}$} satisfy {\small$\text{Prob}(\Vert f'-f\Vert^2\le\alpha\delta_f^2)>1-\tau$}.

To simplify the implementation, we make the approximation {\small$\mathbb{E}_{\delta\sim\mathcal{N}(0,I)}[\Vert f'-f\Vert^2]\approx\beta\sigma_f^2$}, where {\small$\beta<\alpha$}, as a replacement of {\small$\text{Prob}(\Vert f'-f\Vert^2\le\alpha\delta_f^2)>1-\tau$}.
In this way, {\small$\lambda$} is determined based on the value of {\small$\beta$}, and we do not need to specify values of {\small$\tau$} and {\small$\alpha$}. \emph{I.e.} we only need to consider the value of {\small$\beta$} to learn {\small${\boldsymbol\sigma}$}.
Thus, the stop criterion of learning {\small${\boldsymbol\sigma}$} can be given as Equation (4) in the paper, \emph{i.e.}
\begin{equation*}
\label{appeqn:lambda}
\min_\sigma\text{Loss}(\boldsymbol\sigma)
\;\text{s.t.}\;\mathbb{E}_{\delta\sim\mathcal{N}(0,I)}[\Vert f'-f\Vert^2]\approx\beta\sigma_f^2,\beta<\alpha.
\end{equation*}

\section{Proof of the relationship between the concentration and the information bottleneck theory}
\label{appsec:information-bottleneck}
We introduce the relationship between the concentration and the information bottleneck theory in the last paragraph of Section~\ref{sec:SidAndConcentration}. In this section, we prove the above relationship, \emph{i.e.} a high concentration usually indicates a high efficiency $\rho$. According to the information bottleneck theory~\cite{InformationBottleneck, InformationBottleneck2}, the efficiency $\rho$ of a DNN can be computed as $\rho=I(F;Y)/I(X;F)$, where $X$, $F$, $Y$ denote input samples, intermediate-layer features and gound-truth labels, respectively. The efficiency can be formulated as follows.
\begin{equation}
\label{eqn:A1}
\begin{aligned}
\rho&=\frac{I(F;Y)}{I(X;F)}\\
&=\frac{I(X;F;Y)+I(F;Y|X)}{H(X)-H(X|F)}\\
&=\frac{I(X;Y)-I(X;Y|F)+I(F;Y|X)}{H(X)-H(X|F)}\\
&=\frac{H(X)-H(X|Y)-I(X;Y|F)+I(F;Y|X)}{H(X)-H(X|F)}\\
&=\frac{H(X)-H(X|Y)-I(X;Y|F)+I(F;Y|X)}{H(X)-\text{CID}}
\end{aligned}
\end{equation}
Note that the intermediate-layer feature $F$ is determined by $X$, thereby $I(F;Y|X)=0$. In this way, the above equation can be rewritten as
\begin{equation}
\label{eqn:A2}
\rho=\frac{H(X)-H(X|Y)-I(X;Y|F)}{H(X)-\text{CID}}
\end{equation}
Since $H(X)$ and $H(X|Y)$ are only related to the dataset, we can consider them as constants.
Given $F$, we assume that the foreground of the input is conditionally independent with the background.
Thus, $I(X;Y|F)$ can be disentangled as $I(X;Y|F)=I(X_\text{fg};Y|F)+I(X_\text{bg};Y|F)$, where $X_\text{fg}$ and $X_\text{bg}$ denote the foreground and background part of the input, respectively. Specifically, we have
\begin{equation}
\label{eqn:A3}
\begin{aligned}
I(X_\text{fg};Y|F)&=H(X_\text{fg}|F)-H(X_\text{fg}|F,Y)=\gamma_\text{fg}H(X_\text{fg}|F)\\
I(X_\text{bg};Y|F)&=H(X_\text{bg}|F)-H(X_\text{bg}|F,Y)=\gamma_\text{bg}H(X_\text{bg}|F)
\end{aligned}
\end{equation}
We assume that there exists a scalar $\gamma_\text{fg}$ to represent the ratio of the foreground information, which is related to the ground-truth label $Y$, \emph{i.e.} $I(X_\text{fg};Y|F)=\gamma_\text{fg} H(X_\text{fg}|F)$.
Similarly, we assume that there exists a scalar $\gamma_\text{bg}$ to represent the ratio of the background information, which is related to the ground-truth label $Y$, \emph{i.e.} $I(X_\text{bg};Y|F)=\gamma_\text{bg} H(X_\text{bg}|F)$.
Since $H(X_\text{fg}|F)>H(X_\text{fg}|F,Y)$ and $H(X_\text{bg}|F)>H(X_\text{bg}|F,Y)$, we have $0<\gamma_\text{fg}<1$, $0<\gamma_\text{bg}<1$.
Since the task is mainly related to the foreground, the information discarded in the foreground is usually less than the information discarded in the background.
In this way, we have $\gamma_\text{fg}\gg \gamma_\text{bg}$, $\gamma_\text{fg}-\gamma_\text{bg}>0$.
Thus, the efficiency $\rho$ can be written as follows.
\begin{eqnarray}
\label{eqn:A4}
\rho&\!\!\!\!\!=\!\!\!\!\!&\frac{H(X)-H(X|Y)-\gamma_\text{fg}H(X_{fg}|F)-\gamma_\text{bg}H(X_{bg}|F)}{H(X)-\text{CID}}\\
\label{eqn:A5}
&\!\!\!\!\!=\!\!\!\!\!&
\frac{\gamma_\text{fg} \text{concentration} - (\gamma_\text{fg}+\gamma_\text{bg})H(X_\text{bg}|F)+H(X)-H(X|Y)}{H(X)-\text{CID}}\\
\label{eqn:A6}
&\!\!\!\!\!=\!\!\!\!\!& -
\frac{\gamma_\text{bg} \text{concentration} + (\gamma_\text{fg}+\gamma_\text{bg})H(X_\text{fg}|F)-H(X)+H(X|Y)}{H(X)-\text{CID}}
\end{eqnarray}
By combining above two equations, we have
\begin{equation}
\begin{aligned}
\rho&=
\frac{(\gamma_\text{fg}-\gamma_\text{bg})\text{concentration}-(\gamma_\text{fg}+\gamma_\text{bg})\text{CID}+2(H(X)-H(X|Y))}{2[H(X)-\text{CID}]}\\
&=\frac{\gamma_\text{fg}+\gamma_\text{bg}}{2}+\frac{(\gamma_\text{fg}-\gamma_\text{bg})\text{concentration}-(\gamma_\text{fg}+\gamma_\text{bg}-2)H(X)-2H(X|Y)}{2[H(X)-\text{CID}]}
\end{aligned}
\end{equation}
Note that $\gamma_\text{fg}$, $\gamma_\text{bg}$, $H(X)$ and $H(X|Y)$ can be considered as constants. \
For simplicity, let $C_1=\frac{\gamma_\text{fg}+\gamma_\text{bg}}{2}$, $C_2=\gamma_\text{fg}-\gamma_\text{bg}>0$, $C_3=(\gamma_\text{fg}+\gamma_\text{bg}-2)H(X)+2H(X|Y)$, $C_4=H(X)>\text{CID}$. Therefore, we have
\begin{equation}
\rho=C_1+\frac{C_2 \text{concentration}-C_3}{2[C_4-\text{CID}]}
\end{equation}
Thus, for DNNs learned for the same task with the similar value of CID, a high value of $\text{concentration}$ usually indicates a high value of efficiency $\rho$, which reflects the connection between our metrics and the information bottleneck theory.

\section{About how to understand the limitation of CID from the perspective of invertible nets}
\label{appsec:invertible}
In this section, we discuss the limitation of CID in invertible nets, which is briefly introduced in the third paragraph of Section~\ref{subsec:discussions}.

Strictly speaking, there is no strict way to quantify the discarding of the input information during the computation of an intermediate-layer feature. Our method is based on the assumption that the concept of a specific object instance is within the range of {\small$\text{Prob}(\Vert f'-f\Vert^2\le\epsilon)\ge 1-\tau$}, which makes the algorithm sensitive to the activation magnitude of each feature dimension. For example, a typical failure case for this assumption is invertible neural networks~\cite{invertible,invertible2,invertible3,invertible4,invertible5,invertible6}. Theoretically, invertible neural networks do not discard any input information; otherwise, the input cannot be inverted from intermediate-layer features. Instead, invertible neural networks usually significantly decrease the magnitude of neural activations caused by unimportant pixels \emph{w.r.t.} the task, and boost the magnitude of neural activations triggered by important pixels \emph{w.r.t.} the task. Similarly, given a pre-trained DNN, if we revise a DNN by selectively halving magnitudes of parameters of 50\% filters $w\leftarrow0.5w$, theoretically, this revision does not discard any input information.

However, information discarding in this paper is defined from another perspective, \emph{i.e.} whether the input information can significantly contribute to the final output of the neural network. For both invertible neural networks and the above revision of halving magnitudes of parameters, these techniques all decrease activation magnitudes caused by certain pixels, thereby letting these pixels contribute less numerical values to the network output.

Therefore, our definition of information discarding does not conflict with the information processing in invertible neural networks. Based on our definition of information discarding, a high information discarding of a pixel indicates that this pixel will contribute a low numerical score to the intermediate-layer feature or the network output.

\section{About the computational cost of CID and RU}
\label{appsec:cost}
In the last paragraph of Section~\ref{subsec:discussions}, we have briefly clarified that the computational cost of the CID and RU is comparable with previous explanation methods. In this section, we will provide more discussions about this issue.
In this paper, the pixel-wise CID and RU were usually generated by letting the DNN recursively conduct 100 inferences. In comparison, IG~\cite{sundararajan2017axiomatic} took 300 inferences to compute the attribution map. LIME~\cite{trust} needed 5000 inferences to learn the explanation result. The computational cost of the Shapley value~\cite{shapley1953value} was NP-hard. All of these explanation methods had a higher computational cost than the proposed metrics. Therefore, the

\section{Further experiments}
\label{appsec:exp}
This section introduces several additional experiments, which are briefly introduced in the last paragraph of the "Comparative studies" section in the paper.

\subsection{Diagnosis of architectural revision (damage)}
\begin{figure*}[!ht]
  \centering
  \includegraphics[width=0.6\linewidth]{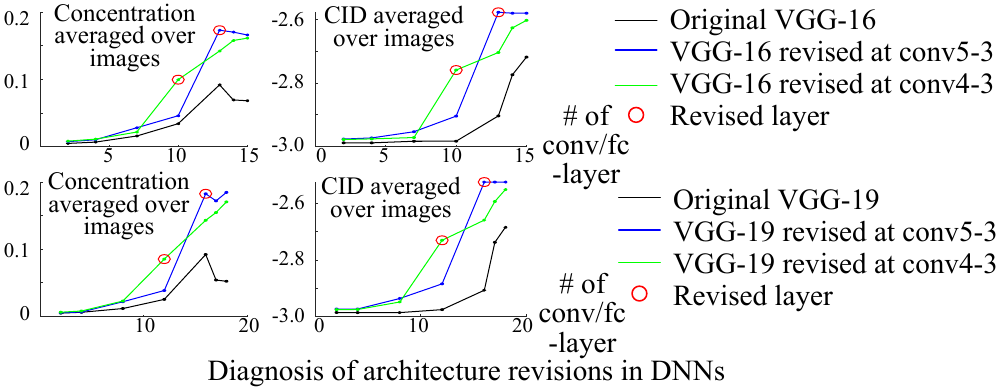}
  \caption{Diagnosis of architectural revision. Values were normalized by the pixel number per image and averaged over images. The architectural revision increased the value of CID and concentration.}
  \label{fig:Damage}
\end{figure*}

In this experiment, we aimed to analyze whether the proposed metrics reflected architectural revisions of DNNs. To this end, we revised the architecture of the VGG-16/VGG-19 network by changing a specific convolutional layer to contain four $7\!\!\times\!\!7\!\!\times\!\!512$ filters with $\text{padding}\!\!=\!\!3$, which hurts the representation capacity of the DNN.
We trained both the original VGG-16/VGG-19 and the revised VGG-16/VGG-19 for binary classification between bird images cropped from the CUB200-2011 dataset~\cite{CUB200} and random images in the ImageNet~\cite{ImageNet}. Figure~\ref{fig:Damage} compares the original and revised DNNs. We found that compared to the original DNN, the architectural revision significantly boosted the information discarding at the revised layer.
Meanwhile, the architectural revision (damage) also slightly increased the concentration of DNNs. The increase of the concentration seemed to conflict with the architectural damage, but this can be explained as follows.

1. Compared to the increase of the information discarding of the revised net, the increase of concentration was significantly lower. Thus, in general, the architectural revision hurt the representation capacity of the DNN.

2. The DNN with the reduced feature dimension could only encode much fewer concepts of object parts. Thus, the revised DNN usually encoded fewer, simpler, but more discriminative features than original DNNs.

3. Original DNNs usually ignored background information and extracted discriminative foreground features at high FC layers (see Figure~\ref{fig:Damage}, whereas the dimension reduction at the revised layer made the DNN ignored background information at much lower layers.

\begin{table}[!ht]
\caption{Relationship between the $\Delta$CID and the adversarial robustness. A higher CID value usually indicates a higher adversarial robustness.}
\label{tab:relat_adv}
\begin{center}
\resizebox{0.6\linewidth}{!}{
\begin{tabular}{|l|c|c|}
\hline
DNN & $\Delta$CID & adversarial robustness $\Vert\epsilon\Vert_2$\\
\hline
Original DNN (with 100\% parameters) & 0 & 0.00276\\
DNN with 38.0\% parameters & -0.003 & 0.00281\\
DNN with 12.8\% parameters & -0.005 & 0.00254\\
DNN with 8.0\% parameters & -0.033 & 0.00109\\
\hline
\end{tabular}
}
\end{center}
\vspace{5pt}
\end{table}


\subsection{Relationship to adversarial noises}
\label{appsubsec:advnoise}
We conducted an experiment to reveal the relationship between the CID value and the adversarial noise of the DNN. We trained a VGG-16 using the CUB200-2011 dataset~\cite{CUB200} for fine-grained classification. Then, the VGG-16 was compressed using the method of~\cite{NetCompress} with different pruning thresholds. Table~\ref{tab:relat_adv} compares the CID value of the last FC layer and the adversarial robustness of the DNN, when the DNN was compressed at different ratios. For each input image, we computed adversarial samples towards top-20 incorrect fine-grained categories with the highest probabilities. For fair comparisons, we added the adversarial noise until the adversarial attack just succeeded, \emph{i.e.} when the adversarial perturbation just pushed the sample to the decision boundary.
For each adversarial noise, we measured its L-2 norm values. The adversarial robustness was reported as the average L-2 norm over all images. We only measured the CID value of the last FC layer, because the CID value of the last layer most fit the logic of the final prediction.
Table~\ref{tab:relat_adv} shows that a higher CID value usually indicates a higher adversarial robustness. This indicated a close relationship between the CID value and the adversarial noise.

\subsection{Analysis of information discarding after different epochs during the learning process}
\begin{figure}[t]
  \centering
  \includegraphics[width=0.6\linewidth]{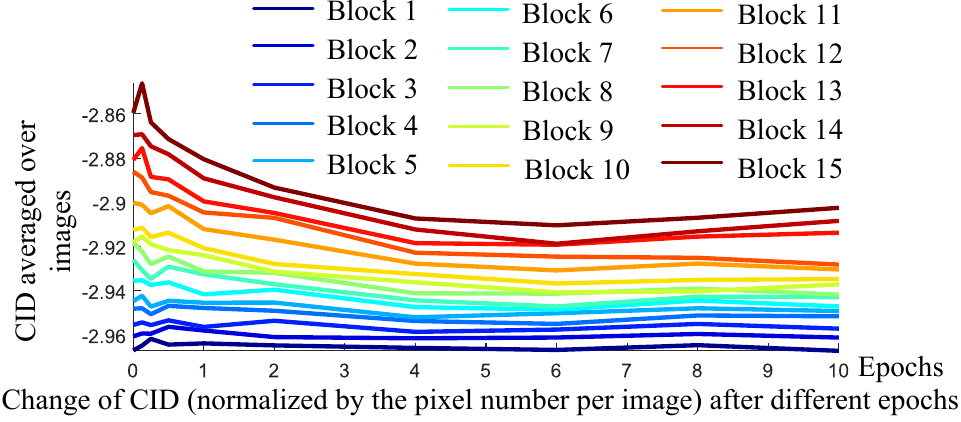}
  \caption{Effects of learning epochs. CID values were normalized by the pixel number per image and averaged over images. Each curve shows the information discarding of the output feature of a specific block during the learning process.}
  \label{fig:DamageAndEpoch}
\end{figure}

We trained the ResNet-32 network using the CIFAR-10 dataset~\cite{cifar}. Figure~\ref{fig:DamageAndEpoch} shows the change of information discarding \emph{w.r.t.} output features of different blocks during the learning process. Information discarding in high layers satisfied the information-bottleneck theory.

\subsection{Analysis of information discarding in the adversarial attack}
\begin{figure}[!ht]
\centering
\includegraphics[width=0.8\linewidth]{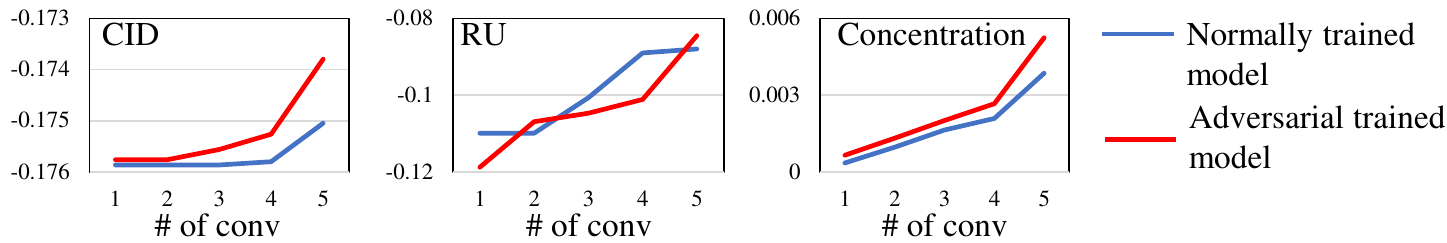}
\caption{Comparison of the CID, RU, and \textit{concentration} value between the adversarially trained AlexNet and the normally trained AlexNet. The adversarial trained AlexNet discarded more information than the normally trained AlexNet, and more focused on the foreground that the normally trained AlexNet.}
\label{fig:advtrain}
\end{figure}

In this experiment, we compared an adversarially trained AlexNet and a normally trained AlexNet on the CUB-200 2011 dataset. During the adversarial training, we used the PGD to generate adversarial samples. Specifically, we used the L-$\infty$ attack, where the number of steps of the attack is $10$, and the step size of the attack is 0.001. Figure~\ref{fig:advtrain} shows that the adversarial trained AlexNet discarded more information than the normally trained AlexNet, which was consistent with results in Appendix~\ref{appsubsec:advnoise}. Besides, the adversarial trained AlexNet more focused on the foreground than the normally trained AlexNet, since the adversarially trained AlexNet had higher \textit{concentration} values.


\section{Visualization of pixel-wise CID}
\label{appsec:visCID}

\subsection{For the U-Net learning using images in the ISBI cell tracking challenge}
In the second paragraph of Section~\ref{sec:comparative}, we have claimed that we applied the metric CID to the U-Net trained for segmenting neuronal structures in medical images as a real-world application. In this subsection, we will provide visualization results of the pixel-wise CID computed on different layers of the U-Net.
{\center
\noindent
\includegraphics[width=0.75\linewidth]{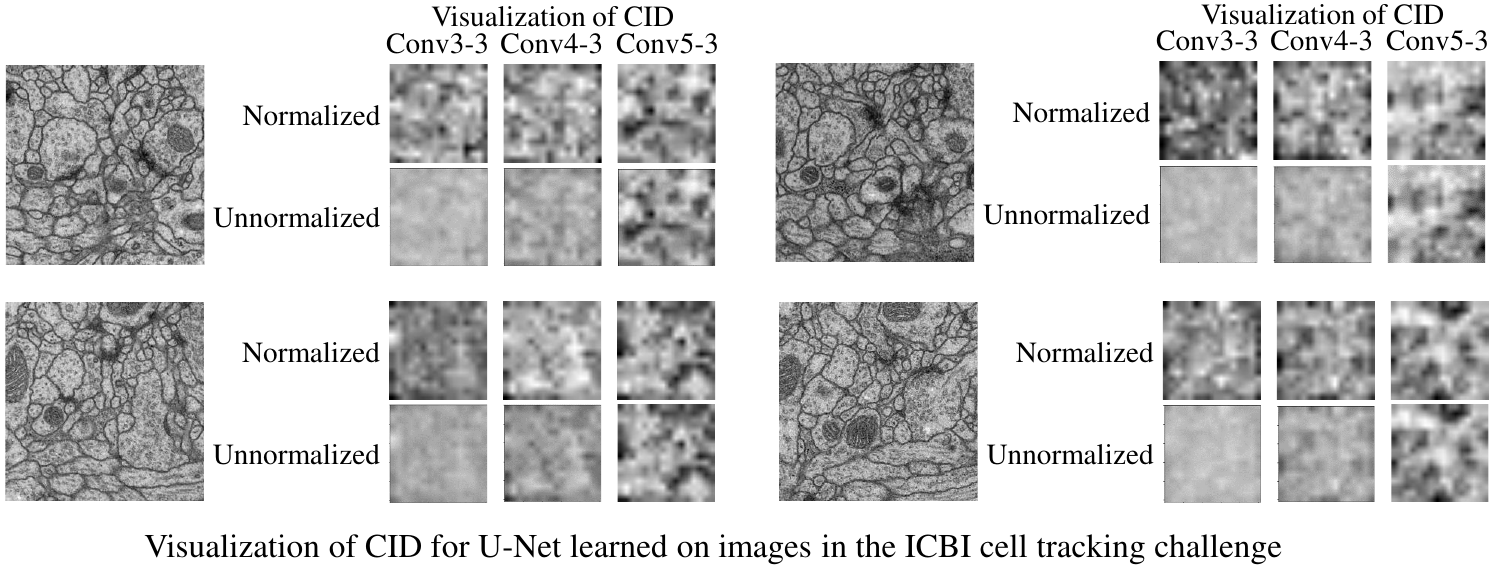}\\
}

\subsection{For the AlexNet learned using the ImageNet dataset}
This subsection provides visualization results of CID on the AlexNet learned using the ImageNet dataset. The visualized results can be used to fairly compare the relative importance of the foreground \emph{w.r.t.} the background over different layers.
{\center
\noindent
\includegraphics[width=0.75\linewidth]{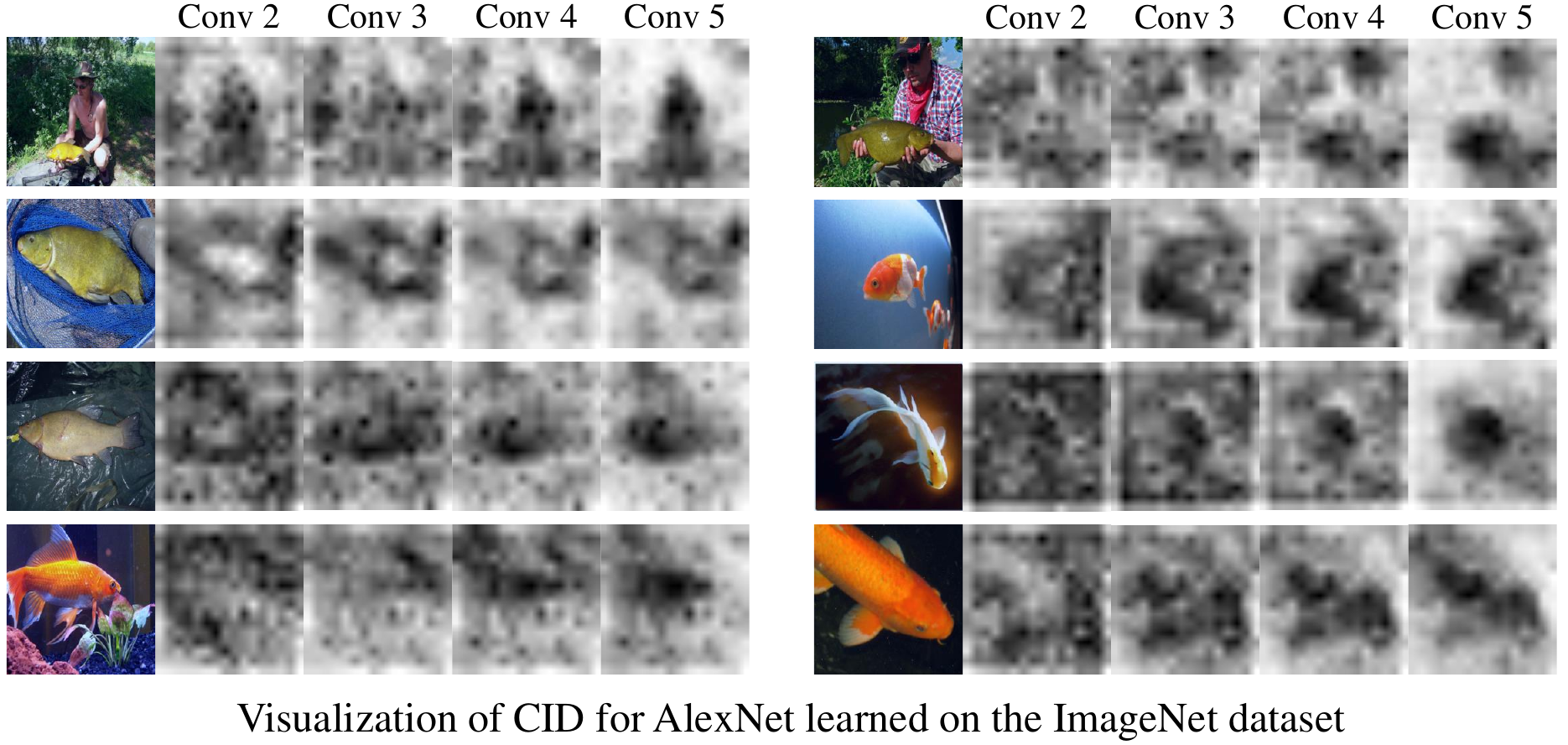}\\
}

\subsection{For the AlexNet/VGG-16/VGG-19 learned using the CUB200-2011 dataset}
This subsection provides visualization results of CID on the AlexNet/VGG-16/VGG-19 learned using the CUB200-2011 dataset. The visualized results can be used to fairly compare the relative importance of the foreground \emph{w.r.t.} the background over different layers.

{\center
\noindent
\includegraphics[width=0.75\linewidth]{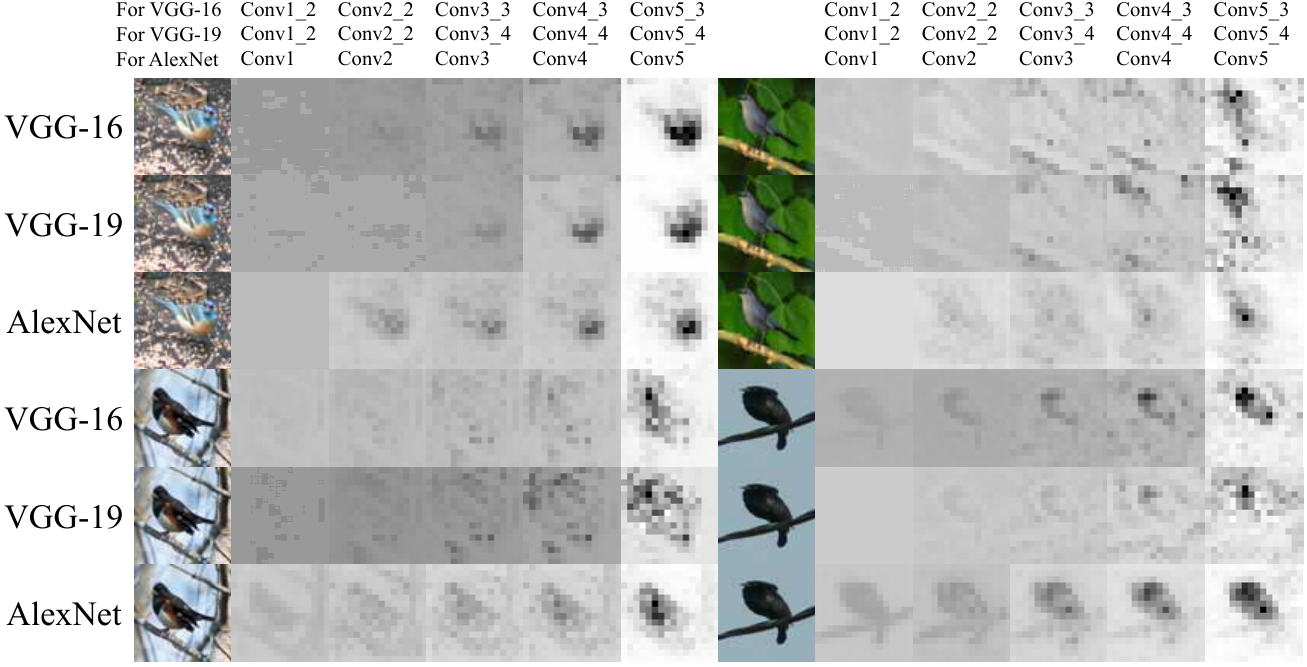}\\
\noindent
\includegraphics[width=0.75\linewidth]{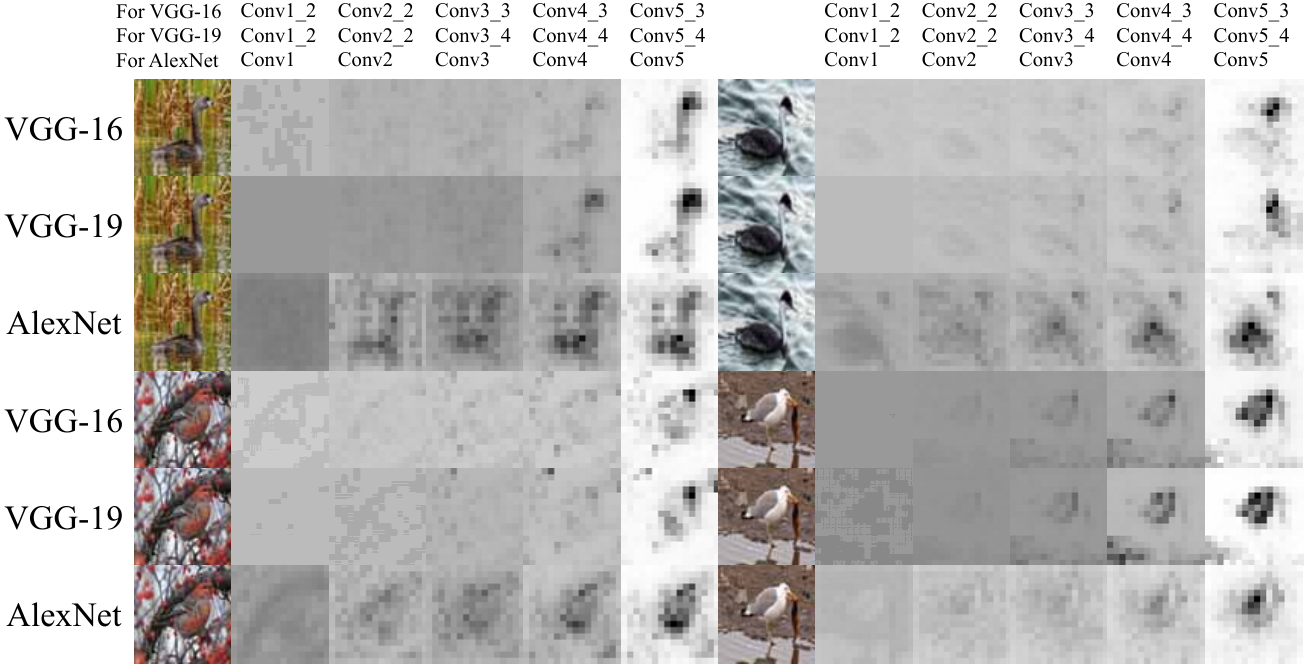}\\
\noindent
\includegraphics[width=0.75\linewidth]{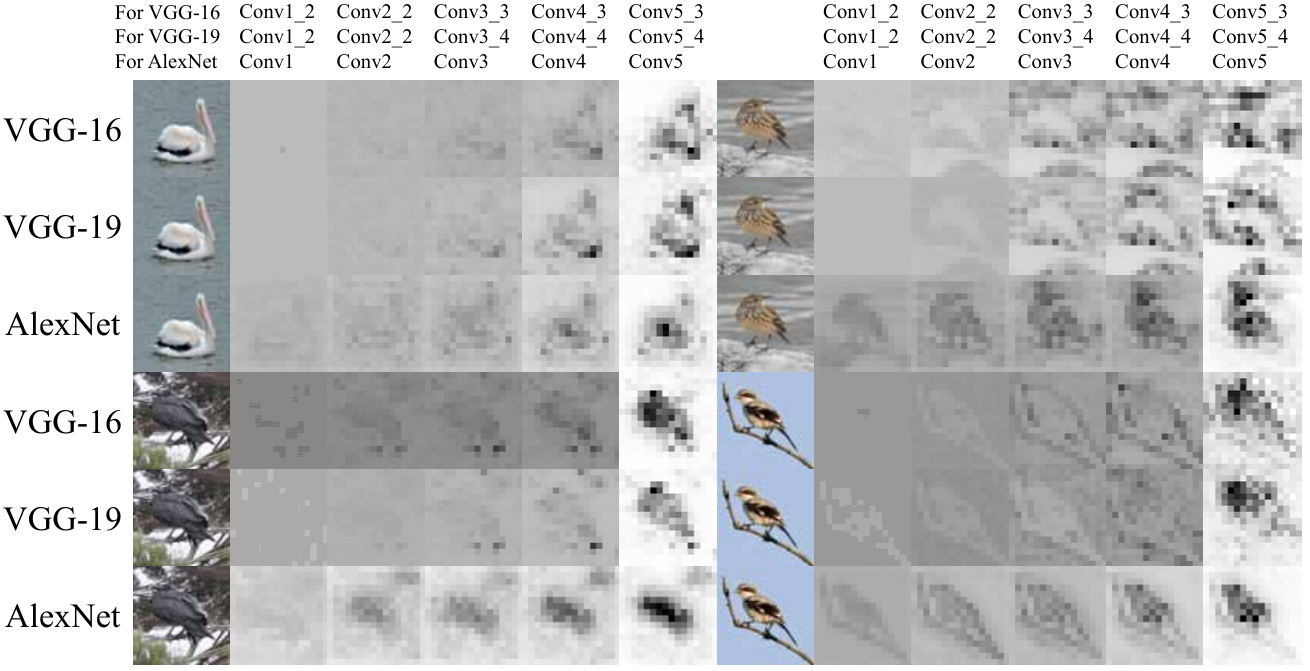}\\
\noindent
\includegraphics[width=0.75\linewidth]{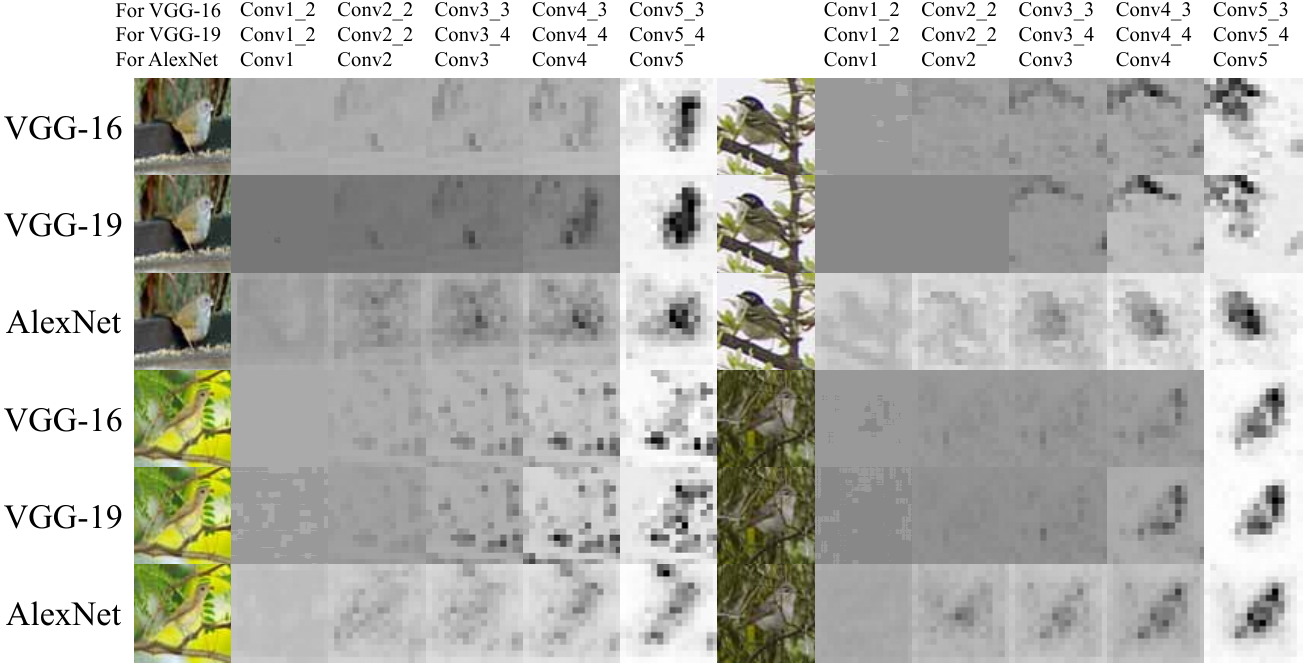}\\
\noindent
\includegraphics[width=0.75\linewidth]{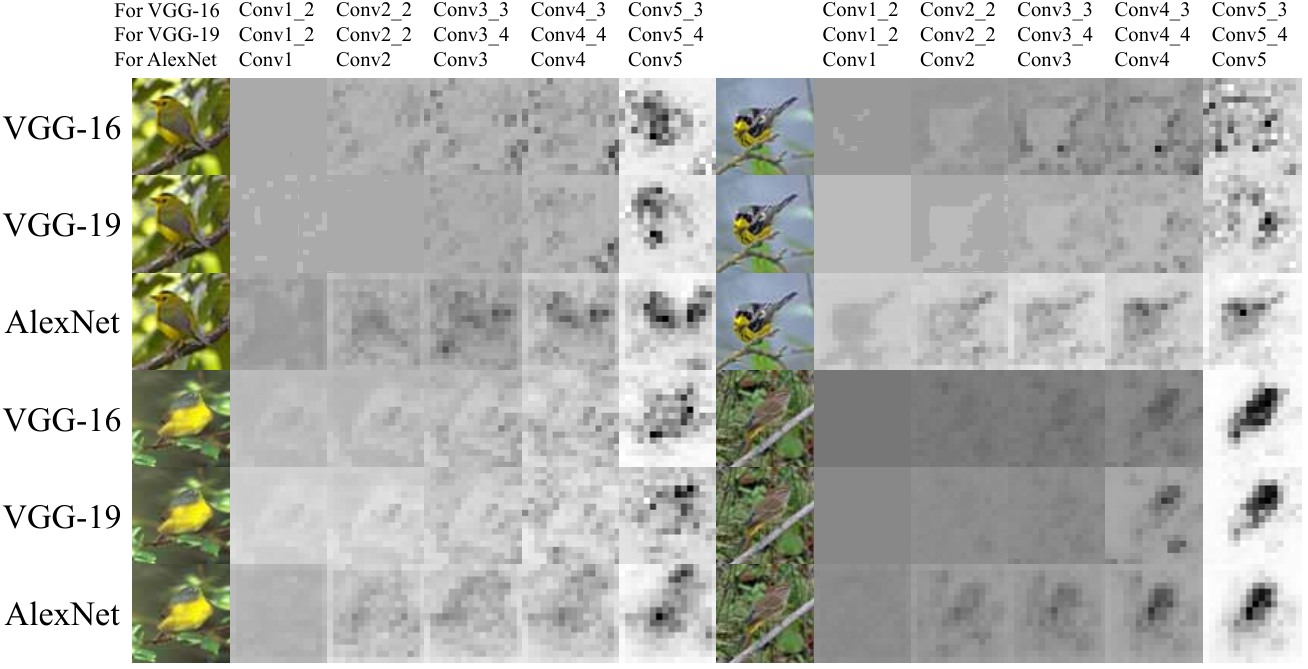}\\
\noindent
\includegraphics[width=0.75\linewidth]{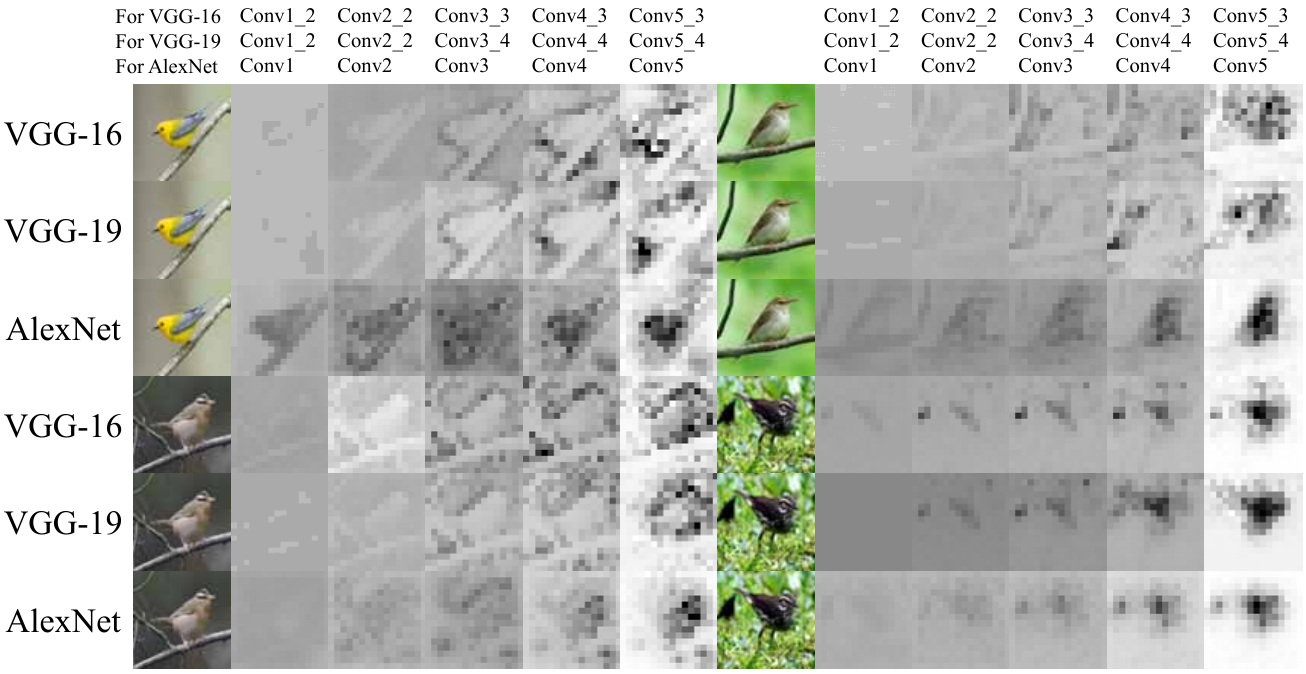}\\
}

\subsection{For the ResNet-20/32/44 learned using the CIFAR-10 dataset}
This subsection provides visualization results of CID on the ResNet-20/32/44 learned using the CIFAR-10 dataset. The visualized results can be used to fairly compare the relative importance of the foreground \emph{w.r.t.} the background over different layers.

{\centering
\noindent
\includegraphics[width=0.8\linewidth]{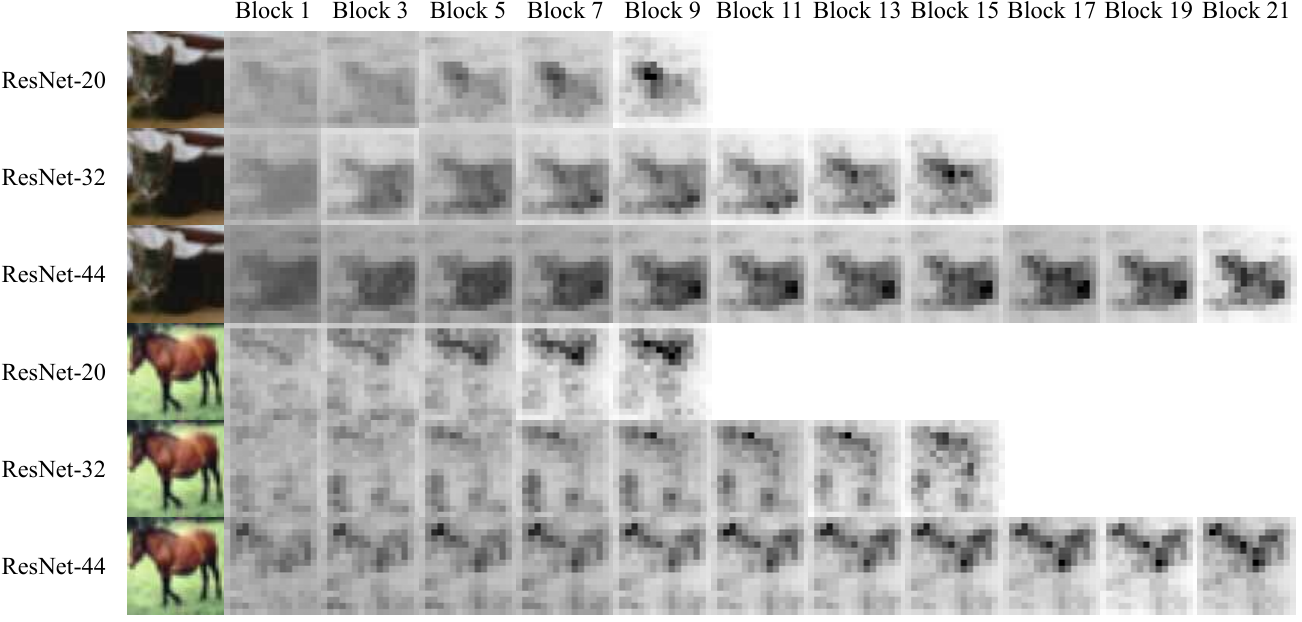}\\
\noindent
\includegraphics[width=0.8\linewidth]{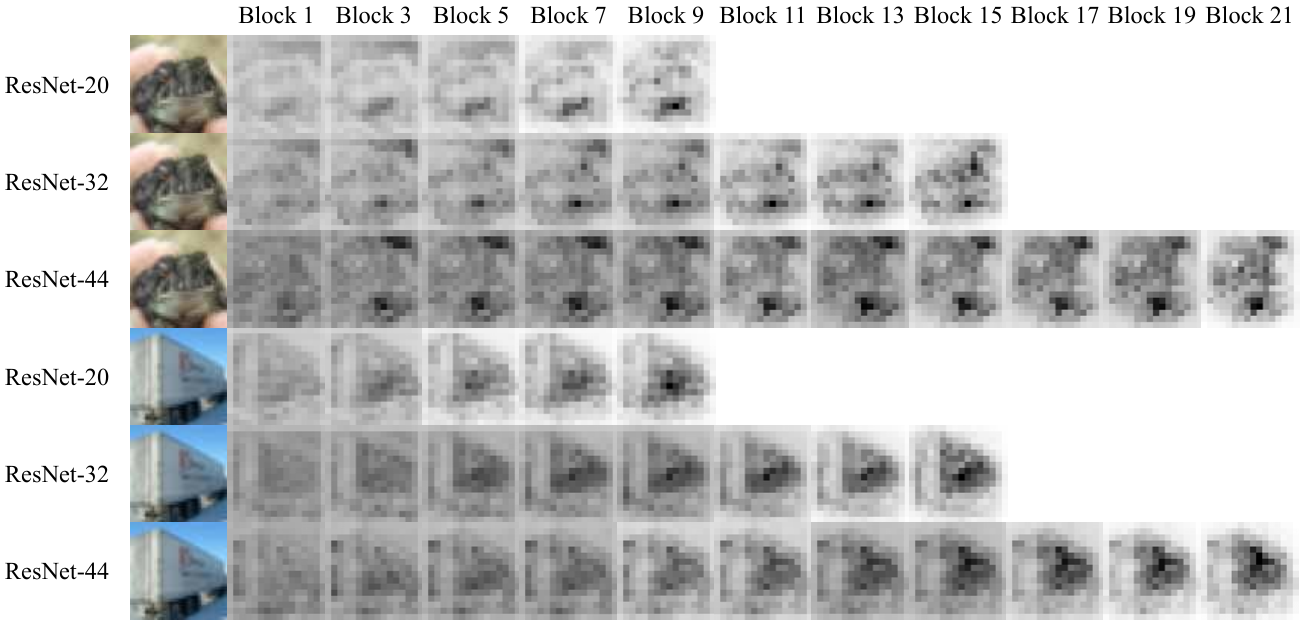}\\
\noindent
\includegraphics[width=0.8\linewidth]{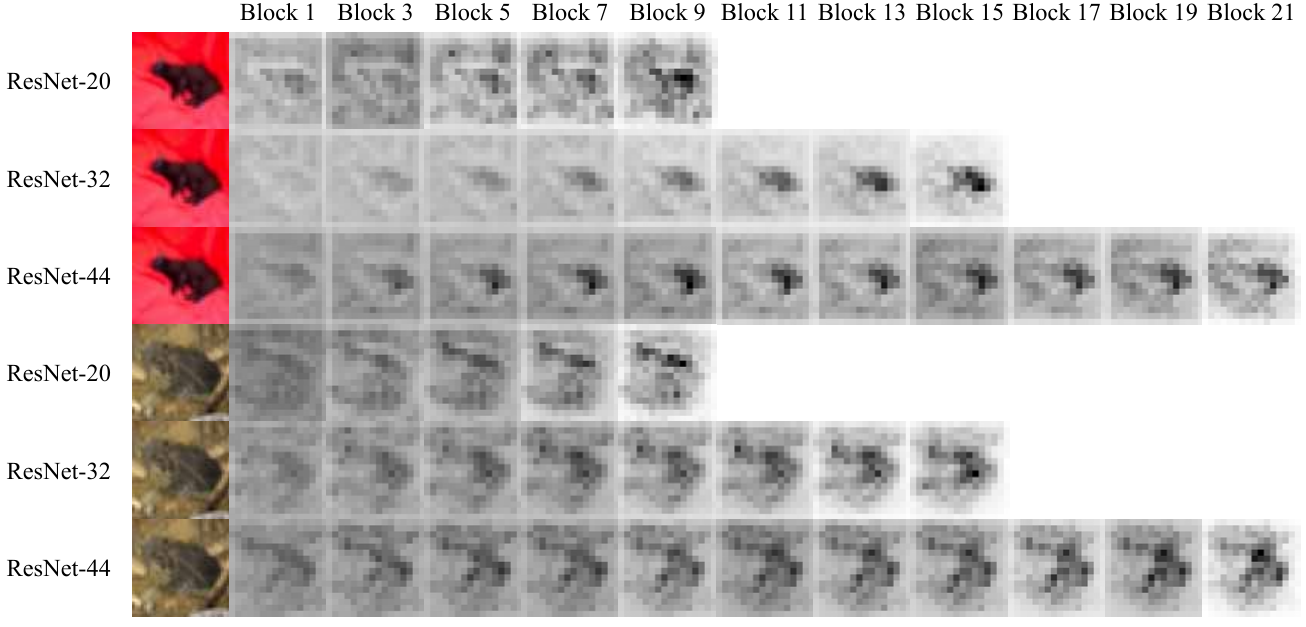}\\
\noindent
\includegraphics[width=0.8\linewidth]{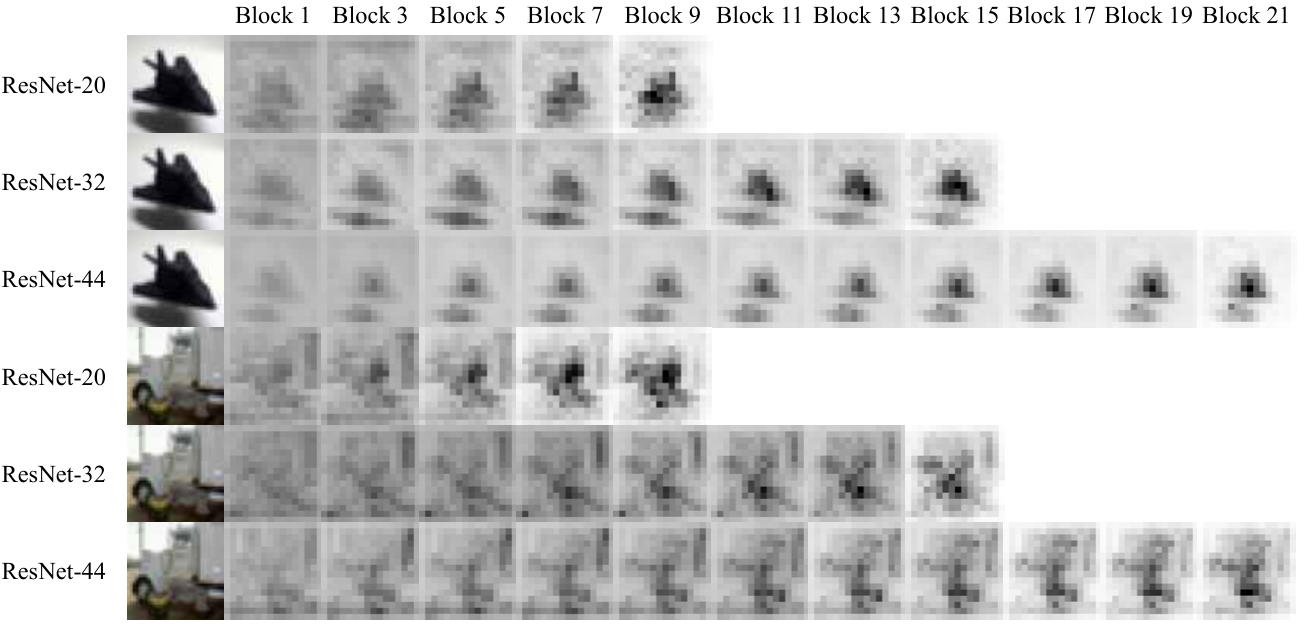}\\
\noindent
\includegraphics[width=0.8\linewidth]{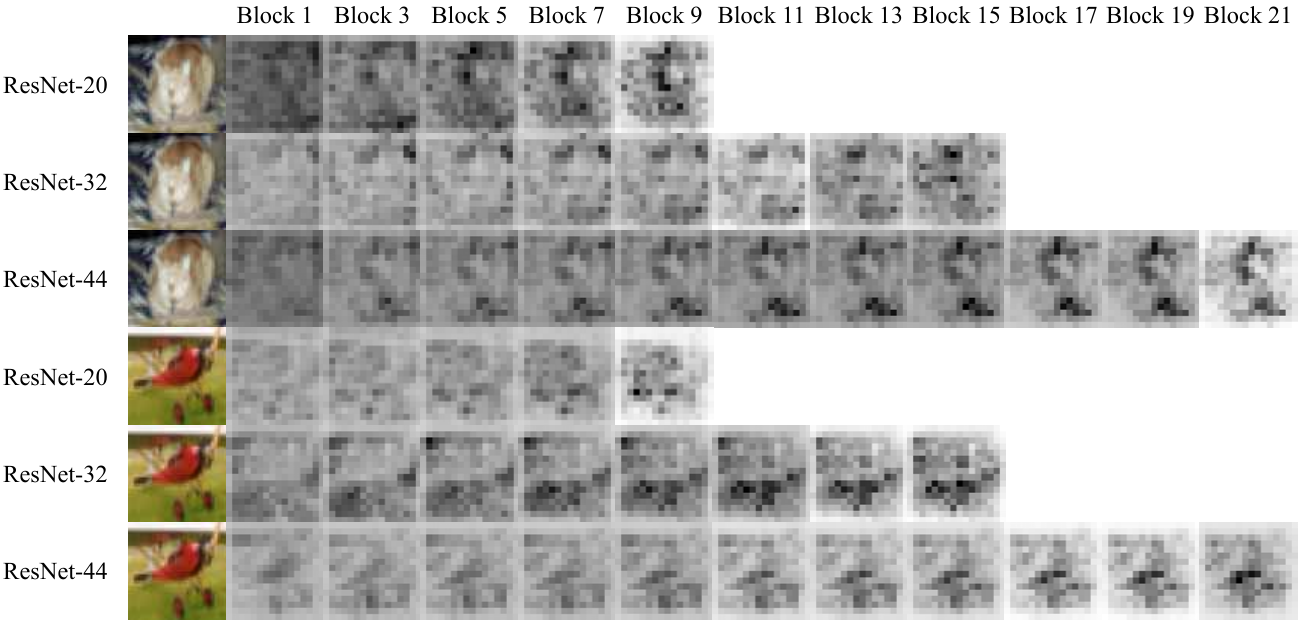}\\
\noindent
\includegraphics[width=0.8\linewidth]{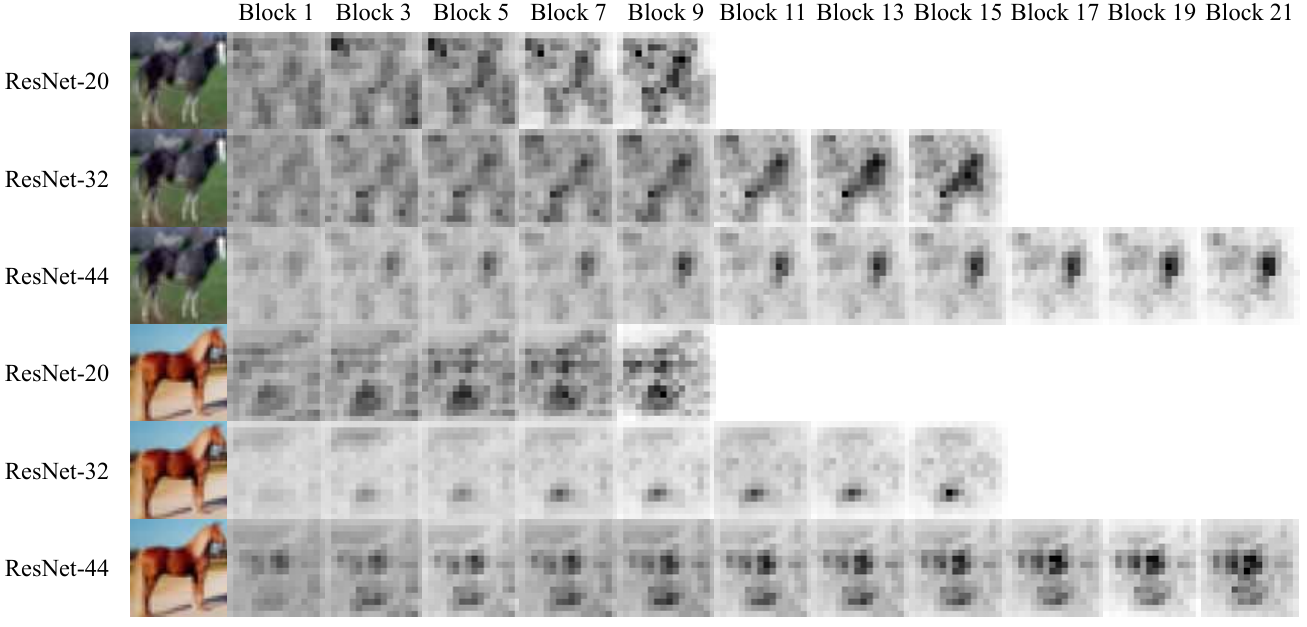}\\
\noindent
\includegraphics[width=0.8\linewidth]{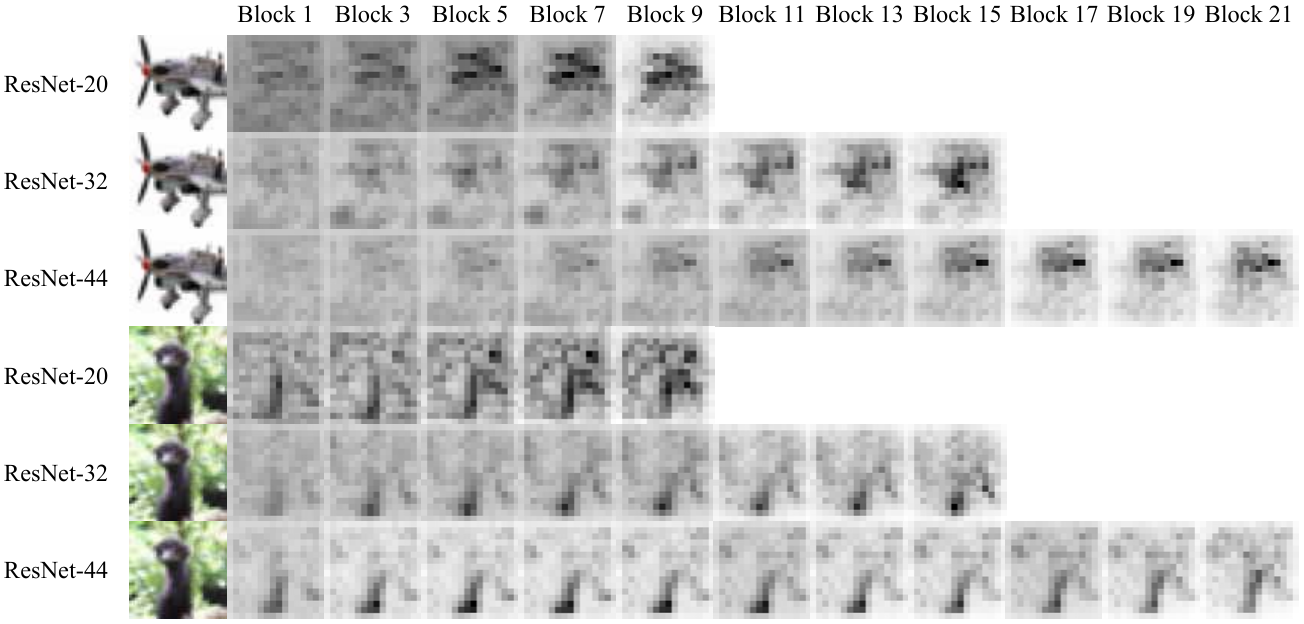}\\
}

\section{Visualization of pixel-wise RU}
\label{appsec:visRU}
\subsection{For the VGG-16 learned using the CUB200-2011 dataset}
This subsection provides visualization results of RU on the VGG-16 learned using the CUB200-2011 dataset. The visualized results can be used to fairly compare the relative importance of the foreground \emph{w.r.t.} the background over different layers.

{\centering
\noindent
\includegraphics[width=0.75\linewidth]{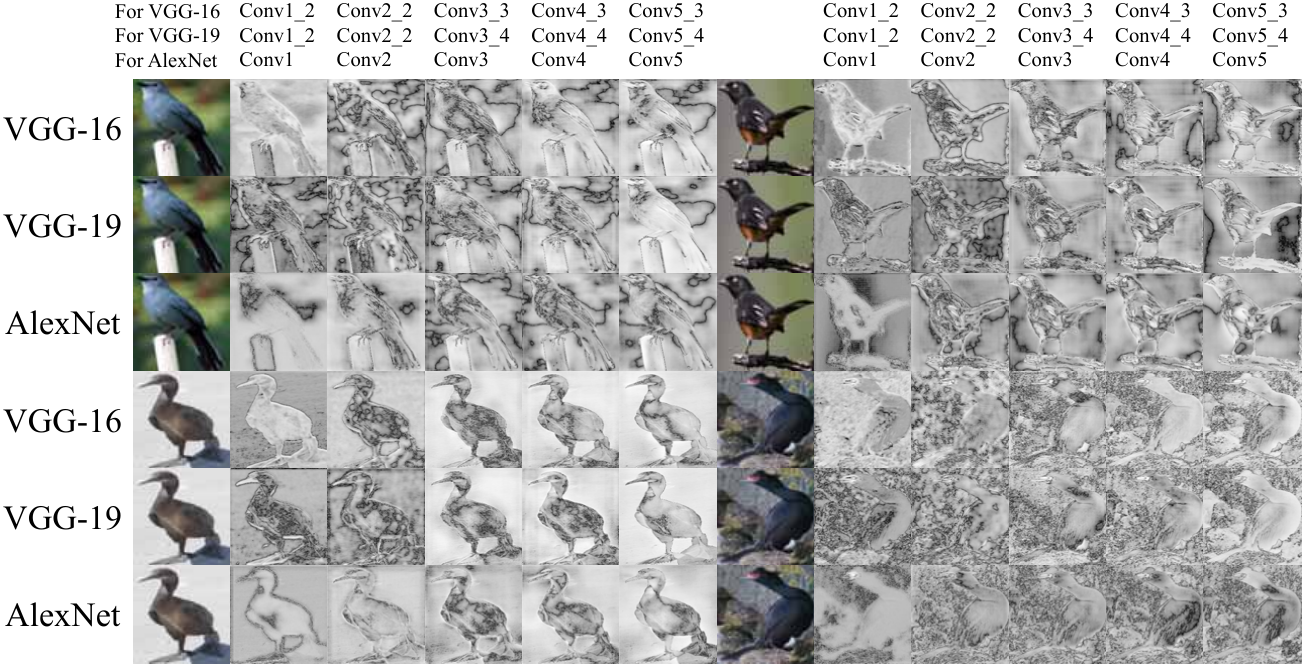}\\
\noindent
\includegraphics[width=0.75\linewidth]{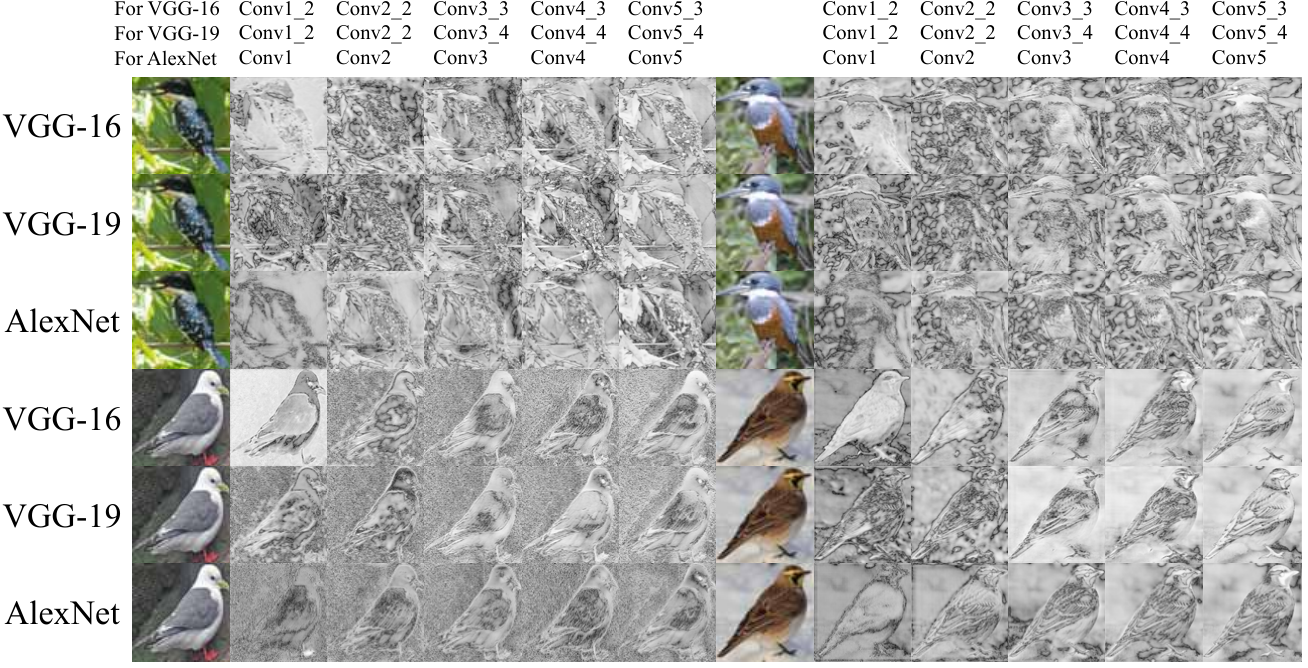}\\
\noindent
\includegraphics[width=0.75\linewidth]{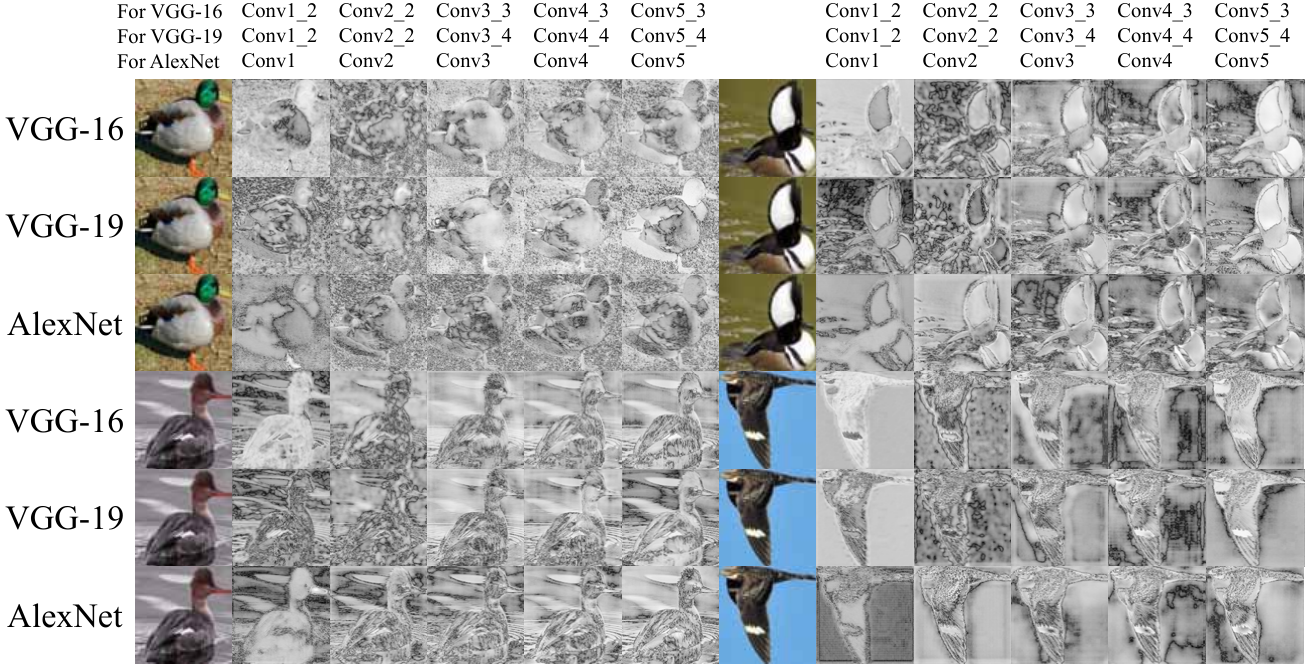}\\
\noindent
\includegraphics[width=0.75\linewidth]{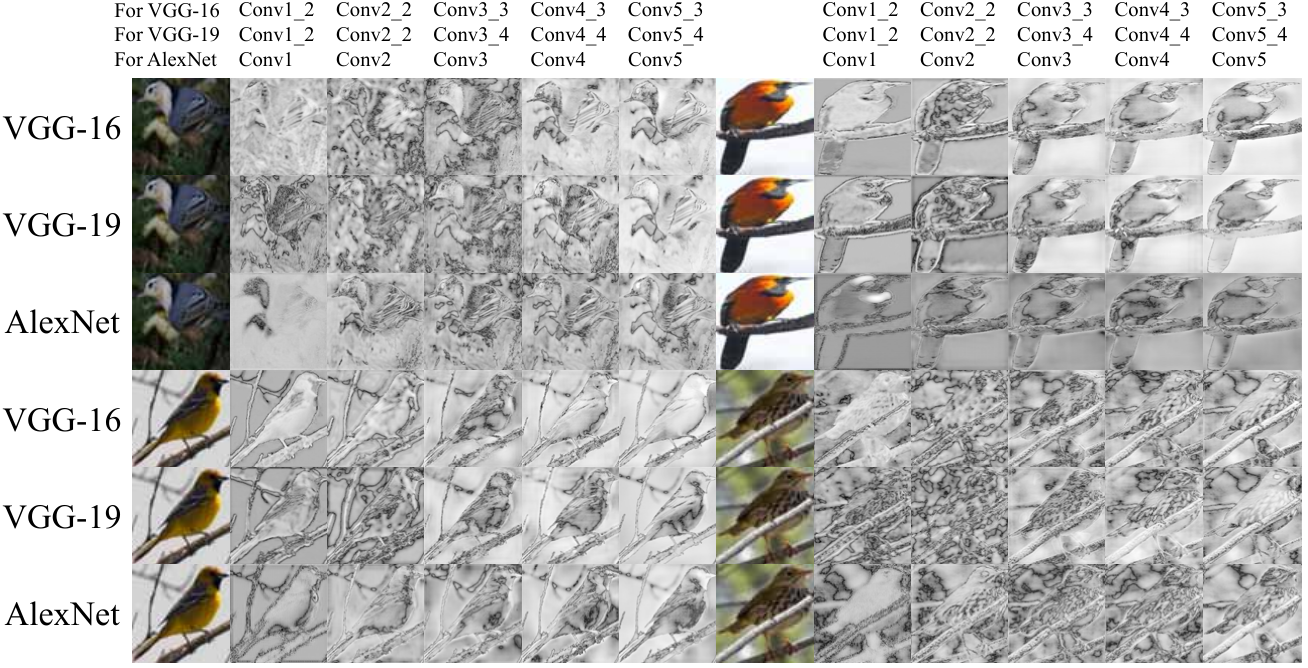}\\
\noindent
\includegraphics[width=0.75\linewidth]{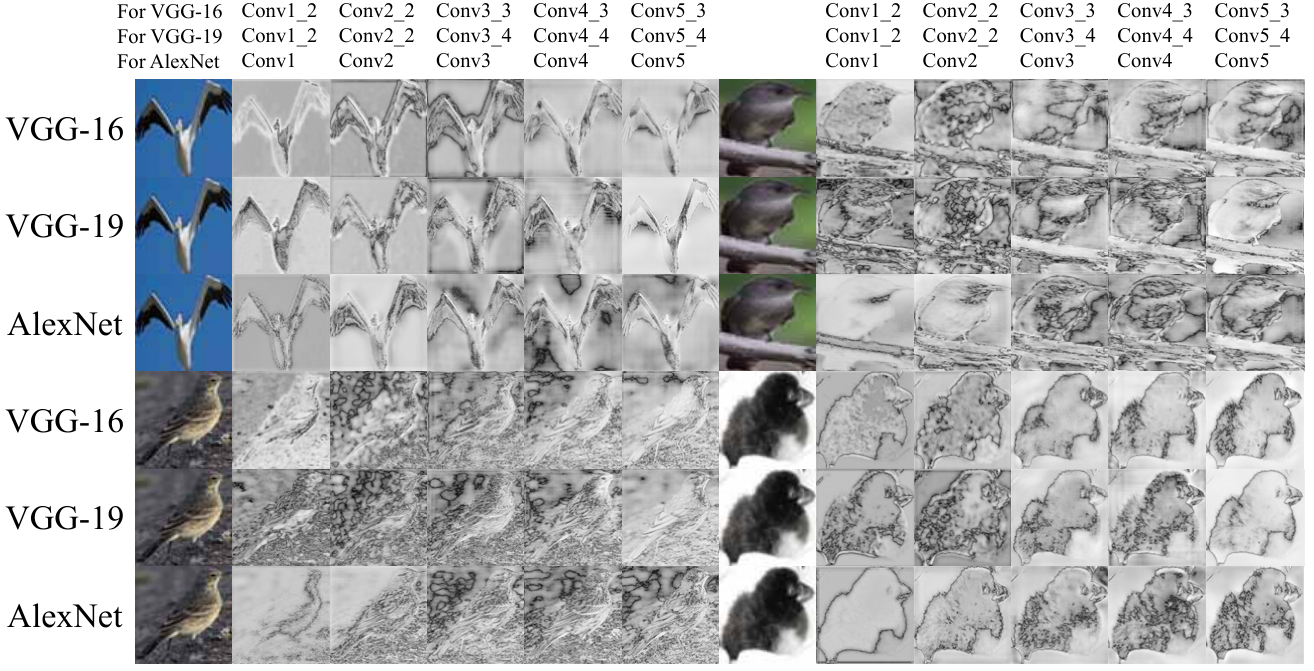}\\
\noindent
\includegraphics[width=0.75\linewidth]{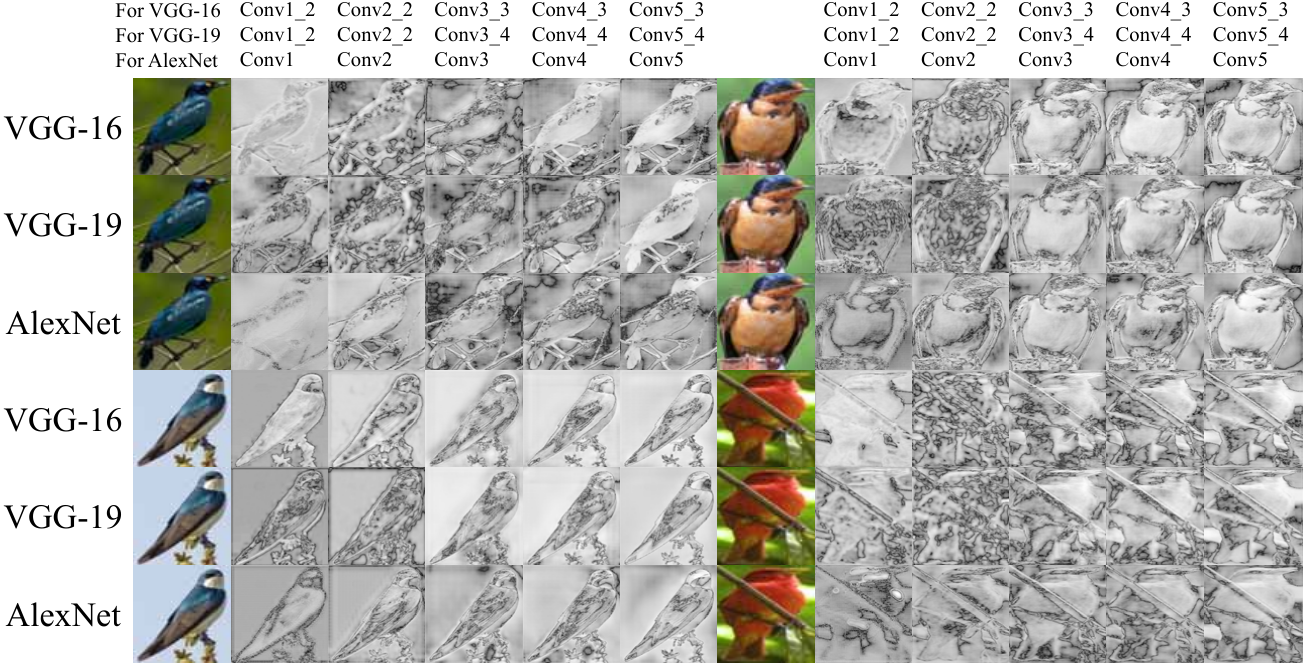}\\
}

\subsection{For the ResNet-20/32/44 learned using the CIFAR-10 dataset}
This subsection provides visualization results of RU on the ResNet-20/32/44 learned using the CIFAR-10 dataset. The visualized results can be used to fairly compare the relative importance of the foreground \emph{w.r.t.} the background over different layers.

{\centering
\noindent
\includegraphics[width=0.8\linewidth]{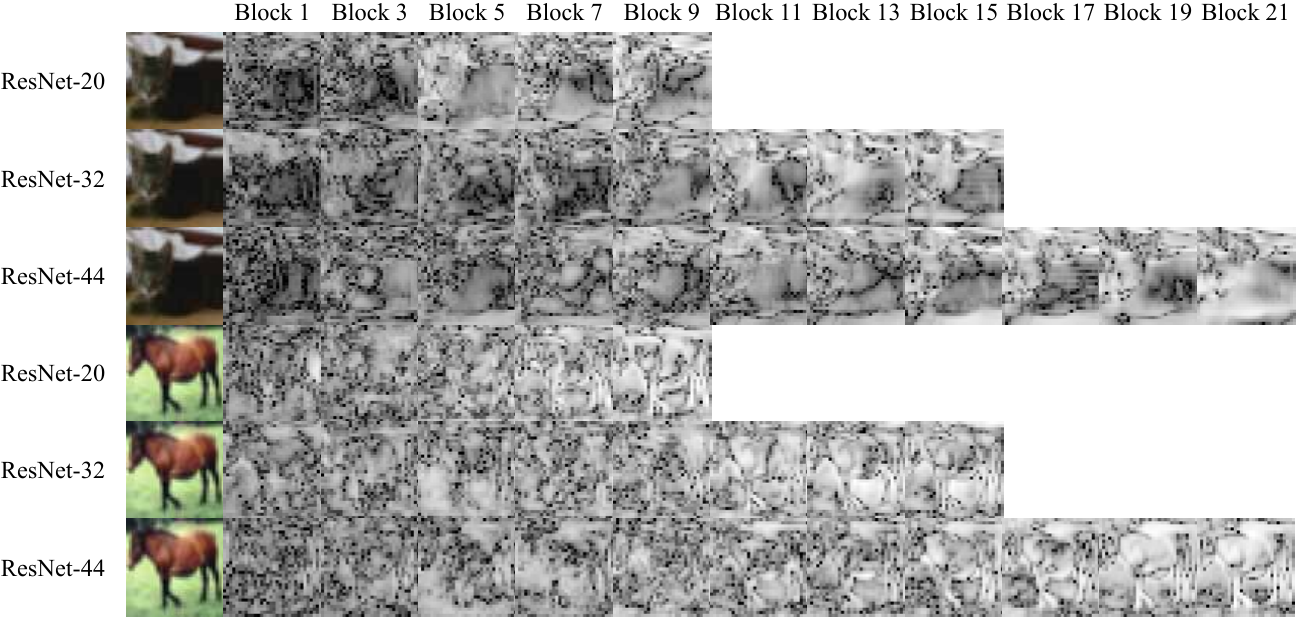}\\
\noindent
\includegraphics[width=0.8\linewidth]{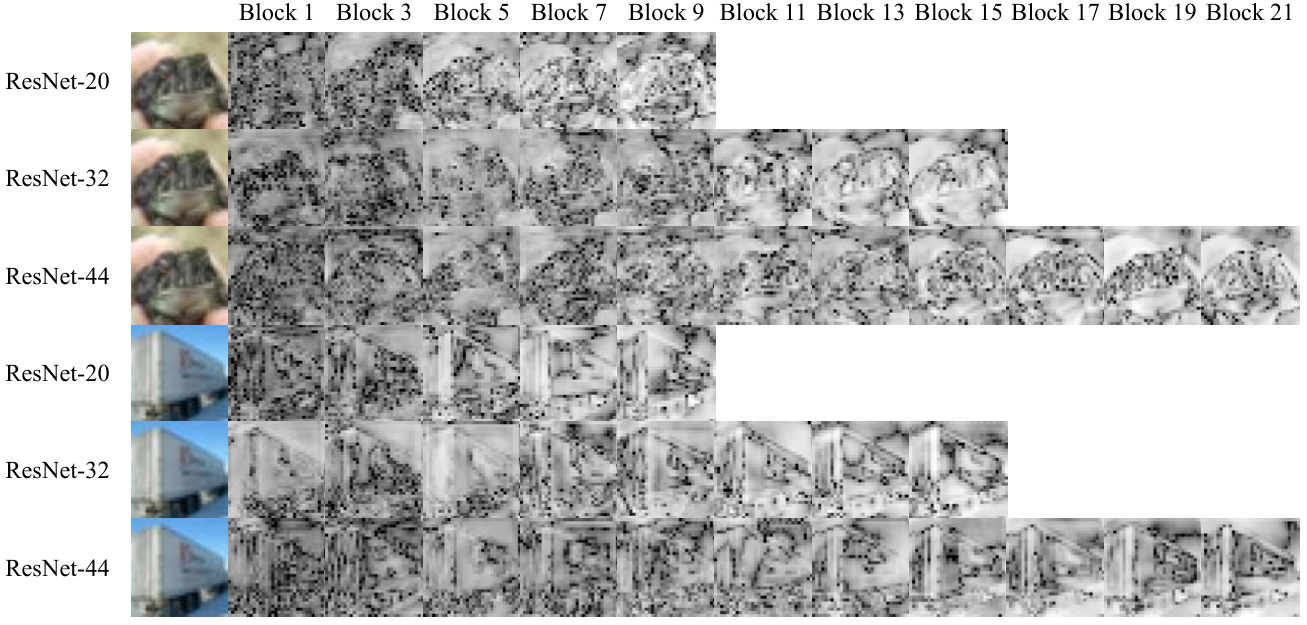}\\
\noindent
\includegraphics[width=0.8\linewidth]{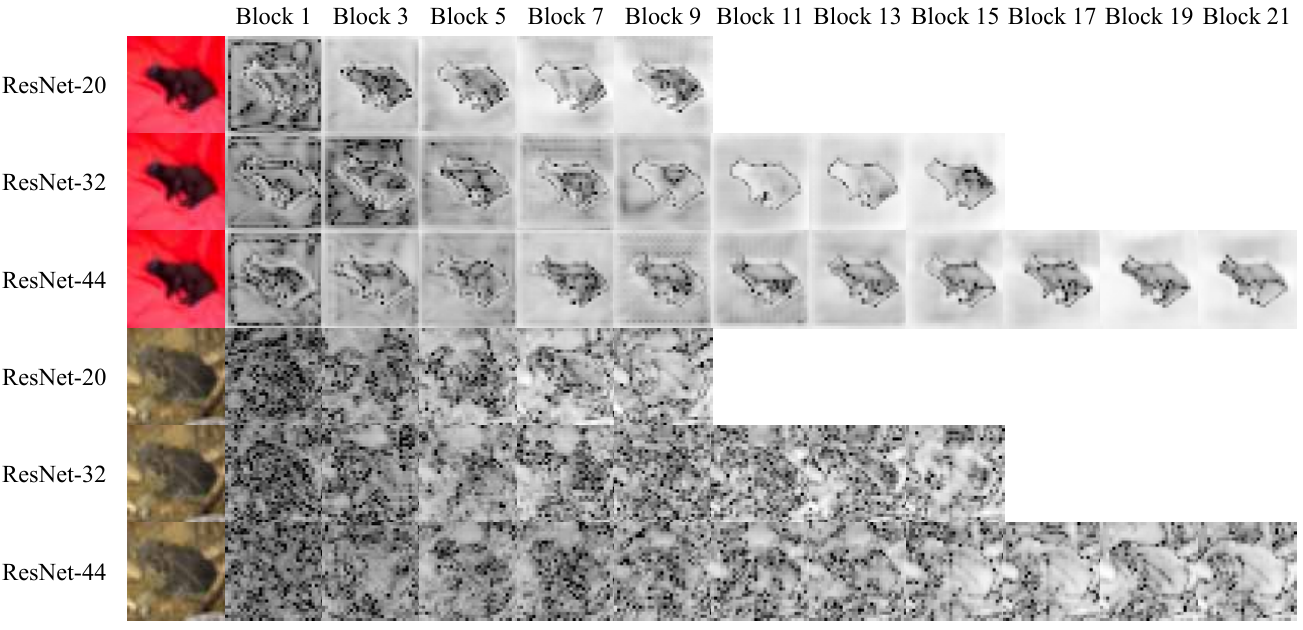}\\
\noindent
\includegraphics[width=0.8\linewidth]{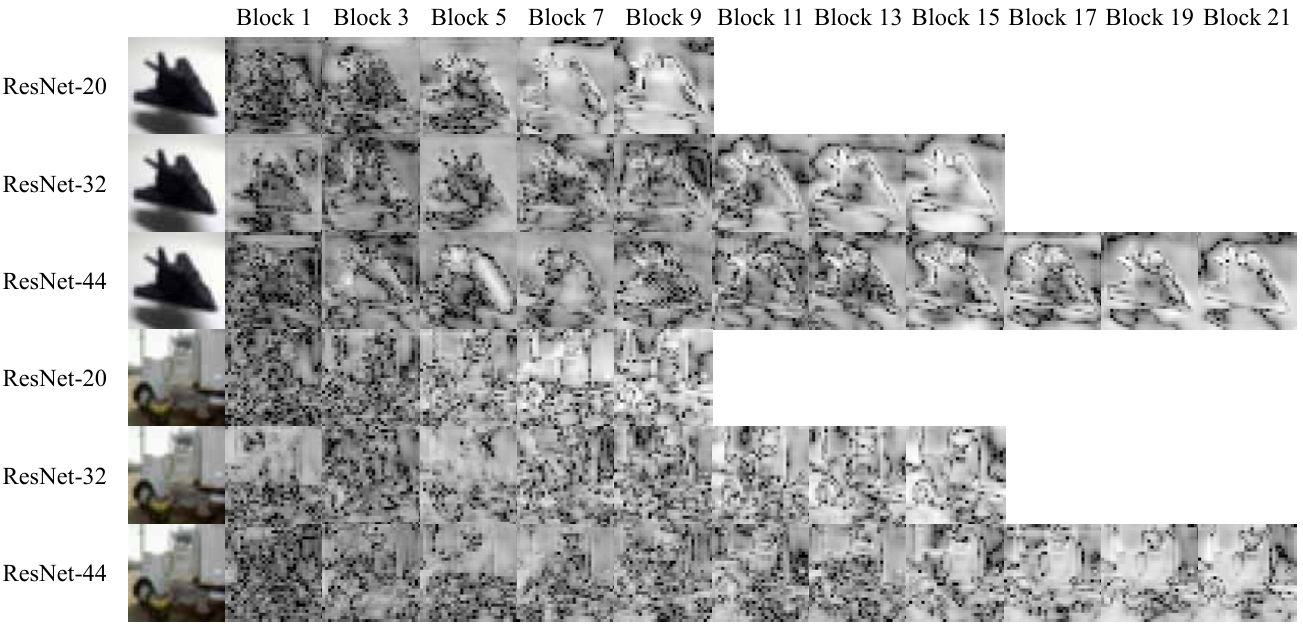}\\
\noindent
\includegraphics[width=0.8\linewidth]{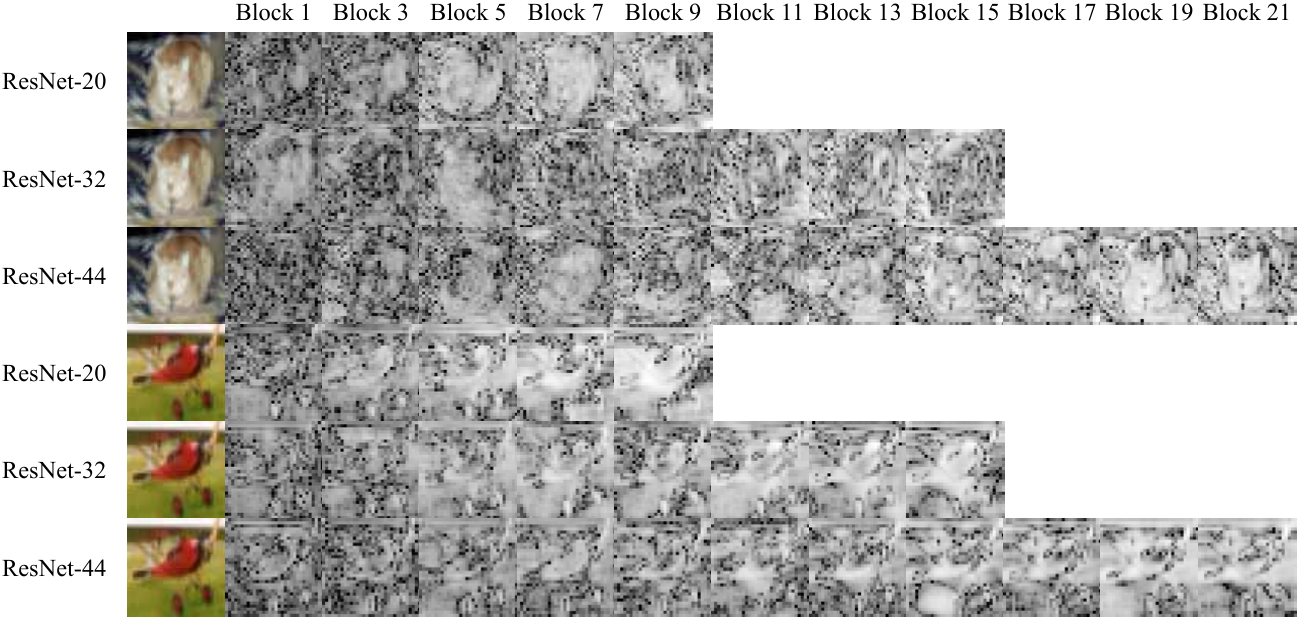}\\
\noindent
\includegraphics[width=0.8\linewidth]{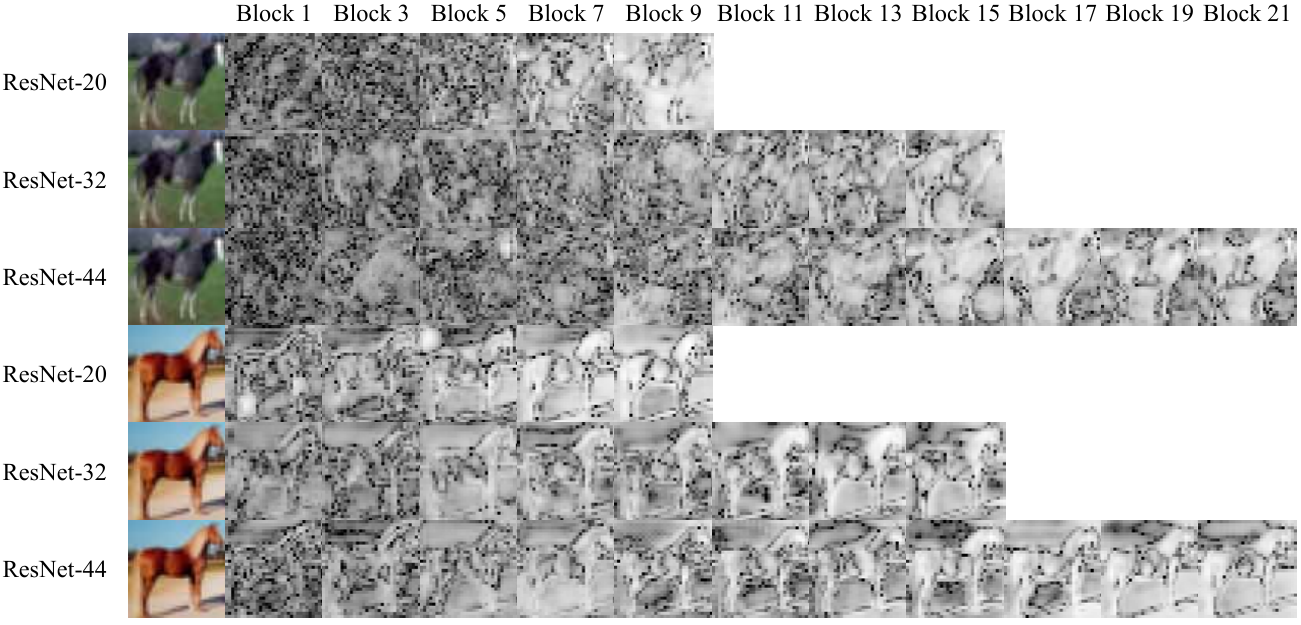}\\
\noindent
\includegraphics[width=0.8\linewidth]{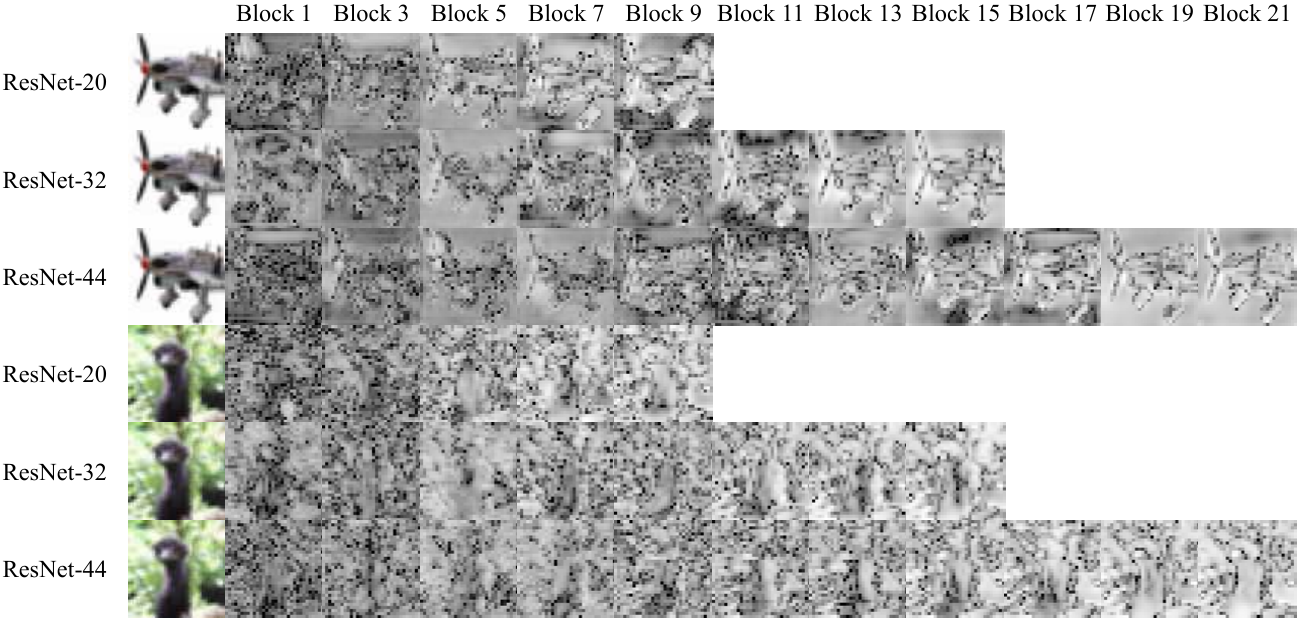}\\
}

\section{Importance maps generated by CAM, Grad-CAM, and Gradient on different layers}
\label{appsec:cam}
The visualization results of importance maps generated by CAM, Grad-CAM, and Gradient on different layers of the VGG-16. The visualized results have been normalized to the unit mean value.
According to Table 1 of the paper, CAM, Grad-CAM, and Gradient cannot generate explanation results that enable fair layer-wise comparisons.

{
\centering
\noindent
\includegraphics[width=0.65\linewidth]{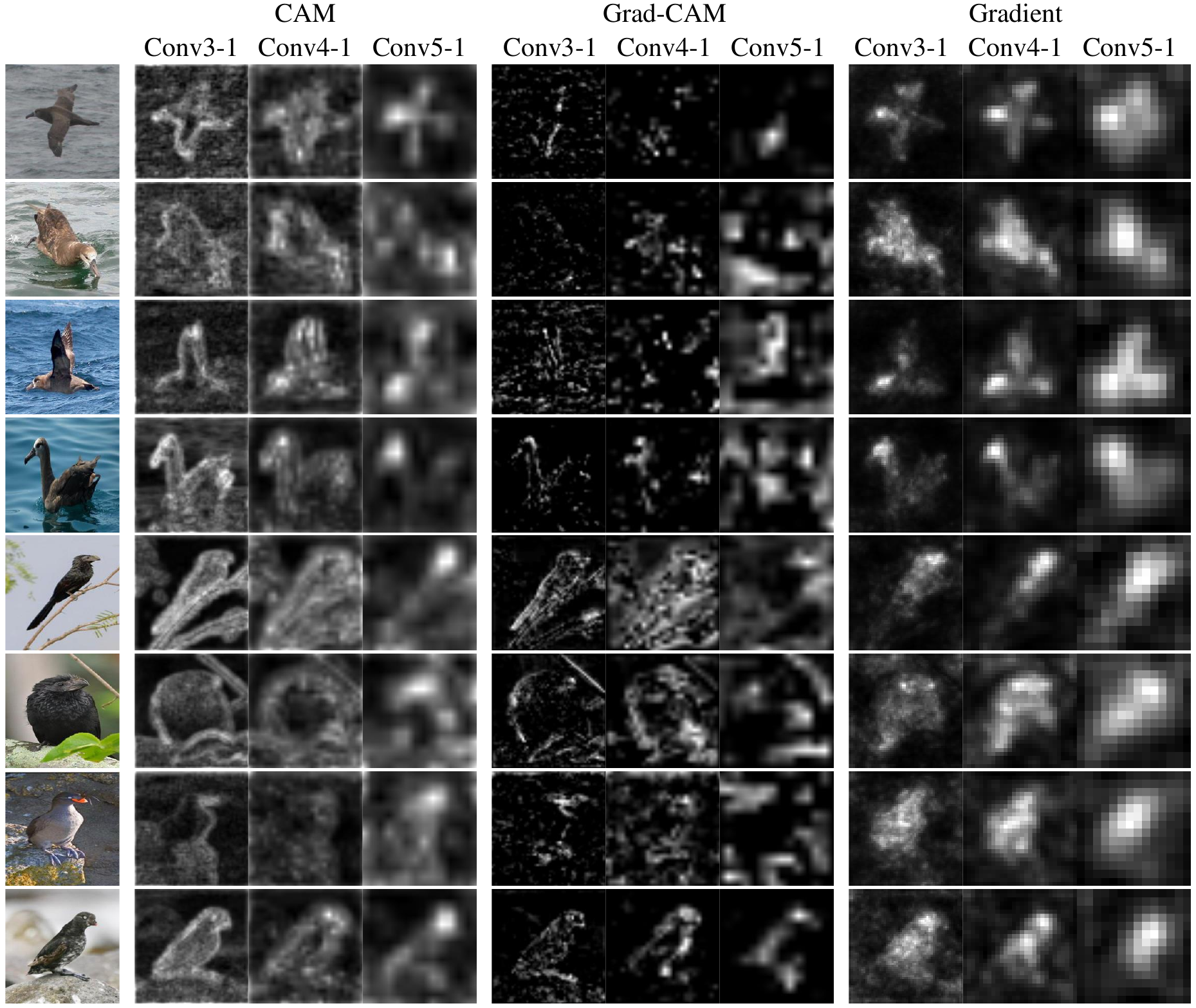}\\
}

\section{Comparisons of pixel-wise RU generated by different decoders}
\label{appsec:diffdecoder}
In this section, we trained two different decoders for the computation of the metric RU, \emph{i.e.} decoders with six or eight residual blocks, respectively. We visualized RU results in the following figure, which shows that the number of residual blocks in the decoder did not affect the results of pixel-wise RU significantly.

{
\centering
\noindent
\includegraphics[width=0.5\linewidth]{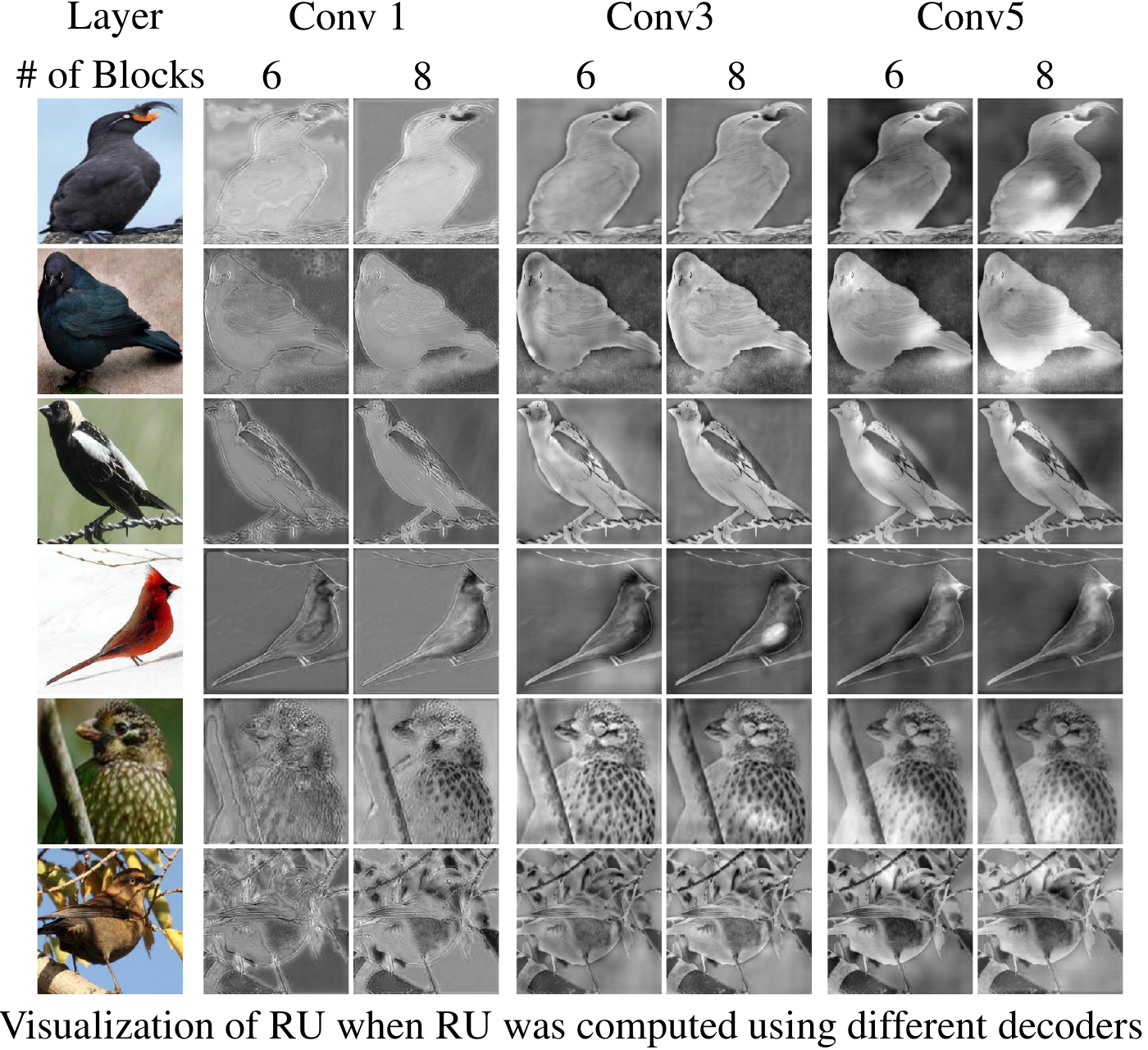}\\
}

\section{Comparisons of pixel-wise CID between the original DNN and the compressed DNN}

When we removed 93.3\% parameters from the VGG-16 network, the network compression did not significantly change the pixel-wise CID of intermediate-layer features.

{\centering
\noindent
\includegraphics[width=0.7\linewidth]{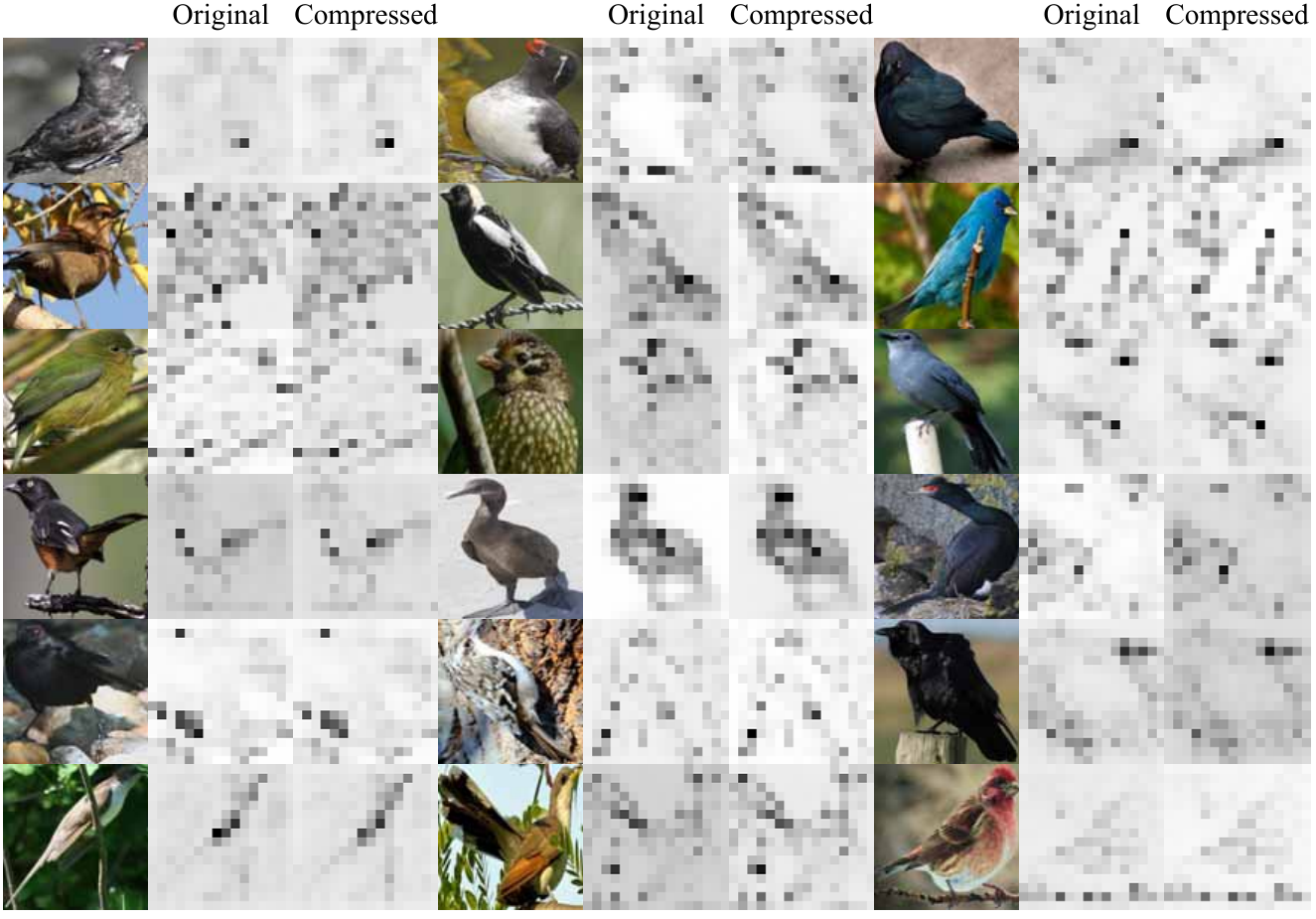}\\

\noindent
\includegraphics[width=0.7\linewidth]{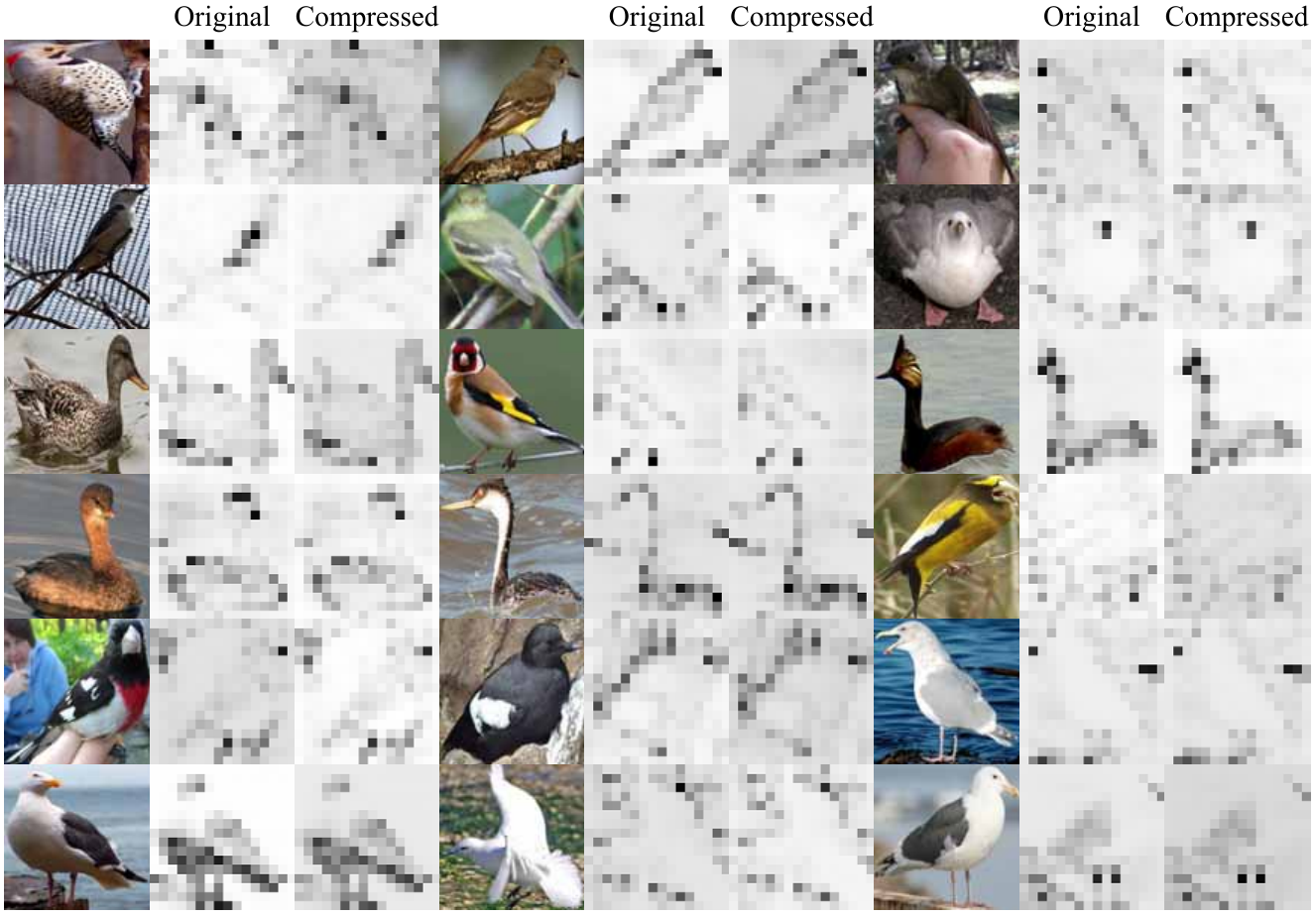}\\

\noindent
\includegraphics[width=0.7\linewidth]{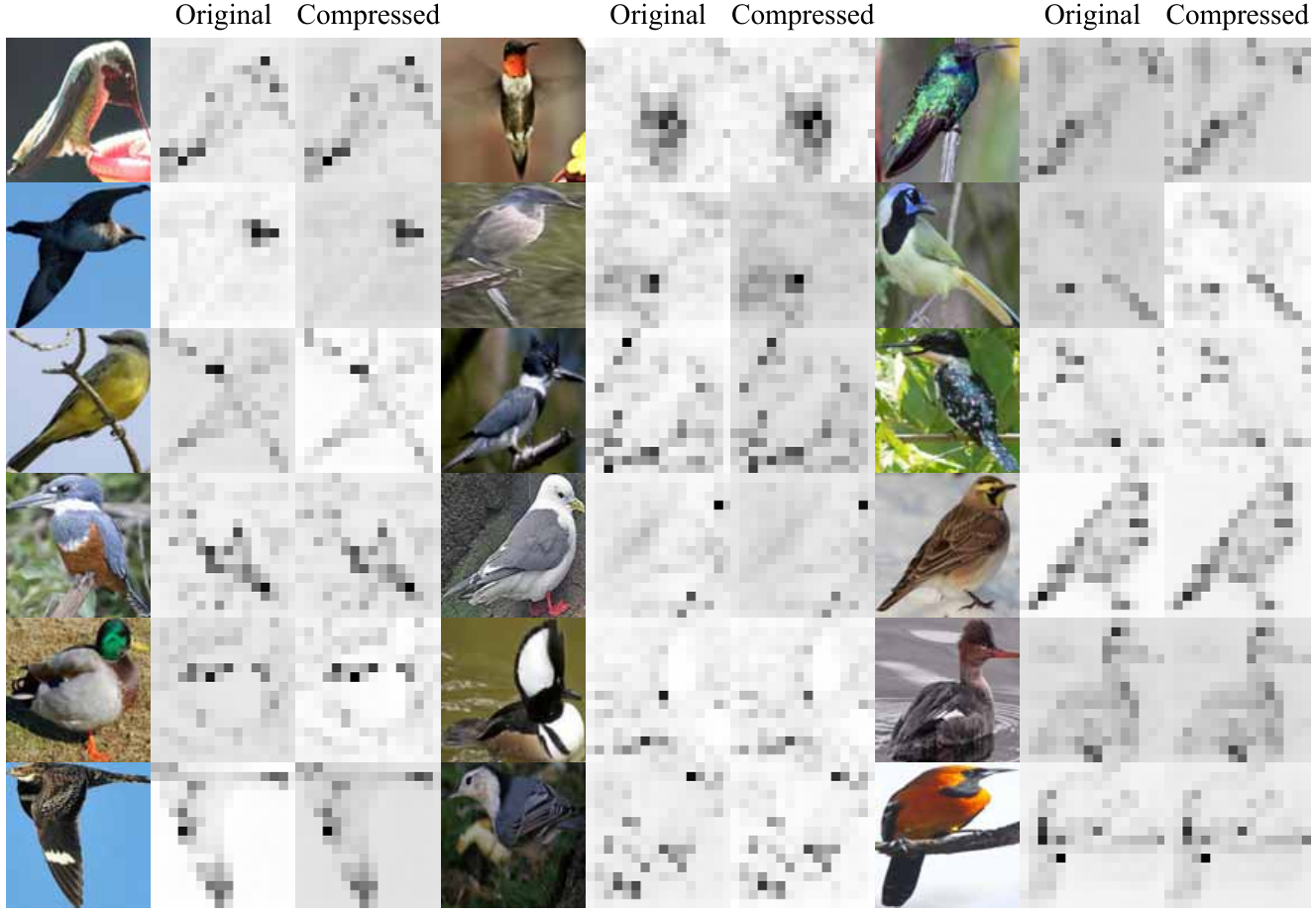}\\

\noindent
\includegraphics[width=0.7\linewidth]{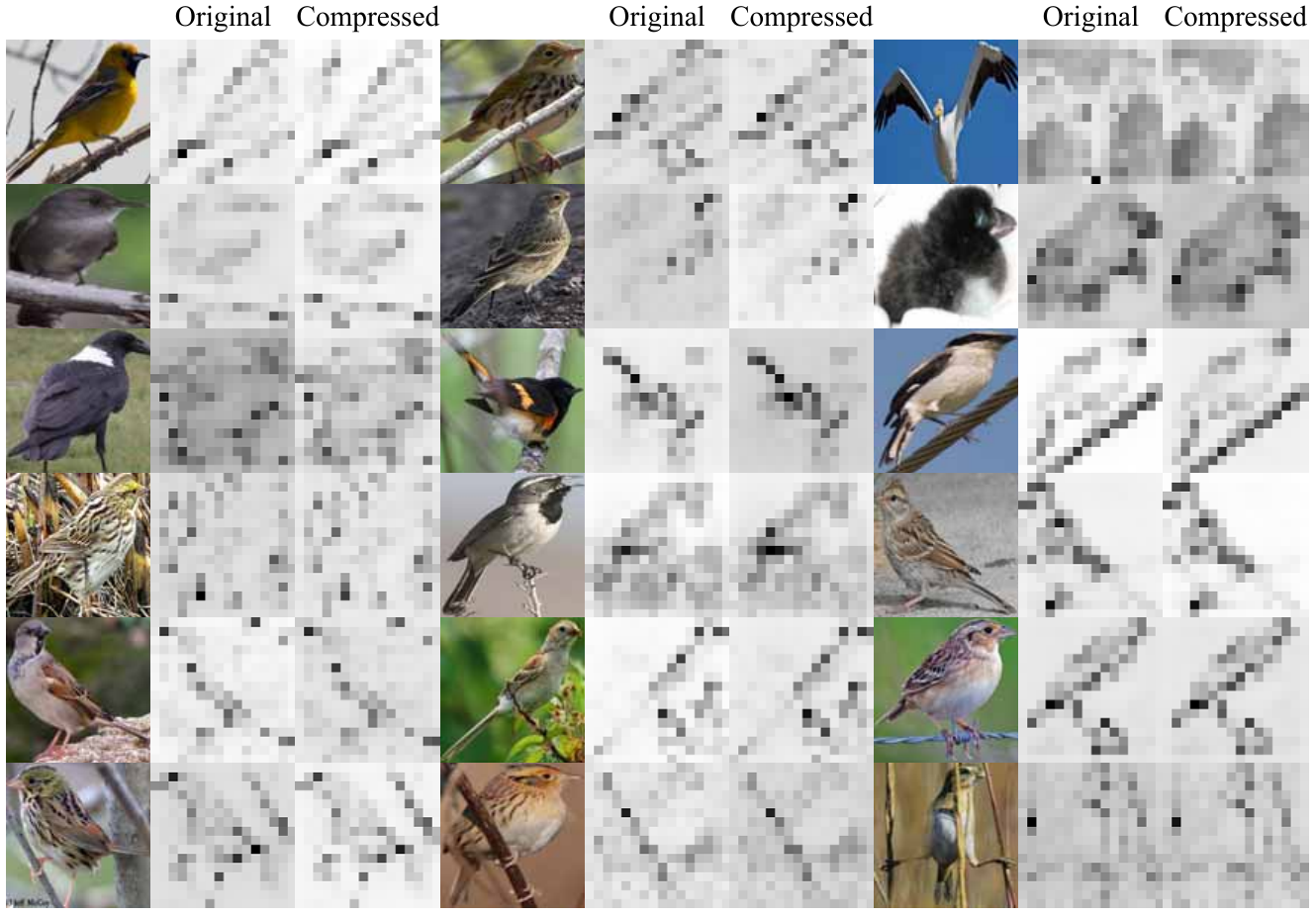}\\

\noindent
\includegraphics[width=0.7\linewidth]{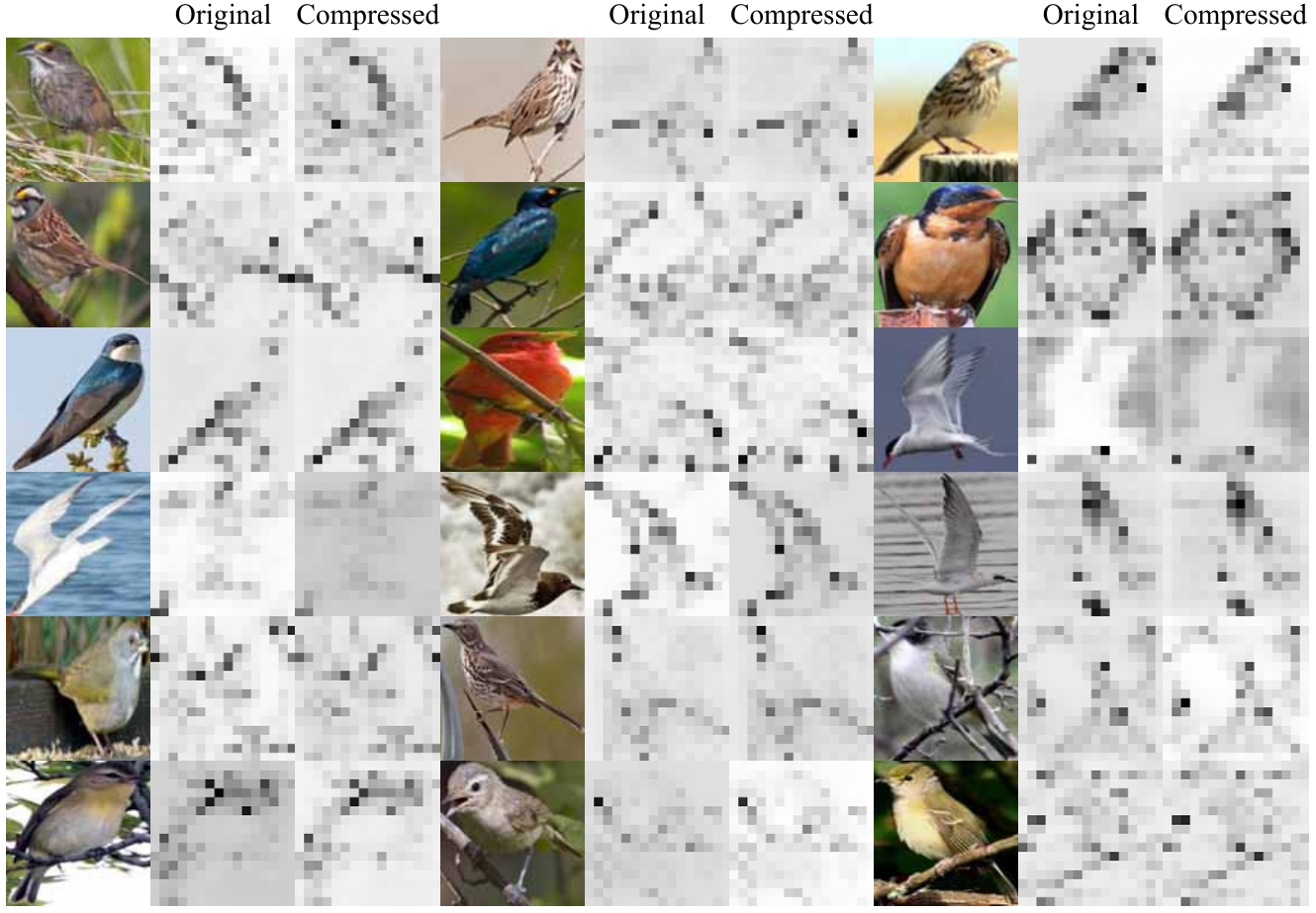}\\
}

\newpage
\section{Comparisons of pixel-wise CID between the original DNN (the teacher) and the DNN learned via knowledge distillation (the student)}

We visualized the pixel-wise CID of VGG-16 networks that were learned using the CUB200-2011 dataset~\cite{CUB200}.

{\centering
\noindent
\includegraphics[width=0.7\linewidth]{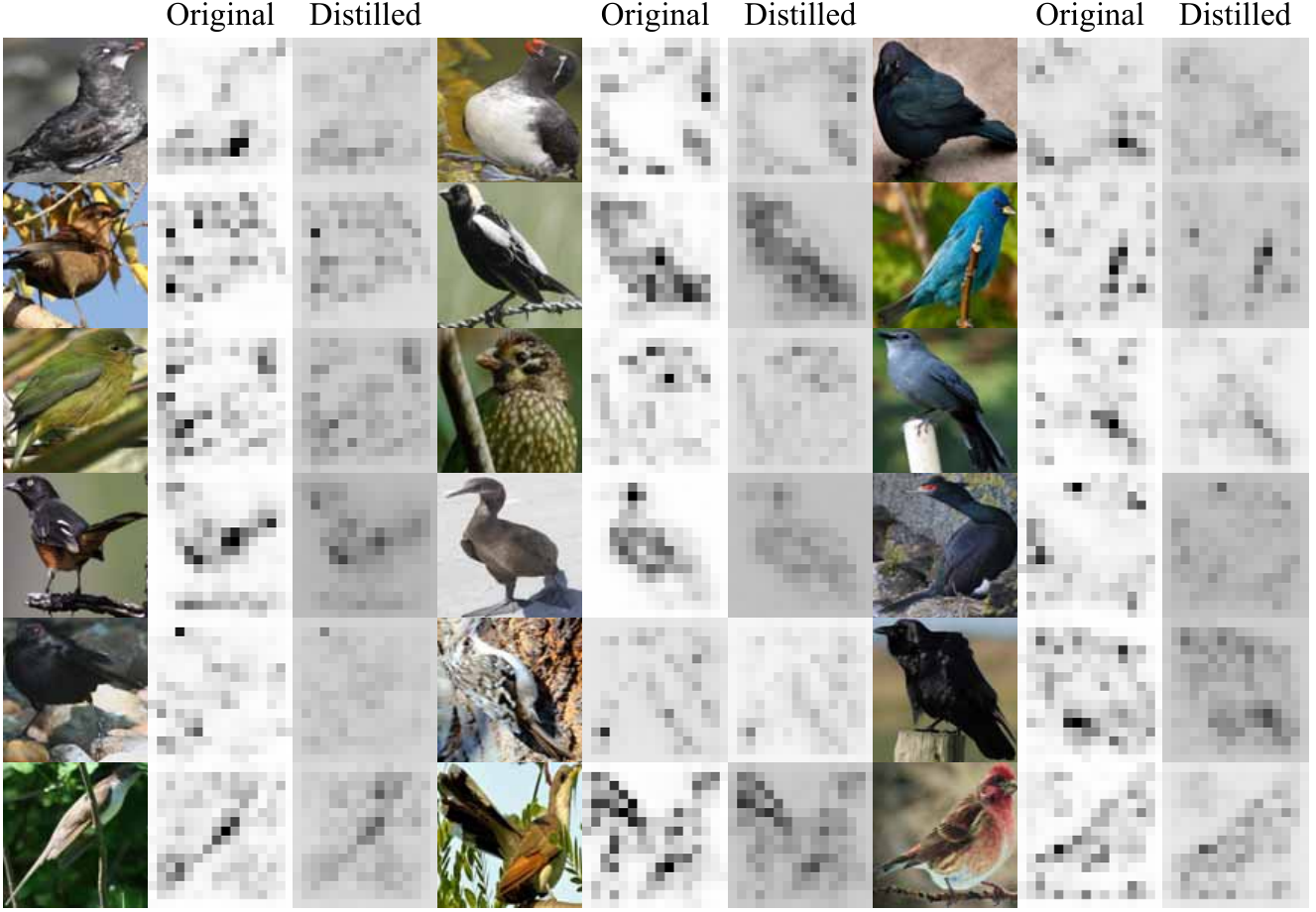}\\

\noindent
\includegraphics[width=0.7\linewidth]{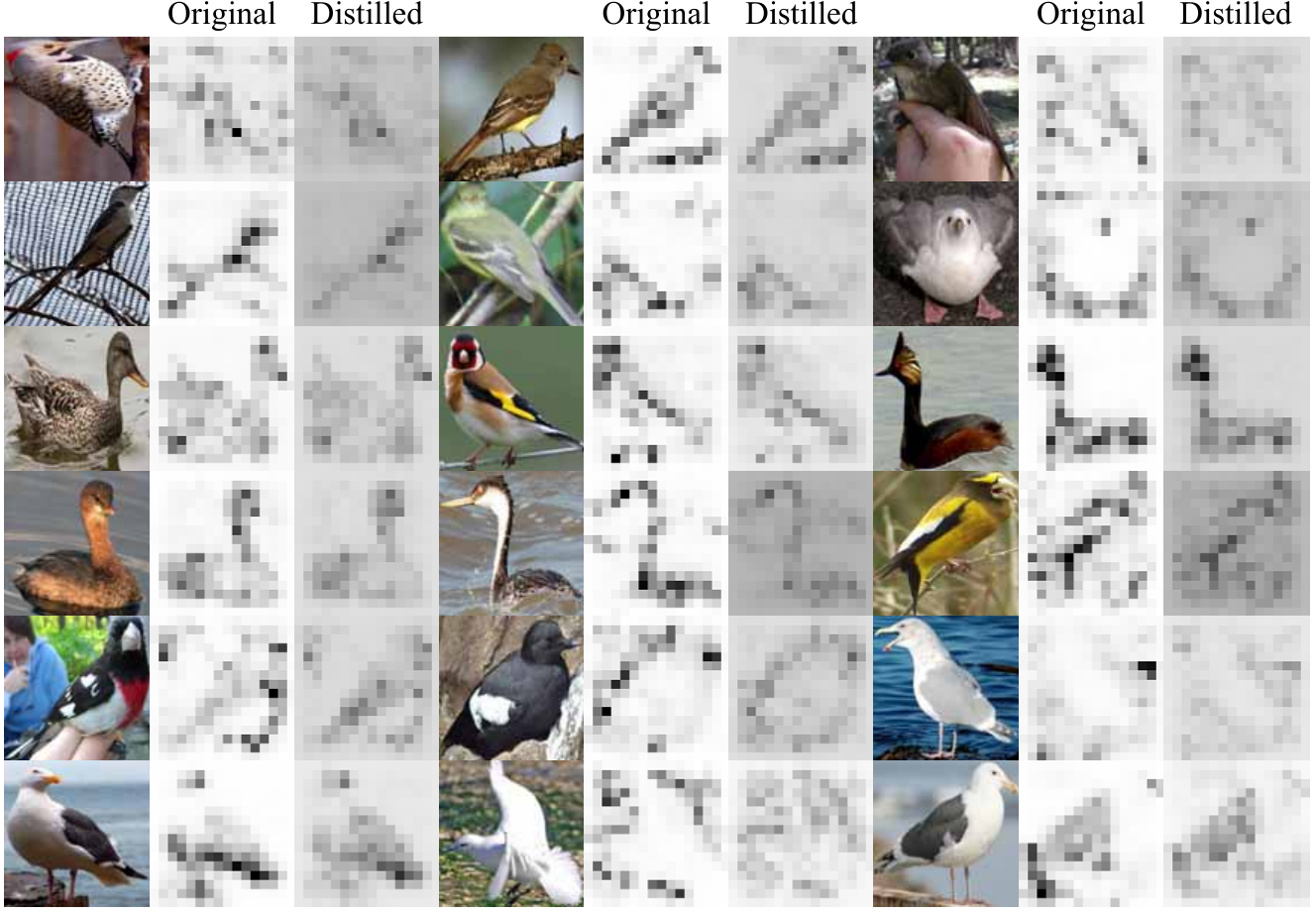}\\

\noindent
\includegraphics[width=0.7\linewidth]{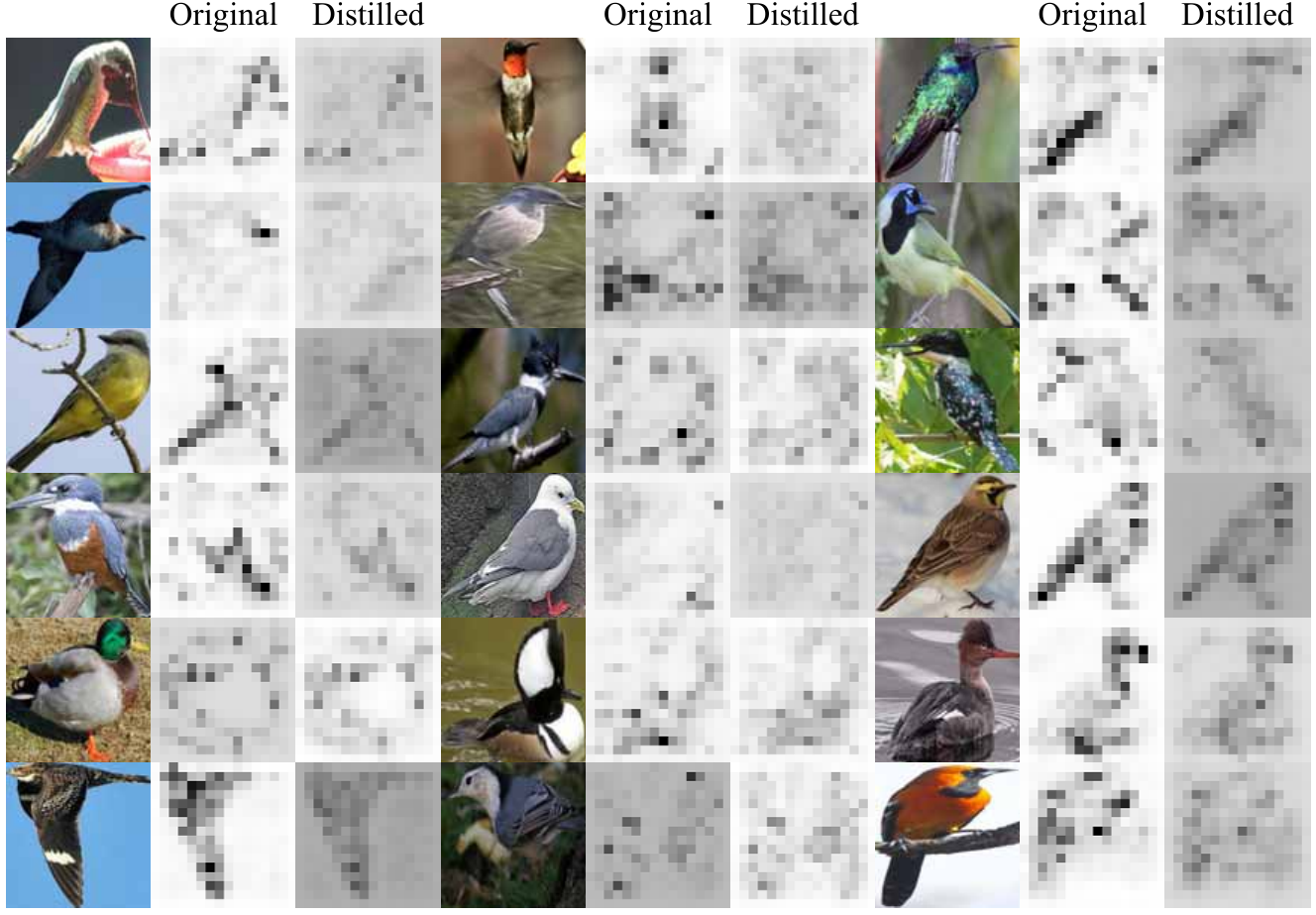}\\

\noindent
\includegraphics[width=0.7\linewidth]{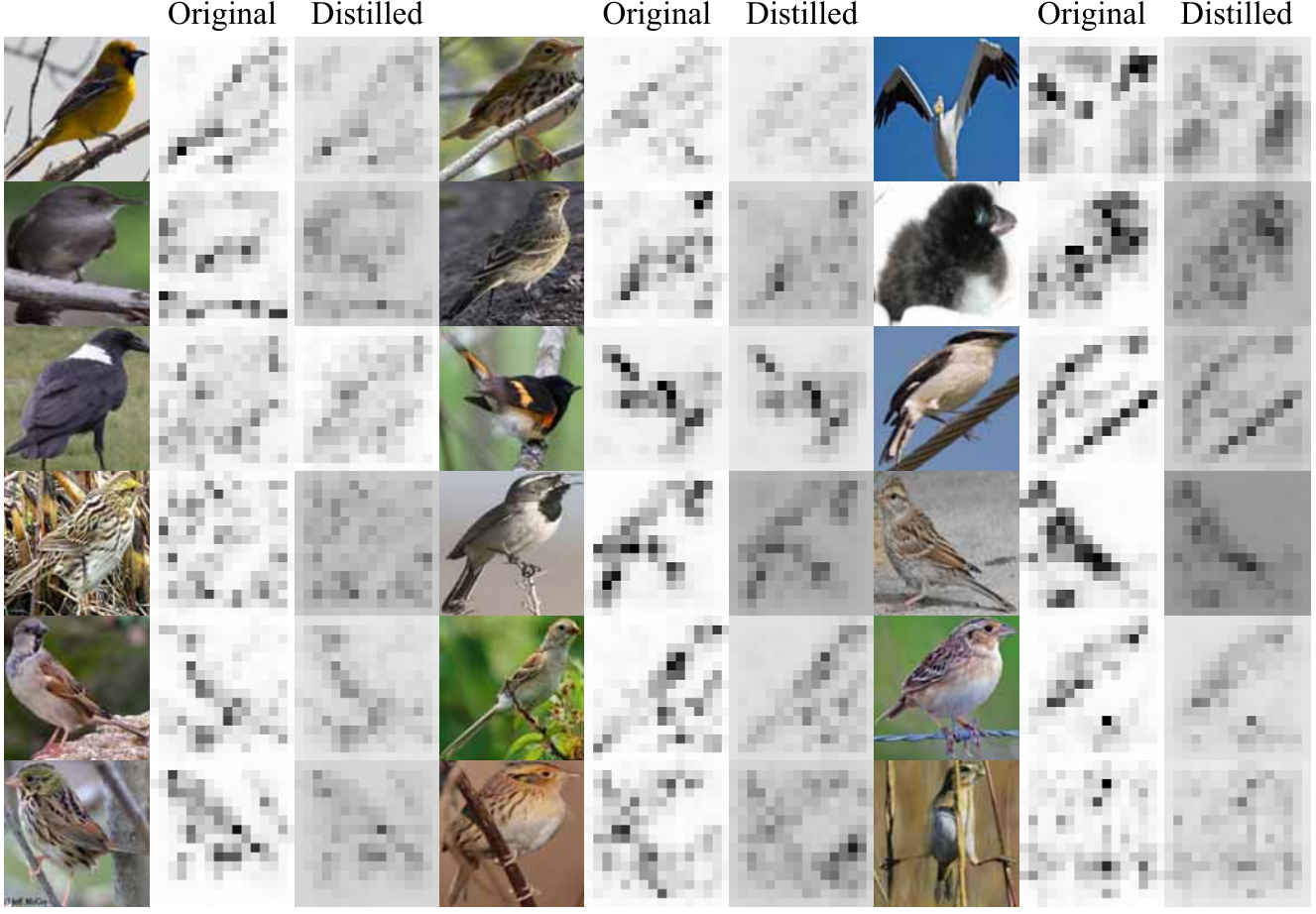}\\

\noindent
\includegraphics[width=0.7\linewidth]{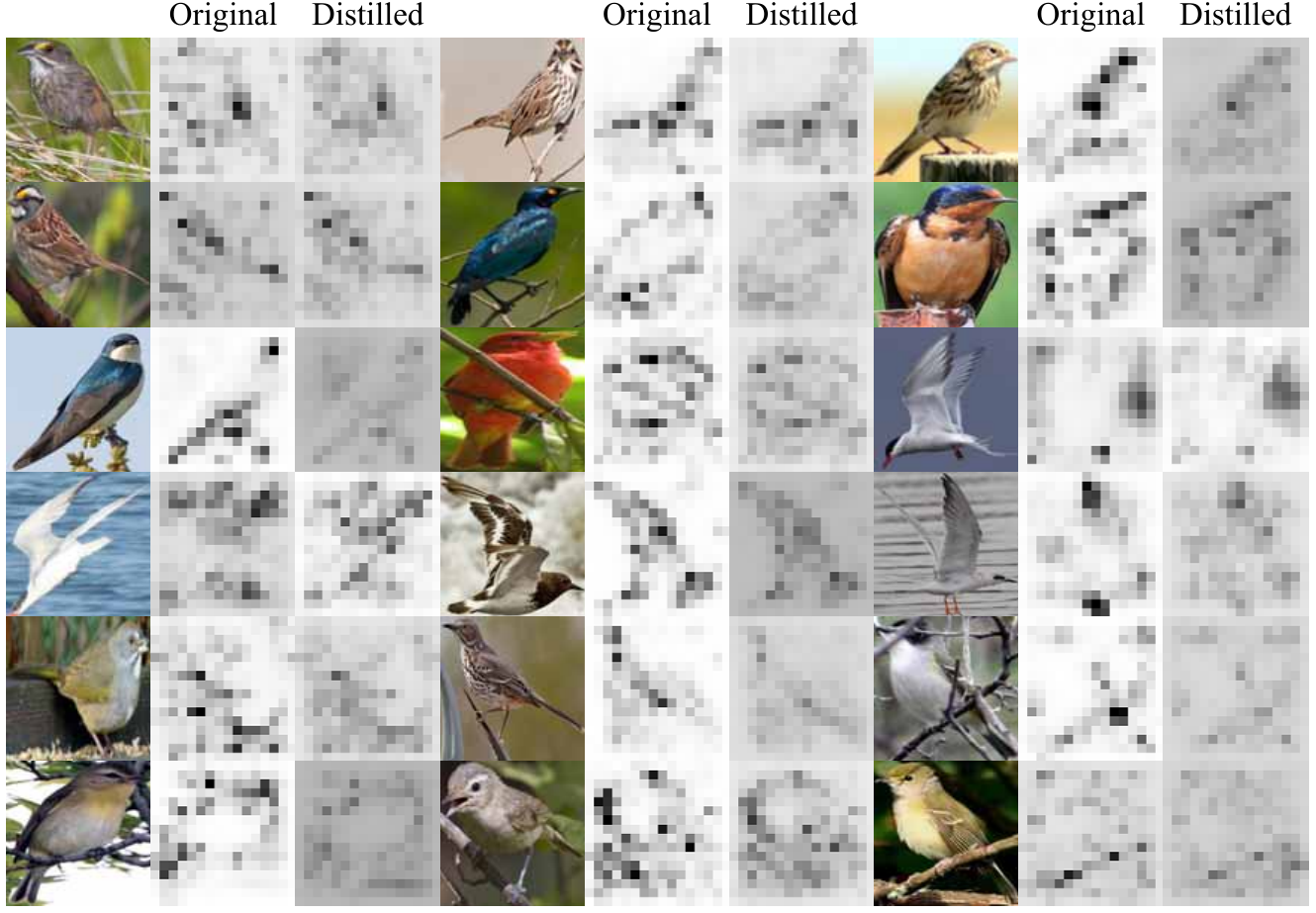}\\
}

\section{Comparisons of pixel-wise CID between the original and the revised DNNs}

We visualized the pixel-wise CID of the original and damaged networks that were learned using the CUB200-2011 dataset~\cite{CUB200}. We focused on the VGG-16 and VGG-19 networks. For each neural network, we revised either the last convolutional layer or the second last convolutional layer to generate the revised networks.

{\centering
\noindent
\includegraphics[width=0.7\linewidth]{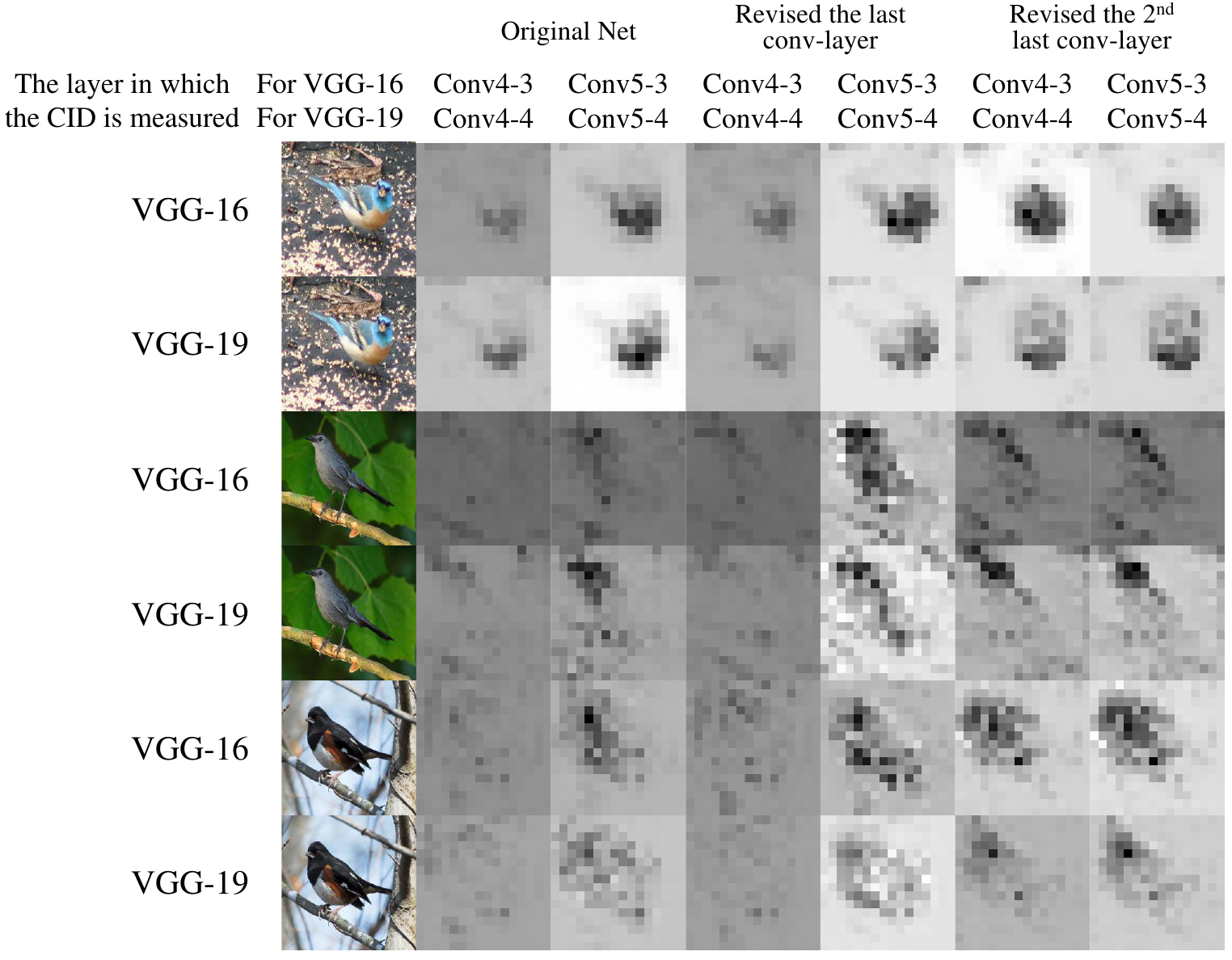}
\\
\noindent
\includegraphics[width=0.7\linewidth]{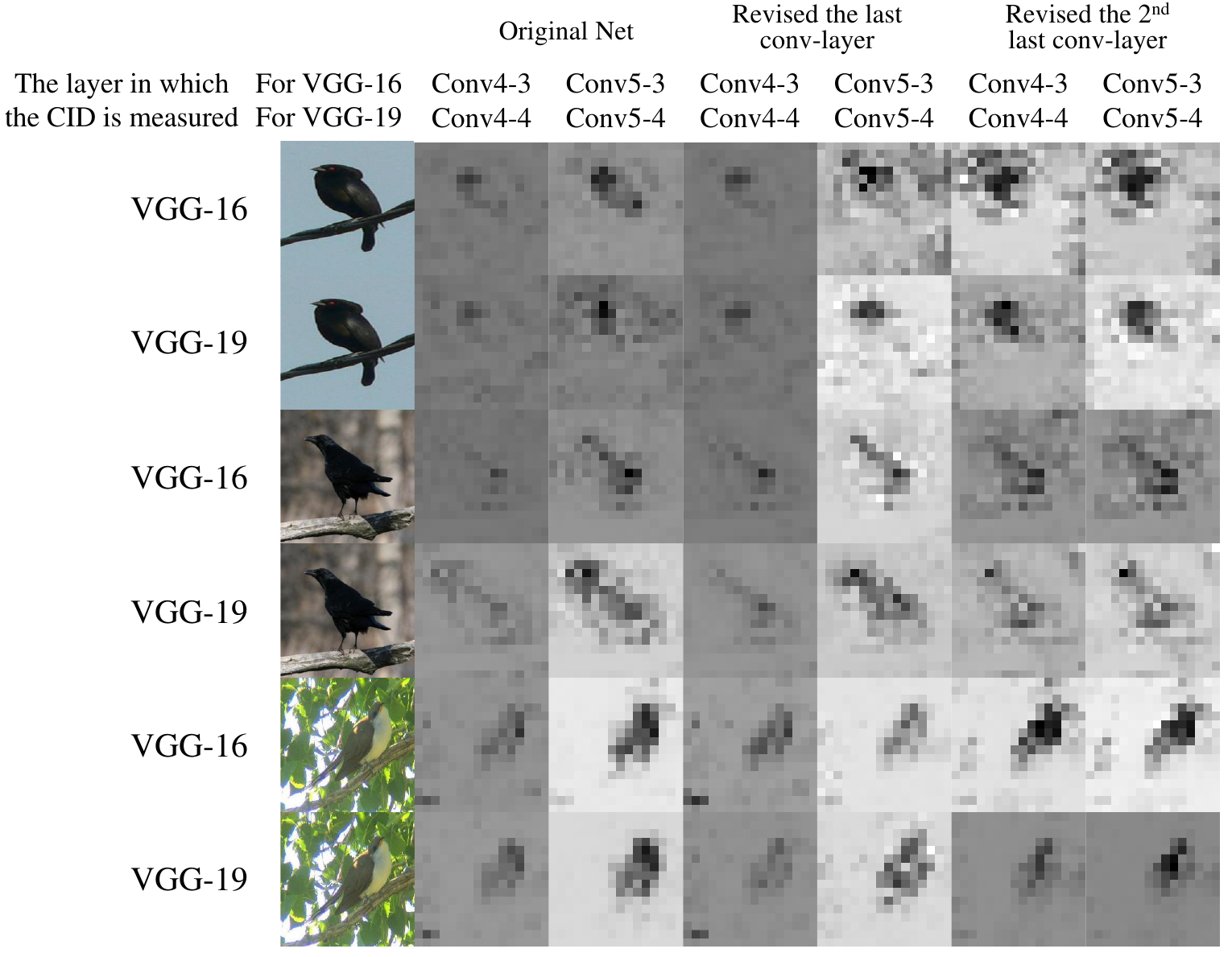}
\\
\noindent
\includegraphics[width=0.7\linewidth]{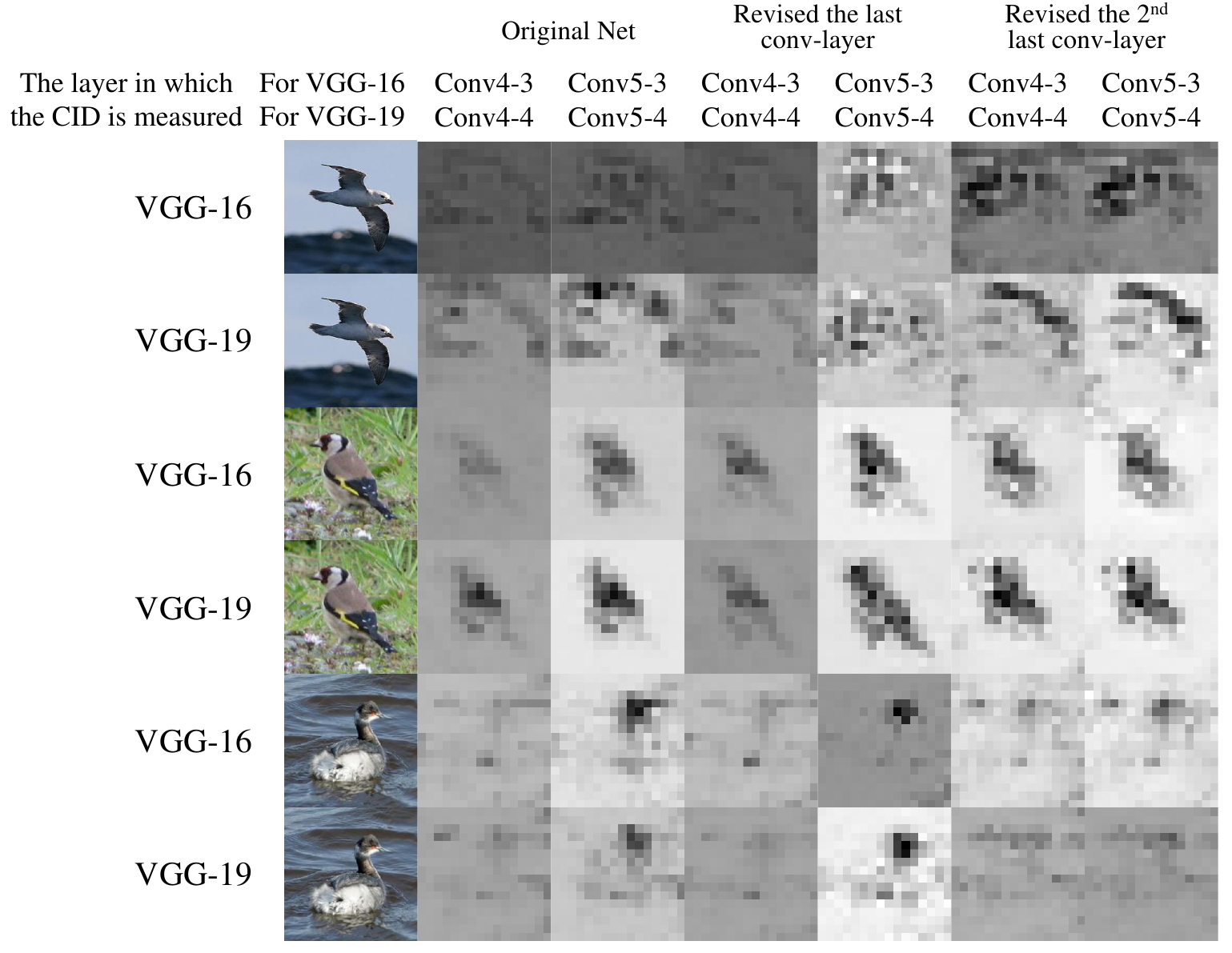}
\\
\noindent
\includegraphics[width=0.7\linewidth]{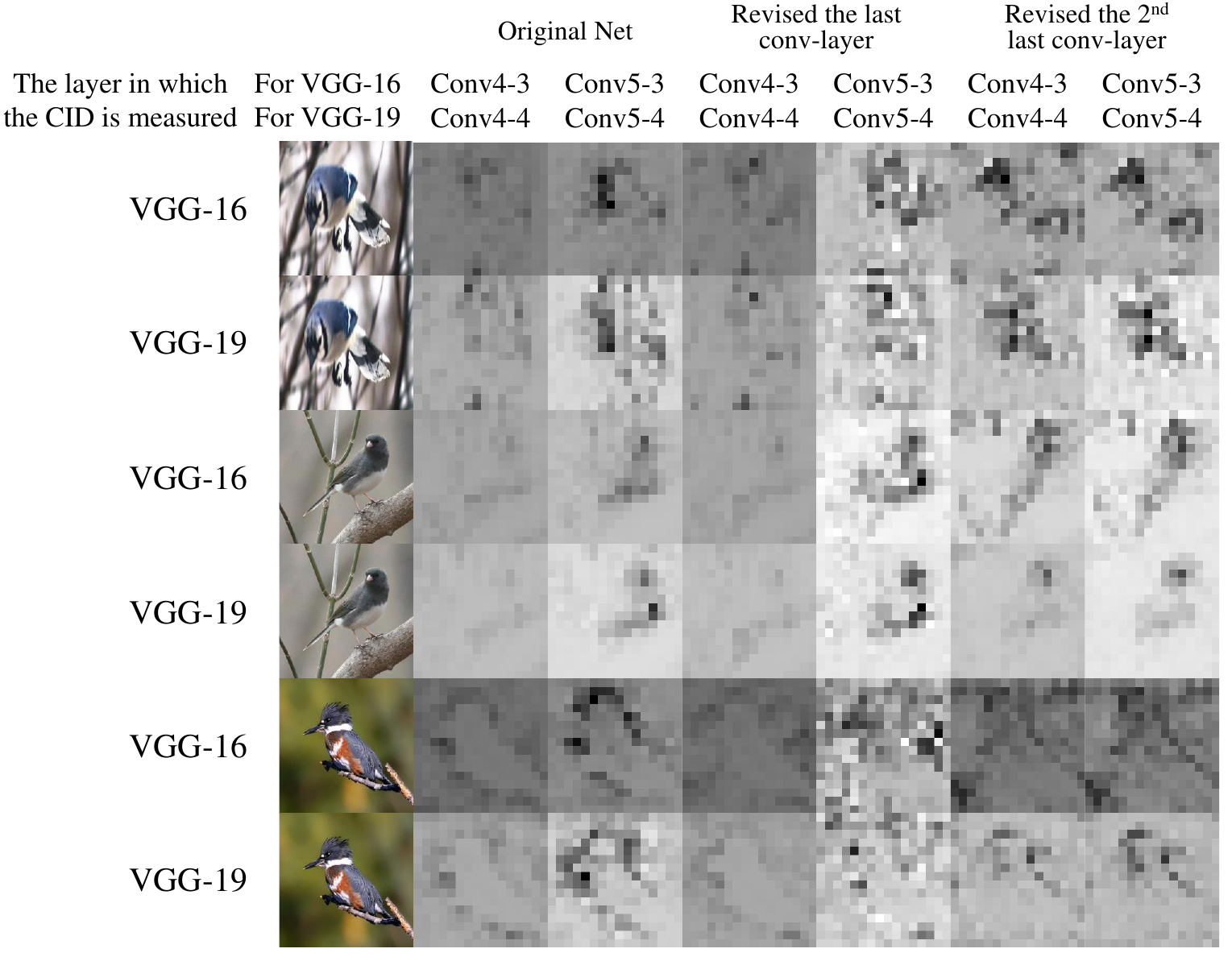}
\\
\noindent
\includegraphics[width=0.7\linewidth]{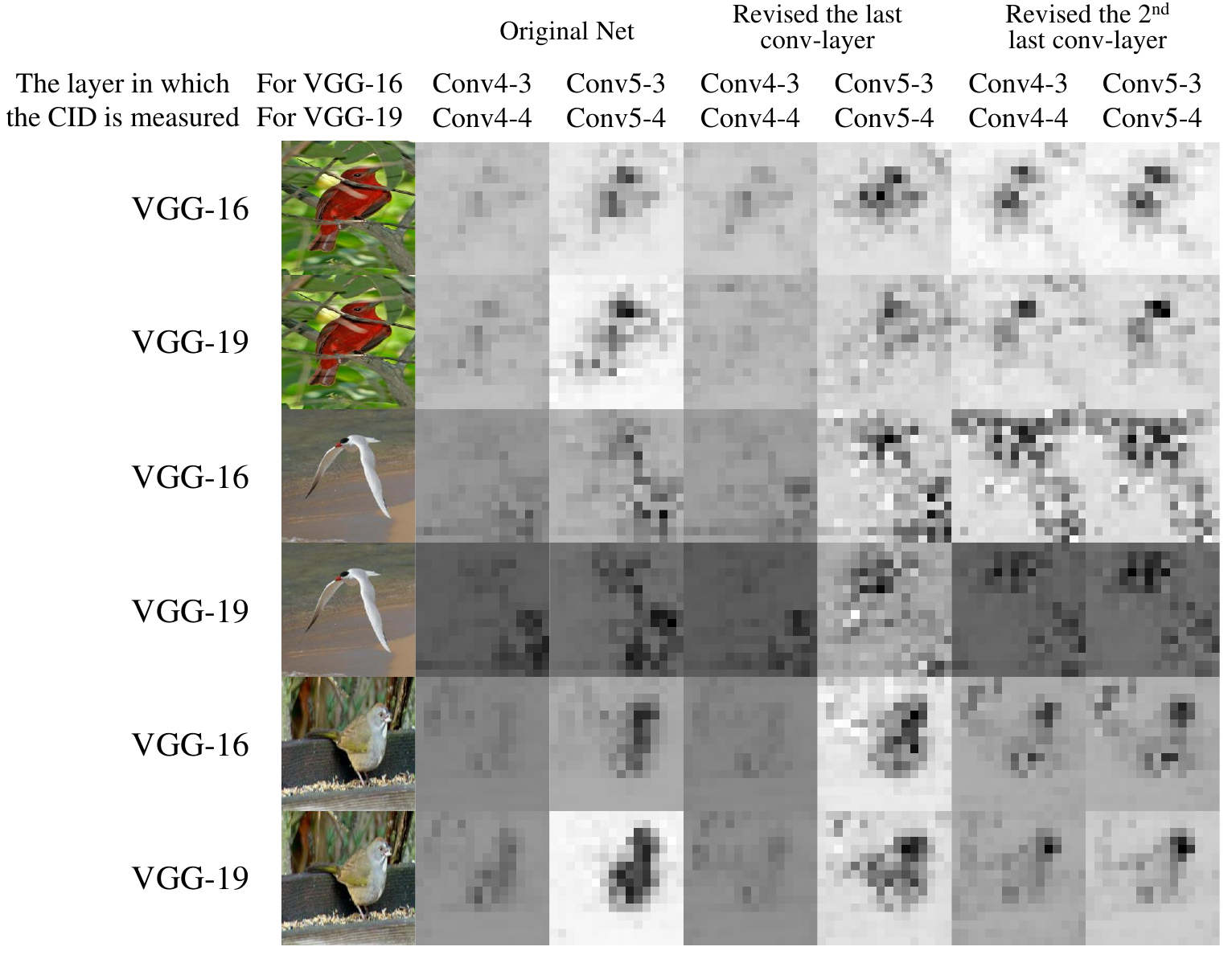}
\\
\noindent
\includegraphics[width=0.7\linewidth]{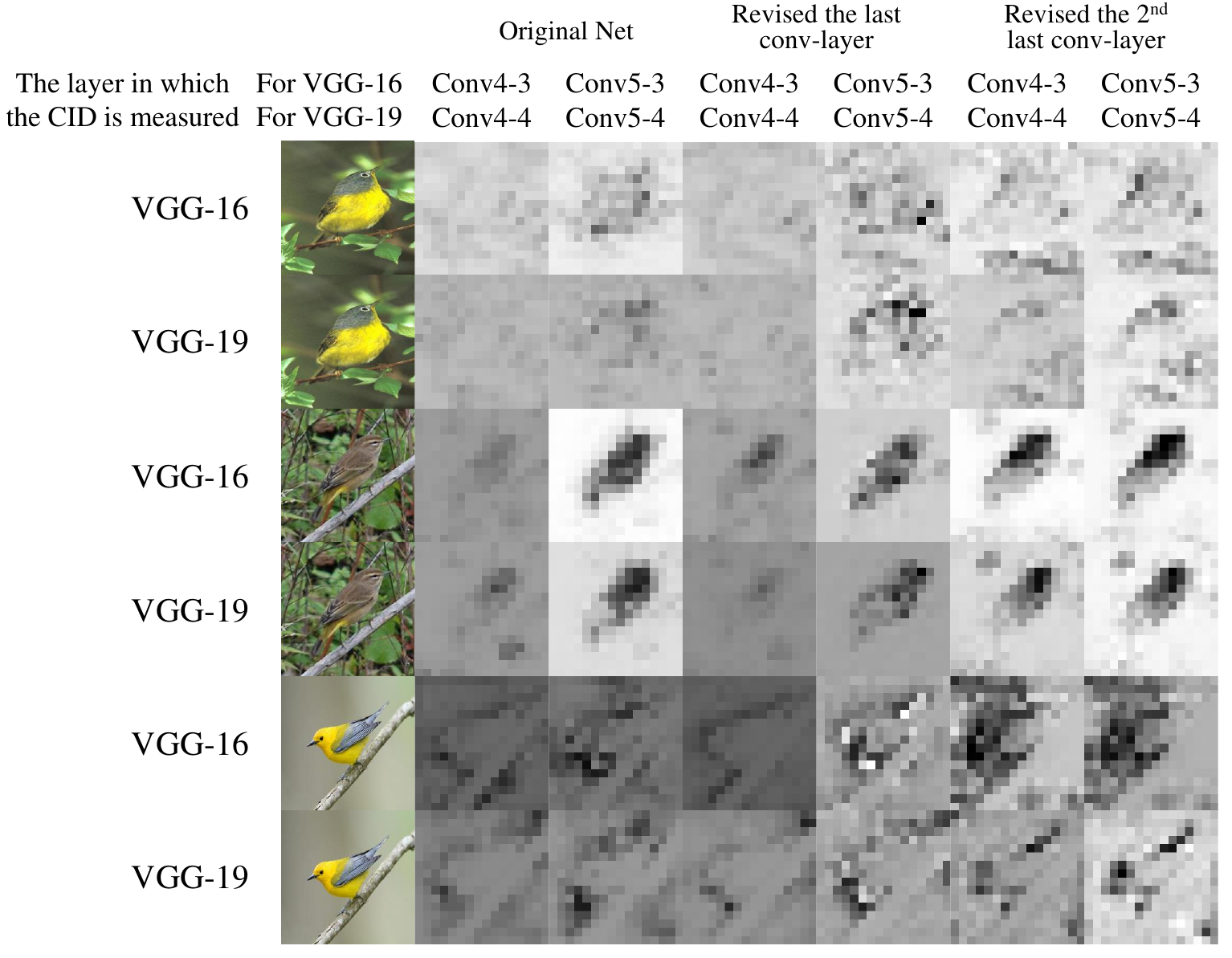}
\\
}


\end{document}